\pdfoutput=1
\documentclass{article}

\PassOptionsToPackage{numbers,sort&compress}{natbib}
\usepackage[main, final]{neurips_2026}
\usepackage[table,dvipsnames]{xcolor}
\usepackage{caption}
\usepackage[utf8]{inputenc}
\usepackage[T1]{fontenc}
\usepackage{hyperref}
\hypersetup{hidelinks}
\usepackage{url}
\usepackage{booktabs}
\usepackage{amsfonts}
\usepackage{nicefrac}
\usepackage{microtype}
\usepackage{multirow}
\usepackage{graphicx}
\usepackage[accsupp]{axessibility}
\usepackage{svg}
\usepackage{float}
\usepackage{amssymb}
\usepackage{pifont}
\usepackage{makecell}
\usepackage{wrapfig}
\usepackage{needspace}
\usepackage{latexml}

\newcommand{\cmark}{\ding{51}}
\newcommand{\xmark}{\ding{55}}
\title{InstantHand: Feed-Forward 3D Gaussian Splatting\\
for Animatable Hand Avatar Reconstruction \\ from a Single Image}

\iflatexml
\author{Chanho Kim, Muhammad Salman Ali, Kyeonghwan Gwak, Muhammad Shaheryar, Incheol Park, Jeongwan On, and Seungryul Baek\textsuperscript{*}\\
UNIST, South Korea\\
\texttt{\{coolho1129, salmanali, khgwak, shaheryar, inchulnim, jeongwan.on, srbaek\}@unist.ac.kr}\\
Project page: \url{https://coolho1129.github.io/InstantHand-page}\\
\textsuperscript{*}Corresponding author}
\else
\author{%
  \textbf{Chanho Kim} \quad
  \textbf{Muhammad Salman Ali} \quad
  \textbf{Kyeonghwan Gwak} \quad
  \textbf{Muhammad Shaheryar} \\
  \textbf{Incheol Park} \quad
  \textbf{Jeongwan On} \quad
  \textbf{Seungryul Baek}\thanks{Corresponding author.} \\[3pt]
  UNIST, South Korea \\[2pt]
\texttt{\{coolho1129, salmanali, khgwak, shaheryar,} \\
\texttt{inchulnim, jeongwan.on, srbaek\}@unist.ac.kr}\\[3pt]
  Project page: \url{https://coolho1129.github.io/InstantHand-page}
}
\fi

\begin{document}

\maketitle

{\centering
\captionsetup{type=figure}
\includegraphics[width=\linewidth]{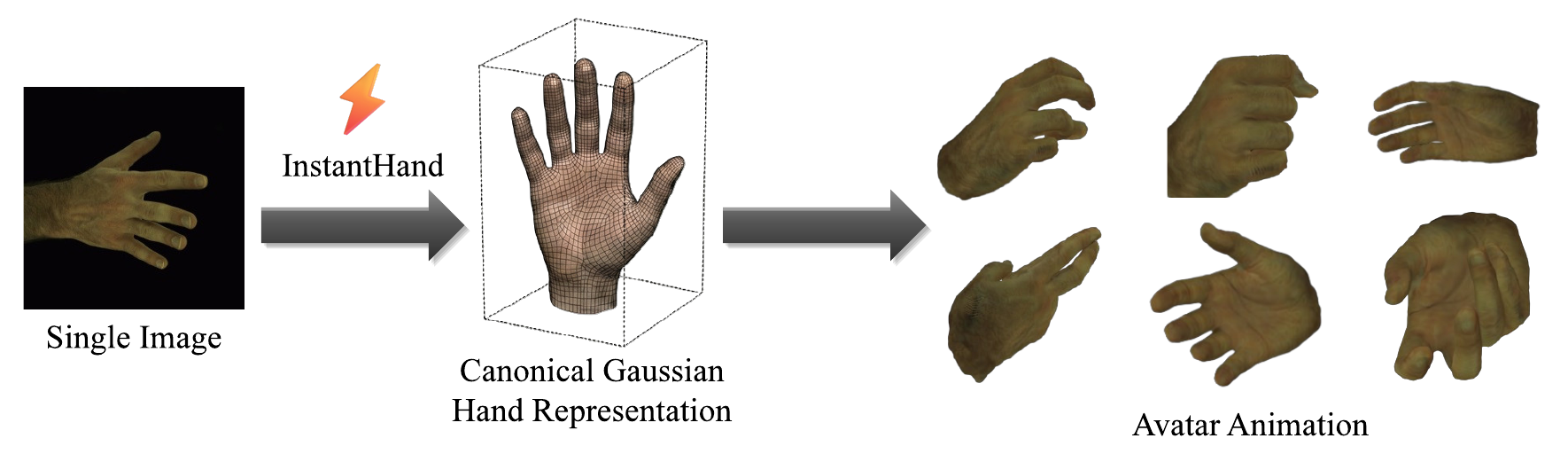}
\captionof{figure}{
Single-image Feed-forward Gaussian Hand Avatar Reconstruction.
InstantHand constructs an animatable 3D Gaussian hand avatar from a single image
without per-subject optimization.
}
\label{fig:hook}
\par}

\begin{abstract}

Hand avatar reconstruction typically relies on multi-view capture or per-subject optimization, limiting its scalability. We introduce \emph{InstantHand}, a feed-forward framework that constructs an animatable 3D Gaussian hand avatar from a single image using off-the-shelf MANO, camera, and mask estimates, without test-time optimization. Image-averaged losses provide insufficient supervision for alpha leakage, which thickens thin finger contours. To address this issue, we propose a \emph{Soft Alpha Band} regularizer that concentrates supervision on silhouette boundaries. Furthermore, colors regressed from image features can drift toward a learned prior even in observed regions. Our \emph{Reliability-gated Appearance Transfer} preserves observed appearance by transferring source colors according to each Gaussian's rendering contribution, while completing unobserved regions using the learned prior conditioned on global source identity. InstantHand constructs an avatar in 2.84 seconds and renders at 161 FPS, matching per-subject-optimized baselines on InterHand2.6M and remaining competitive in zero-shot evaluation on HanCo.

\end{abstract}

\section{Introduction}
\label{sec:intro}

The rapid evolution of digital human technologies is transforming immersive interaction across augmented and virtual reality, telepresence, gaming, and social communication. 
This progress has been driven by the combination of parametric human models, such as FLAME~\citep{FLAME:SiggraphAsia2017} for faces, SMPL~\citep{SMPL:2015} and SMPL-X~\citep{SMPL-X:2019} for bodies, and MANO~\citep{MANO:SIGGRAPHASIA:2017} for hands, with high-fidelity neural rendering techniques including NeRF~\citep{mildenhall2020nerf} and 3D Gaussian Splatting (3DGS)~\citep{kerbl2023gaussiansplatting}. 
Among these components, hands play a central role as the primary interface for fine-grained interaction and dexterous manipulation~\citep{tzionas2016capturing, hasson19_obman, GRAB:2020, chao:cvpr2021, sayem2026handvqa}.
However, despite their importance, creating personalized, animatable, and photorealistic hand avatars from minimal input remains a key challenge for scalable digital human systems.
While substantial progress has been made in single-image 3D avatar reconstruction for head avatars~\citep{chu2024gpavatar, chu2024gagavatar, he2025lam} and full-body avatars~\citep{zhuang2024idolinstant, qiu2025anigs, qiu2025LHM, sim2025persona}, high-fidelity hand avatar modeling from a single image remains comparatively underexplored and particularly challenging.
Hands are highly articulated, thin, and frequently self-occluded~\citep{tzionas2016capturing, zb2017hand, hasson19_obman}, which makes reconstruction from sparse observations inherently 
ill-posed.
Existing hand avatar pipelines address this ill-posedness through additional observations or per-subject optimization. Most methods rely on multi-view capture or monocular video with subject-specific fitting~\citep{handavatar, mundra2023livehand, karunratanakul2023harp}; single-image approaches still require optimization-based inversion at inference~\citep{zheng2024ohta}; and Gaussian-based methods remain dependent on multi-view or video-based reconstruction~\citep{Zhao2025GaussianHand}. Consequently, scalable single-image hand avatar reconstruction without test-time optimization remains largely unresolved.

To eliminate the dependence on dense observations and costly test-time optimization, we aim to directly predict an animatable hand avatar from a single image, given off-the-shelf MANO, camera, and mask estimates, in a feed-forward manner.
We adopt 3DGS~\citep{kerbl2023gaussiansplatting, ali2026compression} as the output representation for its efficient rendering and controllable geometry and appearance. However, feed-forward Gaussian avatar methods developed for heads~\citep{he2025lam} and bodies~\citep{qiu2025LHM} do not transfer directly to hands. First, hands are thin, low-volume, and highly articulated, so even minor errors in Gaussian position, scale, or orientation produce over-expanded or overlapping primitives, thickness artifacts, and degraded fidelity under novel poses. Second, appearance from a single image is ambiguous: existing feed-forward pipelines~\citep{he2025lam, qiu2025LHM} regress appearance directly from image features and tend to fall back on learned priors even in observed regions, weakening input-specific cues such as skin tone and fine texture, while unobserved regions must still be completed plausibly. Feed-forward hand avatar reconstruction therefore requires (i) geometry stabilization for thin, articulated structures and (ii) source-conditioned appearance modeling that preserves reliable input evidence while completing missing regions from priors.

To address these challenges, we propose InstantHand (Fig.~\ref{fig:hook}), a feed-forward 3DGS framework for single-image animatable hand avatar reconstruction. For geometry, we constrain Gaussian placement, scale, and orientation around MANO anchors and supervise rendered alpha on the silhouette band to keep thin fingers compact. For appearance, we introduce Reliability-gated Appearance Transfer, where Gaussians supported by input observations are selectively enhanced with reliable local appearance cues to preserve input-specific details such as skin tone and fine textures. For unobserved regions, appearance is inferred using the learned prior together with a global identity representation, enabling consistent completion under occlusion. Our contributions can be summarized as follows:

\begin{itemize}
\setlength{\itemsep}{1pt}
\setlength{\parskip}{0pt}

\item We present \emph{InstantHand}, a feed-forward framework that reconstructs an animatable canonical 3D Gaussian hand avatar from a single image without per-subject optimization.

\item We propose a Soft Alpha Band (SAB) regularizer that concentrates alpha supervision on silhouette boundaries, reducing alpha leakage and preserving thin finger structures.

\item We introduce Reliability-gated Appearance Transfer, which preserves observed appearance based on each Gaussian's rendering contribution while completing unobserved regions using the learned prior and a global source identity.

\item InstantHand achieves reconstruction quality comparable to per-subject-optimized baselines and competitive zero-shot performance, with avatar construction in 2.84\,s and
rendering at 161\,FPS.

\end{itemize}

\section{Related Work}

\textbf{Hand Pose and Shape Estimation.}
Monocular 3D hand pose and mesh estimation provides the geometric foundation for many hand-centric vision systems~\citep{baek2019pushing, baek2020weakly, Moon_2020_ECCV_I2L-MeshNet, lin2021metro, lin2021mesh_graphormer, 2022MobRecon, cho2023transformer, zhou2024simple, pavlakos2024hamer, dong2024hamba, ismayilzada2025qort, potamias2025wilor}. These methods either predict mesh vertices directly or regress the parameters of the MANO model~\citep{MANO:SIGGRAPHASIA:2017} from RGB images~\citep{baek2019pushing, baek2020weakly}. For example, METRO~\citep{lin2021metro} and Mesh Graphormer~\citep{lin2021mesh_graphormer} use transformers to regress the vertices of a MANO-topology hand mesh from a single image. More recent methods, including HaMeR~\citep{pavlakos2024hamer} and WiLoR~\citep{potamias2025wilor}, regress MANO parameters together with camera parameters from unconstrained images. Our work takes off-the-shelf pose, camera, and mask estimates as input and focuses on reconstructing and rendering an animatable hand avatar.

\textbf{Hand Avatar Reconstruction.}
Hand avatar reconstruction aims to recover subject-specific geometry and appearance for rendering under novel viewpoints and poses. Most existing approaches rely on multi-view images or monocular video sequences and employ neural implicit representations~\citep{mildenhall2020nerf, haider2026inr} or explicit mesh-based pipelines~\citep{corona2022lisa,handavatar,mundra2023livehand,karunratanakul2023harp,guo2023handnerf,kalshetti2024intrinsichand}. Representative methods include HandAvatar~\citep{handavatar}, HARP~\citep{karunratanakul2023harp}, and LiveHand~\citep{mundra2023livehand}, which adopt implicit, mesh-based, and multi-view neural rendering formulations, respectively. To reduce dependence on dense observations, single-image approaches such as OHTA~\citep{zheng2024ohta} incorporate learned priors for one-shot reconstruction, but still require test-time fitting or inversion. Gaussian-based methods, including GaussianHand~\citep{Zhao2025GaussianHand}, use explicit 3D Gaussian primitives for efficient rendering. As summarized in Table~\ref{tab:method_comparison} of the appendix, existing methods require either dense observations or per-subject optimization, regardless of representation. In contrast, our method constructs an animatable Gaussian hand avatar from a single image without test-time optimization.

\textbf{Feed-Forward 3D Gaussian Avatar Reconstruction.}
 Feed-forward avatar methods predict the parameters of animatable 3D Gaussians from one or a few images, avoiding per-subject optimization. This approach has been applied to full- and upper-body avatars by LHM~\citep{qiu2025LHM} and GUAVA~\citep{GUAVA}, and to head avatars by GAGAvatar~\citep{chu2024gagavatar} and LAM~\citep{he2025lam}. However, reconstructing an animatable hand avatar from a single image remains challenging, as thin finger boundaries are sensitive to alpha leakage and the source view provides incomplete appearance information (Sec.~\ref{sec:intro}; Appendix~\ref{sec:supp_related}). We address these challenges with a Soft Alpha Band regularizer that targets boundary leakage and Reliability-gated Appearance Transfer that combines observed appearance with a learned prior and a global source identity.

\begin{figure}[t]
  \centering
   \includegraphics[width=\linewidth]{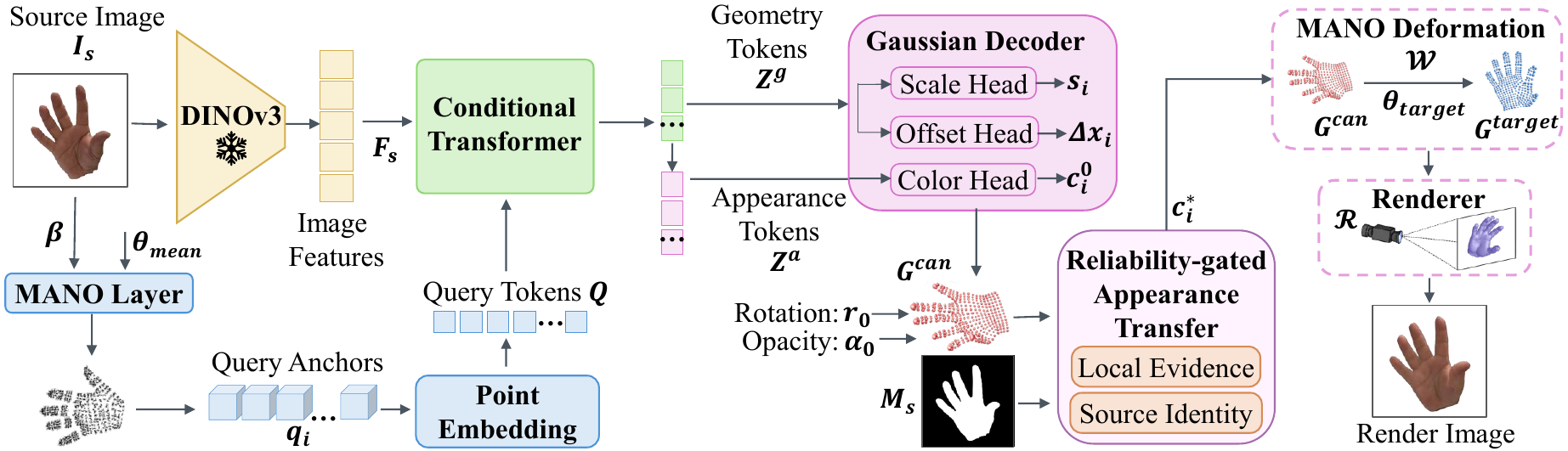}
   \caption{Overview of InstantHand.
A frozen DINOv3 encoder provides image features $\mathbf{F}_s$ that serve as keys and values for a conditional transformer, whose query tokens are point-embedded MANO-surface anchors obtained from the MANO layer $(\boldsymbol{\beta}, \boldsymbol{\theta}_{\mathrm{mean}})$.
The resulting geometry and appearance tokens are decoded by scale, offset, and color heads into a canonical Gaussian hand $\mathbf{G}^{can}$ with fixed rotation $\mathbf{r}_0$ and opacity $\alpha_0$.
Reliability-gated Appearance Transfer then composes local source evidence and a global source identity under the source mask $M_s$ to update only the Gaussian colors ($\mathbf{c}^{0}_i \!\to\! \mathbf{c}^{*}_i$), and the avatar is animated by MANO-driven deformation $\mathcal{W}$ and rendered by $\mathcal{R}$.}
  \label{fig:pipeline}
  \vspace{-0.5cm}
\end{figure}

 \vspace{-0.2cm}
\section{Method}
\label{sec:method}
\vspace{-0.3cm}
Given a single source image $I_s$, InstantHand reconstructs an animatable 3D hand in a single forward pass (Fig.~\ref{fig:pipeline}). Source-image features are fused with MANO-surface query anchors to form per-Gaussian tokens (Sec.~\ref{sec:image_mano_fusion}), which parameterize a canonical Gaussian representation with surface-anchored centers, learned scales, and base colors, while opacity and canonical rotation remain fixed (Sec.~\ref{sec:canonical_gs}). Reliability-gated Appearance Transfer then refines only the Gaussian colors using contribution-aware local evidence and a global source appearance identity (Sec.~\ref{sec:source_conditioned_gaussian_appearance}). The resulting Gaussians are animated via MANO-driven deformation and rendered from arbitrary viewpoints. The two-stage training objective is described in Sec.~\ref{sec:training_objective}.

\subsection{Image--MANO Fusion for Geometry-aligned Tokens}
\label{sec:image_mano_fusion}

\textbf{Image features.}
We extract dense visual features from the source image using a frozen pretrained DINOv3 encoder~\citep{simeoni2025dinov3}.
Intermediate features are aggregated through a lightweight fusion head to produce image tokens:
\begin{equation}
    \mathbf{F}_{s}
    =
    E_{\mathrm{img}}
    \left(
    I_s
    \right)
    \in
    \mathbb{R}^{N_f \times d_{\mathrm{img}}},
\label{eq:img_tokens}
\end{equation}
where $N_f$ denotes the number of image tokens and $d_{\mathrm{img}}$ their dimension; each token captures localized visual evidence, and the resulting tokens are linearly projected to the transformer dimension $d$ when used as keys and values.

\textbf{MANO-surface query anchors.}
To anchor Gaussian primitives on an anatomically valid hand surface, we instantiate canonical query anchors from the MANO layer. Given the shape parameter $\boldsymbol{\beta}$ and the mean pose $\boldsymbol{\theta}_{\mathrm{mean}}$, we define the query anchors as vertices of a subdivided canonical mesh and map them to query tokens via a point-embedding layer $\phi_q$:
{\small
\begin{equation}
\left\{
\mathbf{q}_i
\right\}_{i=1}^{N}
=
\mathrm{Vert}
\left(
\mathcal{S}_{k}
\left(
\mathrm{MANO}
(
\boldsymbol{\beta},
\boldsymbol{\theta}_{\mathrm{mean}}
)
\right)
\right),
\quad
\mathbf{Q}
=
\phi_q
\left(
\left\{
\mathbf{q}_i
\right\}_{i=1}^{N}
\right)
\in
\mathbb{R}^{N \times d}.
\label{eq:anchors}
\end{equation}
}
Here, $\mathcal{S}_{k}$ denotes $k$-level mesh subdivision, with $k=0$ corresponding to the original MANO mesh; we use $k=3$ ($N=49{,}281$; ablated in Appendix~\ref{sec:supp_mano_subdivision}). These MANO-surface query anchors provide a pose-normalized and anatomically structured prior for Gaussian primitives, constraining the reconstruction space before incorporating image evidence.

\textbf{Geometry and appearance tokens.}
We fuse MANO-surface query tokens with image tokens using a conditional transformer (Fig.~\ref{fig:pipeline}), implemented as two sequential streams $\Phi_g$ and $\Phi_a$ and inspired by recent multi-stream transformer designs~\citep{Esser2024ScalingRF}. Query tokens encode canonical surface anchors, while image tokens serve as keys and values that provide dense visual evidence from the source image. Given query tokens $\mathbf{Q}$ and image tokens $\mathbf{F}_{s}$, we compute geometry and appearance tokens as:
\begin{equation}
    \mathbf{Z}^{g}
    =
    \Phi_{g}
    \left(
    \mathbf{Q};
    \mathbf{F}_{s}
    \right),
    \qquad
    \mathbf{Z}^{a}
    =
    \Phi_{a}
    \left(
    \mathbf{Z}^{g};
    \mathbf{F}_{s}
    \right),
\label{eq:tokens}
\end{equation}
\noindent
where $\mathbf{Z}^{g}=\{\mathbf{z}^{g}_i\}_{i=1}^{N}$ and $\mathbf{Z}^{a}=\{\mathbf{z}^{a}_i\}_{i=1}^{N}$, with $\mathbf{Z}^{g},\mathbf{Z}^{a}\in\mathbb{R}^{N\times d}$. 
The geometry tokens remain anchored to the canonical surface while aggregating local visual context, whereas the appearance tokens re-attend to image features under the same MANO-surface indexing, allowing fine-grained appearance cues to be captured independently of the geometry stream.
\vspace{-0.2cm}

\subsection{Canonical Gaussian Representation}
\vspace{-0.2cm}
\label{sec:canonical_gs}

Given the geometry and appearance tokens, we construct a canonical Gaussian hand representation:
\begin{equation}
    \mathbf{G}^{can}
    =
    \left\{
    \left(
    \mathbf{x}_i,
    \mathbf{s}_i,
    \alpha_0,
    \mathbf{r}_0,
    \mathbf{c}^{0}_i
    \right)
    \right\}_{i=1}^{N},
\label{eq:gcan}
\end{equation}
where $\mathbf{x}_i$ denotes the Gaussian center, $\mathbf{s}_i=(s_i,s_i,s_i)^\top$ is an isotropic scale, $\alpha_0$ is a fixed opacity, $\mathbf{r}_0$ is a fixed canonical rotation, and $\mathbf{c}^{0}_i$ is the base RGB color.
Rather than predicting unconstrained Gaussian parameters, we parameterize each Gaussian relative to a MANO-surface anchor $\mathbf{q}_i$ and decode the tokens with a lightweight Gaussian decoder comprising three prediction heads (Fig.~\ref{fig:pipeline}): an offset head $f_\Delta$ and a scale head $f_s$ applied to the geometry token $\mathbf{z}^{g}_i$, and a color head $f_{\mathrm{rgb}}$ applied to the appearance token $\mathbf{z}^{a}_i$:
{\small
\begin{equation}
\Delta \mathbf{x}_i=f_\Delta(\mathbf{z}^{g}_i),\quad
s_i=\rho(f_s(\mathbf{z}^{g}_i)),\quad
\mathbf{s}_i=(s_i,s_i,s_i)^\top,\quad
\mathbf{x}_i=\mathbf{q}_i+\Delta \mathbf{x}_i,\quad
\mathbf{c}^{0}_i=f_{\mathrm{rgb}}(\mathbf{z}^{a}_i).
\label{eq:decoder}
\end{equation}
}
Here, $\rho(\cdot)$ enforces a positive scalar scale, and $\Delta \mathbf{x}_i$ is constrained to remain within a local neighborhood of the anchor.
This surface-anchored residual formulation keeps Gaussians close to the canonical MANO surface while allowing controlled deviations beyond the mesh. To reduce ambiguity from a single image, we fix opacity and canonical rotation and restrict each Gaussian to a learned isotropic scale, which removes unstable interactions among opacity, scale, and orientation and prevents the model from compensating for reconstruction errors through arbitrary Gaussian densities, rotations, or anisotropic stretching (ablation in Appendix~\ref{sec:supp_attr_ablation}). The predicted base color $\mathbf{c}^{0}_i$ is geometry-aligned, but since it is inferred from compact tokens under reconstruction supervision, it may miss source-specific cues such as skin tone and fine texture, so we treat it as a stable anchor and refine it with explicit source-view evidence in Sec.~\ref{sec:source_conditioned_gaussian_appearance}.

\begin{figure}[t]
  \centering
  \includegraphics[width=\linewidth]{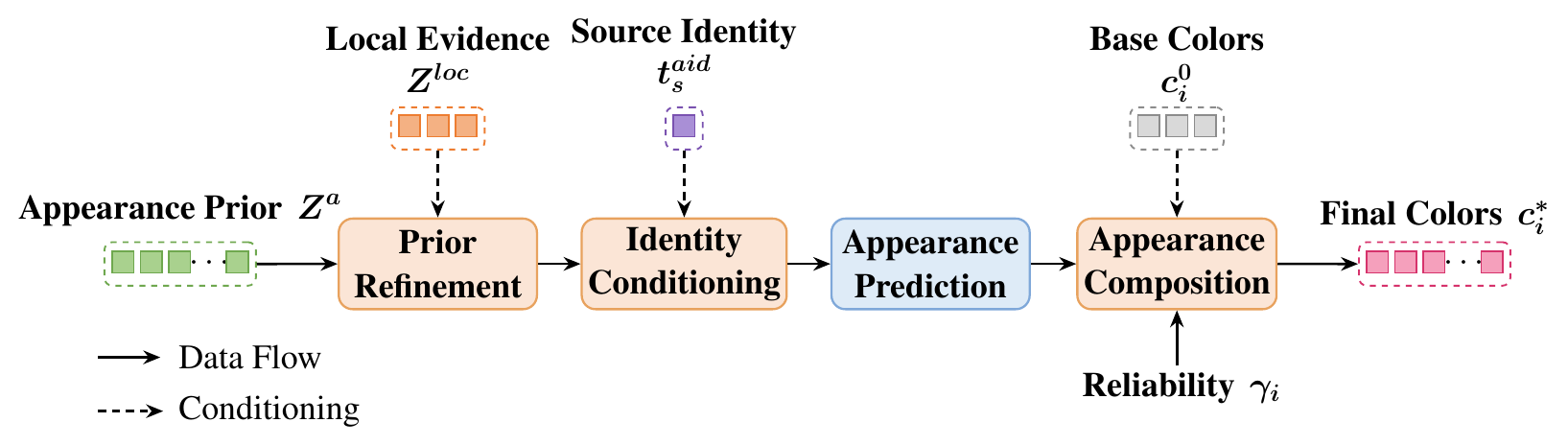} 
\caption{
Reliability-gated Appearance Transfer.
The Stage-1 appearance prior is refined with Gaussian footprint-aligned local evidence (Prior Refinement) and conditioned on the global source identity, which also generates a source-conditioned prior bank (Identity Conditioning).
The composed representation is decoded into an observed color, an appearance gate, and a source-level shift (Appearance Prediction), and Appearance Composition blends these with the base colors under the reliability $\gamma_i$ to produce the final Gaussian colors $\mathbf{c}^{*}_i$.
Dashed arrows denote conditioning. Three further inputs of Eq.~\ref{eq:prior_refine} are omitted for clarity. The source identity also enters Prior Refinement, and the local evidence and base colors also enter Identity Conditioning.
}
  \label{fig:source_appearance}
\end{figure}

\vspace{-0.2cm}
\subsection{Reliability-gated Appearance Transfer}
\label{sec:source_conditioned_gaussian_appearance}

As illustrated in Fig.~\ref{fig:source_appearance}, Reliability-gated Appearance Transfer adapts the canonical appearance to the source image while preserving the underlying 3D geometry. The key idea is to transfer source appearance according to how reliably each Gaussian is observed, using its rendering contribution as a soft reliability cue. Local updates therefore focus on Gaussians that contribute to the source-view rendering and are supported by the foreground mask, while a global source appearance identity token propagates image-level attributes, such as skin tone and illumination bias, across the full Gaussian set, including unobserved regions. The module comprises four steps: Prior Refinement, Identity Conditioning, Appearance Prediction, and Appearance Composition.

\textbf{Gaussian contribution-aware source observation.}
We first animate the canonical Gaussian hand to the source pose and render it from the source camera:
\begin{equation}
    \mathbf{G}^{s}
    =
    \mathcal{W}
    \left(
    \mathbf{G}^{can};
    \boldsymbol{\theta}_{s},
    \boldsymbol{\beta}
    \right),
    \qquad
    \left(
    \hat{I}_{s},
    \{\Omega_i\}_{i=1}^{N},
    \{C_i\}_{i=1}^{N}
    \right)
    =
    \mathcal{R}
    \left(
    \mathbf{G}^{s};
    \mathbf{K}_{s},
    \mathbf{T}_{s}
    \right).
\end{equation}
Here, $\hat{I}_{s}$ denotes the rendered source view, $\Omega_i$ is the projected footprint of the $i$-th Gaussian, and $C_i$ is its accumulated rendering contribution. This source-view rendering establishes a Gaussian-to-image correspondence and provides a soft measure of each Gaussian's participation in forming source-view pixels. Rather than relying on a single projected pixel, we extract local evidence using radius-adaptive Gaussian footprint aggregation. For an image or feature map $X$ and an optional support mask $M$, we define:
\vspace{-0.2cm}
\begin{equation}
    \mathrm{Agg}_{\Omega_i}
    \left(
    X;M
    \right)
    =
    \frac{
    \sum_{\mathbf{p}\in\Omega_i}
    w_i(\mathbf{p})
    M(\mathbf{p})
    X(\mathbf{p})
    }{
    \sum_{\mathbf{p}\in\Omega_i}
    w_i(\mathbf{p})
    M(\mathbf{p})
    +
    \epsilon
    } .
\label{eq:agg}
\end{equation}
Here, $w_i(\mathbf{p})$ denotes a Gaussian kernel centered at the projected center $\mathbf{u}_i$, with bandwidth determined by the projected radius. Setting $M=\mathbf{1}$ yields unmasked aggregation. This footprint-weighted sampling aligns the extracted RGB and feature evidence with the projected support of each Gaussian.
We estimate the reliability of each projected Gaussian by combining its rendering contribution with source foreground support. Specifically, $m_i=\mathrm{Agg}_{\Omega_i}(M_s;\mathbf{1})$ measures foreground support within the projected footprint. We then define the contribution-aware observation reliability, which gates subsequent appearance transfer, as:
{\small
\begin{equation}
    \gamma_i
    =
    \min\!\left(
    \frac{\sqrt{C_i}\, m_i}
    {\mathrm{P}_{\zeta}\!\left(
    \left\{ \sqrt{C_j}\, m_j \right\}_{j\in\mathcal{O}}
    \right)}
    ,\;
    1
    \right)
    \cdot
    \mathbf{1}\!\left[\, i \in \mathcal{O} \,\right],
    \qquad
    \mathcal{O}
    =
    \left\{\,
    j
    \,:\,
    C_j > \kappa\,\mathrm{med}(C_{+}),\;
    m_j > \tau_m
    \,\right\}.
    \label{eq:gamma}
\end{equation}
}

Here, $\mathcal{O}$ denotes the observed set, $\mathrm{med}(C_{+})$ the median of nonzero rendering contributions, and $\mathrm{P}_{\zeta}(\cdot)$ the $\zeta$-th percentile operator. We fix $(\kappa,\tau_m,\zeta)=(0.5,0.5,95)$ a priori without tuning on evaluation data. The reliability $\gamma_i$ is computed deterministically and is not learned; performance is insensitive to these constants (Appendices~\ref{sec:supp_reliability} and~\ref{sec:supp_gamma_sensitivity}). A Gaussian receives high reliability only when it both contributes to the source-view rendering and is supported by the foreground mask. Unlike the binary visible/invisible split that FlexiAvatar~\citep{tiruneh2026flexiavatar} uses to restrict per-subject optimization to observed body regions, $\gamma_i$ provides graded reliability within $\mathcal{O}$, with a bounded discontinuity at its boundary due to the indicator function.

\vspace{-0.1cm}
\textbf{Contribution-aware local evidence and global source appearance identity.}
For each Gaussian, we aggregate source evidence over its projected footprint:
\begin{equation}
    \bar{\mathbf{c}}^{src}_{i}
    =
    \mathrm{Agg}_{\Omega_i}
    \left(
    I_s;
    M_s
    \right),
    \qquad
    \bar{\mathbf{c}}^{ren}_{i}
    =
    \mathrm{Agg}_{\Omega_i}
    \left(
    \hat{I}_{s};
    M_s
    \right),
    \qquad
    \mathbf{f}^{x}_{i}
    =
    \mathrm{Agg}_{\Omega_i}
    \left(
    \mathbf{F}^{x}_{s};
    M_s
    \right).
\label{eq:footprint}
\end{equation}
Here, $\mathbf{F}^{x}_{s}$ denotes local image features used for footprint aggregation. The local evidence token is encoded as:
\vspace{-0.2cm}
\begin{equation}
    \mathbf{z}^{loc}_{i}
    =
    f_{loc}
    \left(
    \left[
    \bar{\mathbf{c}}^{src}_{i},
    \bar{\mathbf{c}}^{ren}_{i},
    \bar{\mathbf{c}}^{src}_{i}
    -
    \bar{\mathbf{c}}^{ren}_{i},
    \mathbf{f}^{x}_{i},
    \gamma_i
    \right]
    \right).
\label{eq:local_evidence}
\end{equation}
We denote the set of local evidence tokens as $\mathbf{Z}^{loc}=\{\mathbf{z}^{loc}_{i}\}_{i=1}^{N}$. The residual term captures the discrepancy between the current Gaussian appearance and the source observation under consistent source-view alignment, while $\gamma_i$ carries the reliability of the observation.
Because a single source image observes only part of the hand, we aggregate reliable source information into a global source appearance identity token $\mathbf{t}^{aid}_{s}$ using masked global and local image features, source and canonical base color statistics, and reliability-weighted local residuals and evidence tokens (Appendix~\ref{sec:supp_architecture}). The token encodes source-level appearance properties, including skin tone, illumination bias, and global color distribution, without describing geometry.
\vspace{-0.1cm}

\textbf{Source-conditioned appearance and color composition.}
We combine three complementary sources of appearance information: local evidence, the Stage-1 appearance prior $\mathbf{Z}^a$, and the global source appearance identity. The Stage-1 appearance prior is updated using local evidence and the global identity (Prior Refinement), while the identity token is expanded into a source-conditioned prior bank for weakly observed regions (Identity Conditioning):
{\small
\begin{equation}
\tilde{\mathbf{Z}}^{a}
=
\mathcal{U}
(
\mathbf{Z}^{a};
\mathbf{Z}^{loc},
\mathbf{t}^{aid}_{s}
),
\qquad
\mathcal{D}_{s}
=
\mathcal{B}
(
\mathbf{t}^{aid}_{s}
),
\quad
\mathbf{h}_i
=
\mathcal{U}_{c}
(
\mathbf{z}^{loc}_{i},
\tilde{\mathbf{z}}^{a}_{i},
\mathbf{t}^{aid}_{s},
\mathbf{c}^{0}_{i},
\mathcal{D}_{s}
).
\label{eq:prior_refine}
\end{equation}
}
This composition enables Gaussians that contribute to rendering to leverage spatially aligned local evidence, while weakly observed or unobserved Gaussians are guided by the base colors and the global source appearance identity.
From the composed appearance token, the module predicts a source-observed color candidate, a learnable appearance gate, and an appearance displacement; the displacements are pooled into a source-level appearance shift (Appearance Prediction):
\begin{equation}
    \left(
    \mathbf{c}^{obs}_{i},
    g_i,
    \boldsymbol{\delta}_{i}
    \right)
    =
    f_{out}
    \left(
    \mathbf{h}_i
    \right),
    \qquad
    g_i
    \in
    [0,1],
    \qquad
    \boldsymbol{\Delta}^{aid}_{s}
    =
    \mathrm{Pool}
    \left(
    \left\{
    \boldsymbol{\delta}_{i}
    \right\}_{i=1}^{N}
    \right).
\label{eq:app_pred}
\end{equation}
The predicted color is not directly assigned to the Gaussian. Instead, the observation reliability modulates the reliance on local source evidence (Appearance Composition):
\begin{equation}
    \mathbf{c}^{*}_{i}
    =
    g_i
    \left[
    \gamma_i
    \mathbf{c}^{obs}_{i}
    +
    \left(
    1-\gamma_i
    \right)
    \mathbf{c}^{0}_{i}
    \right]
    +
    \left(
    1-g_i
    \right)
    \mathbf{c}^{0}_{i}
    +
    \boldsymbol{\Delta}^{aid}_{s}.
    \label{eq:compose}
\end{equation}
Expanding Eq.~\ref{eq:compose} gives $\mathbf{c}^{*}_{i}=\mathbf{c}^{0}_{i}+g_i\gamma_i(\mathbf{c}^{obs}_{i}-\mathbf{c}^{0}_{i})+\boldsymbol{\Delta}^{aid}_{s}$, so $\gamma_i$ and $g_i$ act on the source evidence only through their product; they are nevertheless not interchangeable, since $\gamma_i$ is closed-form and receives no gradient whereas $g_i$ is learned from the evidence (Appendix~\ref{sec:supp_reliability}).
This design separates local evidence transfer from global appearance 
propagation: local evidence enters only through the reliability-weighted term, whereas global appearance cues are consistently propagated across all Gaussians via the source-level shift. The composed colors $\mathbf{c}^{*}_{i}$ replace $\mathbf{c}^{0}_{i}$ in $\mathbf{G}^{can}$, and all non-color attributes are preserved. Novel-pose and novel-view rendering then deform $\mathbf{G}^{can}$, now carrying $\mathbf{c}^{*}_{i}$, with the same MANO-driven deformation and Gaussian renderer:
\begin{equation}
    \mathbf{G}^{target}
    =
    \mathcal{W}
    \left(
    \mathbf{G}^{can}\big|_{\mathbf{c}^{0}_{i}\leftarrow\mathbf{c}^{*}_{i}};
    \boldsymbol{\theta}_{\mathrm{target}},
    \boldsymbol{\beta}
    \right),
    \qquad
    \hat{I}_{\mathrm{target}}
    =
    \mathcal{R}
    \left(
    \mathbf{G}^{target};
    \mathbf{K}_{\mathrm{target}},
    \mathbf{T}_{\mathrm{target}}
    \right).
\end{equation}
Since only the colors of $\mathbf{G}^{can}$ are replaced, the rendering geometry remains governed by the MANO-surface-anchored canonical representation and MANO-driven deformation, while the appearance is determined by the composed colors.
Since the predicted colors are independent of viewing direction and driving pose, the model does not capture view- or pose-dependent effects such as self-shadowing. We discuss this limitation and potential extensions to view-dependent appearance modeling in Appendix~\ref{sec:supp_viewdep}.

\vspace{-0.2cm}

\subsection{Training Objective}
\label{sec:training_objective}
\vspace{-0.1cm}

We train the model in two stages. The first stage learns a stable canonical Gaussian hand representation, while the second stage focuses on source-conditioned appearance modeling with the entire Stage-1 canonical predictor, including the base colors $\mathbf{c}^{0}_i$, kept fixed.

\textbf{Canonical Gaussian training.}
In the first stage, we supervise rendered images using mask-aware photometric and perceptual losses, together with regularization on the canonical Gaussian geometry:
{\small
\begin{equation}
\begin{aligned}
\mathcal{L}_{\mathrm{can}}
&=
\mathcal{L}_{\mathrm{rec}}
+
\lambda_{\mathrm{sil}}
\mathcal{L}_{\mathrm{sil}}
+
\mathcal{L}_{\mathrm{geo}}
+
\mathcal{L}_{\mathrm{band}},
\\
\mathcal{L}_{\mathrm{rec}}
&=
\lambda_{\mathrm{rgb}}
(
\mathcal{L}_{1}
+
\lambda_{\mathrm{lpips}}
\mathcal{L}_{\mathrm{lpips}}
),
\quad
\mathcal{L}_{\mathrm{geo}}
=
\lambda_{\mathrm{off}}
\mathcal{L}_{\mathrm{off}}
+
\lambda_{\mathrm{lap}}
\mathcal{L}_{\mathrm{lap}}
+
\lambda_{\mathrm{scale}}
\mathcal{L}_{\mathrm{scale}} .
\end{aligned}
\label{eq:can_loss}
\end{equation}
}
Here, $\mathcal{L}_{\mathrm{rec}}$ enforces image reconstruction fidelity, $\mathcal{L}_{\mathrm{sil}}$ supervises silhouette consistency, and $\mathcal{L}_{\mathrm{geo}}$ regularizes residual Gaussian displacements, local smoothness, and Gaussian scales.

\begin{table}[t]
\centering
\caption{Quantitative comparison on InterHand2.6M.
Best, second-best, and third-best are marked
\textcolor{ForestGreen}{\textbf{bold green}}, \textcolor{blue}{\underline{underlined blue}}, and
\textcolor{orange}{\textit{italic orange}}.}
\label{tab:main_quant}
\footnotesize
\setlength{\tabcolsep}{4.2pt}
\begin{tabular}{@{}llccc@{}}
  \toprule
  \textbf{Method} & \textbf{Protocol}
  & PSNR$\uparrow$ & SSIM$\uparrow$ & LPIPS$\downarrow$ \\
  \midrule
  HARP~\citep{karunratanakul2023harp}
    & Per-subj.\ opt. & 19.82 & 0.761 & 0.225 \\
  LiveHand~\citep{mundra2023livehand}
    & Per-subj.\ opt. & 23.01 & 0.763 & 0.186 \\
  HandAvatar~\citep{handavatar}
    & Per-subj.\ opt. & 23.79 & \textcolor{orange}{\textit{0.820}} & 0.178 \\
  Handy~\citep{Potamias_2023_CVPR_Handy}
    & Per-subj.\ opt.
    & \textcolor{orange}{\textit{25.56}}
    & 0.794
    & \textcolor{orange}{\textit{0.150}} \\
  OHTA~\citep{zheng2024ohta}
    & Per-subj.\ opt.
    & \textcolor{blue}{\underline{26.11}}
    & \textcolor{blue}{\underline{0.864}}
    & \textcolor{ForestGreen}{\textbf{0.129}} \\
  BIGS$^{*}$~\citep{on2025bigs}
    & Per-subj.\ opt. & 20.46 & 0.749 & 0.212 \\
  \midrule
  \rowcolor{gray!8}
  \textbf{Ours}
    & \textbf{Feed-forward}
    & \textcolor{ForestGreen}{\textbf{26.21}}
    & \textcolor{ForestGreen}{\textbf{0.866}}
    & \textcolor{blue}{\underline{0.133}} \\
  \bottomrule
\end{tabular}
\end{table}

\begin{table}[t]
\centering
\caption{
Single-image reconstruction and efficiency on HanCo.
Time is the avatar-acquisition cost; FPS and memory are measured for $256\times256$ rendering (protocol in Appendix~\ref{sec:supp_timing}). Best in \textbf{bold}.
}
\label{tab:hanco_results}
\footnotesize
\setlength{\tabcolsep}{3.2pt}
\begin{tabular}{@{}lccccccc@{}}
\toprule
Method
&
Protocol
&
\multicolumn{3}{c}{Reconstruction}
&
\multicolumn{3}{c}{Efficiency}
\\
\cmidrule(lr){3-5}
\cmidrule(lr){6-8}
&
&
PSNR$\uparrow$
&
SSIM$\uparrow$
&
LPIPS$\downarrow$
&
Time (s)$\downarrow$
&
FPS$\uparrow$
&
Mem. (GB)$\downarrow$
\\
\midrule
HandAvatar
&
Per-subj. opt.
&
\textbf{20.32}
&
0.841
&
0.198
&
5579
&
0.32
&
14.80
\\
OHTA
&
Per-subj. opt.
&
20.06
&
\textbf{0.843}
&
\textbf{0.186}
&
3360
&
0.26
&
47.80
\\
BIGS$^{*}$
&
Per-subj. opt.
&
17.71
&
0.731
&
0.246
&
1380
&
43.11
&
\textbf{1.50}
\\
\midrule
\rowcolor{gray!8}
\textbf{Ours}
&
\textbf{Zero-shot feed-forward}
&
20.18
&
0.836
&
0.188
&
\textbf{2.84}
&
\textbf{160.71}
&
1.79
\\
\bottomrule
\end{tabular}
\end{table}

\textbf{Soft Alpha Band regularization.}
We design a task-specific Soft Alpha Band regularizer $\mathcal{L}_{\mathrm{band}}$ to stabilize thin articulated fingers. Given the ground-truth silhouette $M$ and the rendered alpha map $A$, we define an outer band $B_{\mathrm{out}}$ via dilation and an inner region $B_{\mathrm{in}}$ via erosion. The regularizer encourages compact yet sufficient rendered-alpha support by penalizing alpha leakage near boundaries and insufficient alpha within the silhouette interior:
{\small
\setlength{\thinmuskip}{2mu}
\setlength{\medmuskip}{2mu}
\setlength{\thickmuskip}{2mu}

\begin{equation}
\mathcal{L}_{\mathrm{band}}
=
\eta(n)
(\lambda_{\mathrm{out}}\mathcal{L}_{\mathrm{out}}
+
\lambda_{\mathrm{in}}\mathcal{L}_{\mathrm{in}}),
\quad
\mathcal{L}_{\mathrm{out}}
=
\mathrm{Avg}_{B_{\mathrm{out}}}
\!\left[
\psi\!\left(
[A-\tau_{\mathrm{out}}]_{+}
\right)
\right],
\quad
\mathcal{L}_{\mathrm{in}}
=
\mathrm{Avg}_{B_{\mathrm{in}}}
\!\left[
\psi\!\left(
[\tau_{\mathrm{in}}-A]_{+}
\right)
\right].
\label{eq:sab}
\end{equation}
}
Here, $\psi(\cdot)$ denotes a robust Huber penalty and $\eta(n)$ is a weighting coefficient scheduled over the optimization step $n$.
The outer-band term suppresses alpha spill around finger boundaries, reducing thickened geometry artifacts, while the inner-region term maintains sufficient coverage inside the hand silhouette.

\textbf{Source-conditioned appearance training.}
In the second stage, we freeze the entire Stage-1 canonical predictor, including the base-color head $f_{\mathrm{rgb}}$, and train only the source-conditioned appearance module. The objective combines source-view reconstruction, feature-statistic alignment, and novel-view supervision:
{\small
\begin{equation}
\mathcal{L}_{\mathrm{app}}
=
\lambda_{s1}
\mathcal{L}_{1}^{\mathrm{src}}
+
\lambda_{sp}
\mathcal{L}_{\mathrm{lpips}}^{\mathrm{src}}
+
\lambda_{\mathrm{stat}}
\mathcal{L}_{\mathrm{stat}}
+
\lambda_{n1}
\mathcal{L}_{1}^{\mathrm{novel}}
+
\lambda_{np}
\mathcal{L}_{\mathrm{lpips}}^{\mathrm{novel}} .
\label{eq:app_loss}
\end{equation}
}
The source-view losses enforce consistency between the rendered output and the input image; $\mathcal{L}_{\mathrm{stat}}$ aligns foreground feature statistics between the source image and the rendered source view, and the novel-view losses maintain appearance consistency under MANO-driven deformation. 
All loss terms are defined in full and all loss weights of both stages are reported in Appendices~\ref{sec:supp_training} and~\ref{sec:supp_losses}.

\vspace{-0.2cm}

\section{Experiments}

\textbf{Training Data and Preprocessing.}
We train on InterHand2.6M~\citep{Moon_2020_ECCV_InterHand2.6M}, HO3Dv3~\citep{hampali2020ho3d,2021ho3dv3}, Goliath~\citep{martinez2024Goliath}, Hand Appearance~\citep{karunratanakul2023harp}, and PALM~\citep{fan2026palm}; each sample pairs a source image with multiple target frames of the same subject. For the appearance stage we add synthetic source--target pairs generated following \citet{Zhao_2025_CVPR} to widen source appearance variation. Images are cropped to hand-centered $256\times256$ regions with intrinsics updated accordingly; where annotations are unavailable, masks and MANO/camera parameters are estimated with SAM 3~\citep{carion2026sam3} and WiLoR~\citep{potamias2025wilor} (details in Appendix~\ref{sec:supp_dataset}).

\vspace{-0.2cm}
\subsection{Comparisons with Prior Methods}
\vspace{-0.2cm}

\textbf{Evaluation Protocol and Metrics.}
We evaluate on InterHand2.6M~\citep{Moon_2020_ECCV_InterHand2.6M} under the single-image reconstruction protocol of prior hand avatar benchmarks~\citep{handavatar,zheng2024ohta}: each method reconstructs a subject-specific hand from one source image and renders novel poses and views. Zero-shot generalization is evaluated on HanCo~\citep{zimmermann2021HanCo} without HanCo training or test-time optimization. All methods share the same source frame and identical target poses and views; we report PSNR, SSIM, and LPIPS over hand-centered regions defined by bounding-box ray masks. The complete protocol, per-baseline input protocol, and provenance of all baseline numbers are given in Appendix~\ref{sec:supp_eval}.

{\textbf{Quantitative Comparisons.}
Table~\ref{tab:main_quant} reports results on InterHand2.6M: without any test-time fitting, InstantHand matches the strongest per-subject-optimized baseline (26.21 vs.\ 26.11\,dB PSNR, 0.866 vs.\ 0.864 SSIM) and attains the second-best LPIPS. Table~\ref{tab:hanco_results} presents zero-shot cross-dataset evaluation and efficiency on HanCo, where the baselines are fitted to each test subject and our model is applied without HanCo training subjects or optimization; InstantHand stays within 0.14\,dB PSNR, 0.007 SSIM, and 0.002 LPIPS of the best fitted baselines at over 400$\times$ lower avatar-acquisition cost. BIGS$^{*}$ denotes our single-hand adaptation of BIGS~\citep{on2025bigs} (Appendix~\ref{sec:supp_bigs}).}

\begin{figure*}[t]
  \centering
  \includegraphics[width=\textwidth]{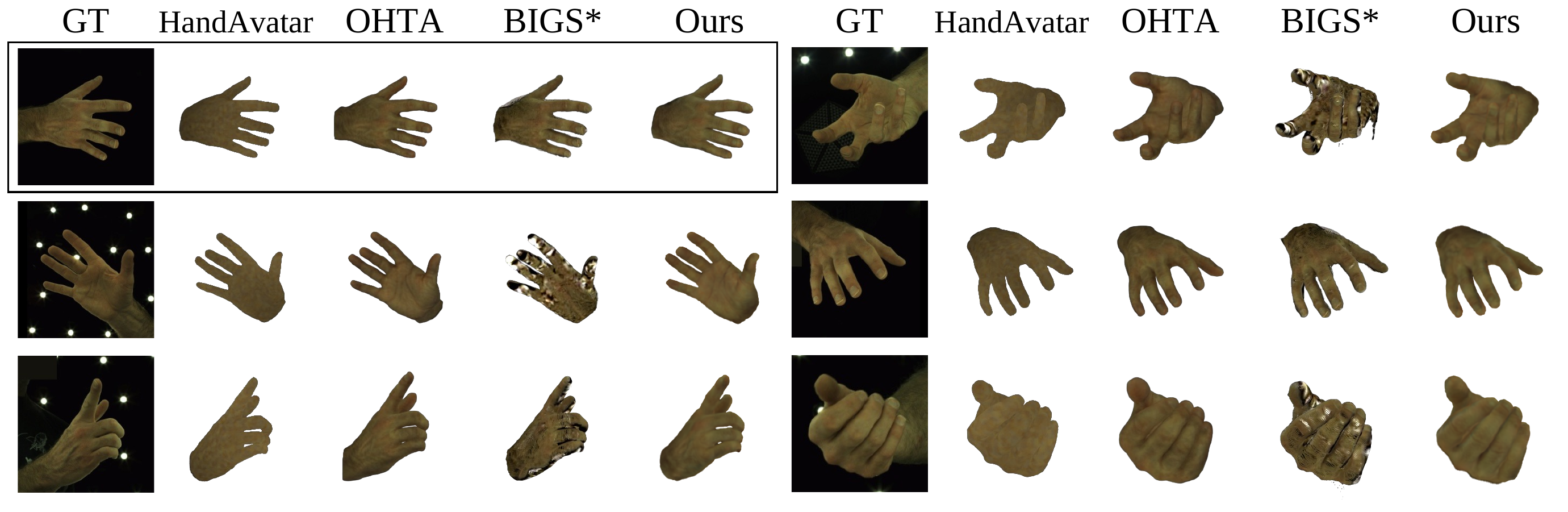}\\[3pt]
  \includegraphics[width=\textwidth]{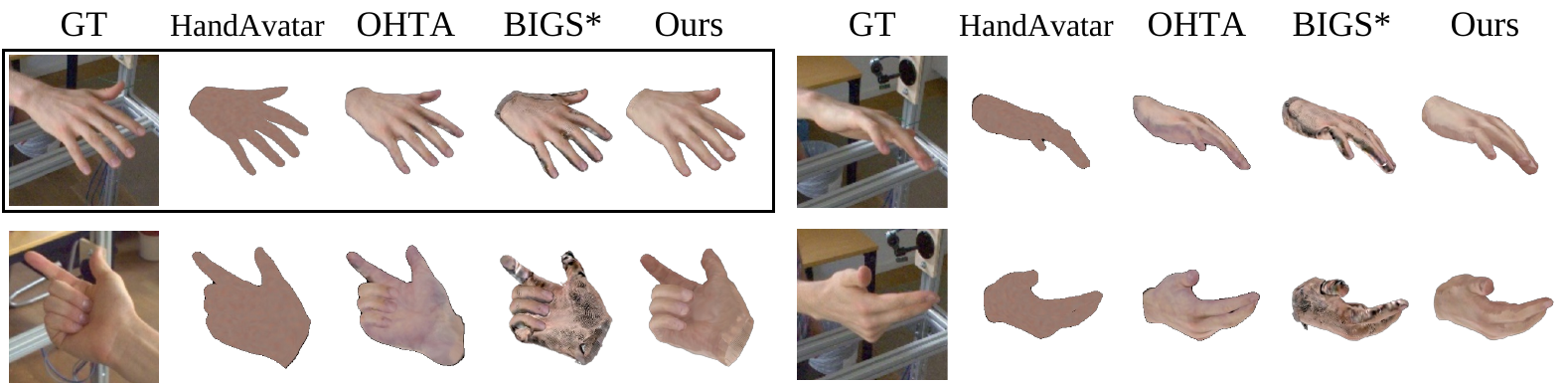}
  \caption{Qualitative comparison with prior methods on InterHand2.6M~\citep{Moon_2020_ECCV_InterHand2.6M} (top three rows) and HanCo~\citep{zimmermann2021HanCo} (bottom two rows). Columns: GT, HandAvatar, OHTA, BIGS$^{*}$, Ours; the black box marks the input image.}
  \label{fig:qual}
  \vspace{-0.3cm}
\end{figure*}

\begin{figure}[t]
  \centering
  \includegraphics{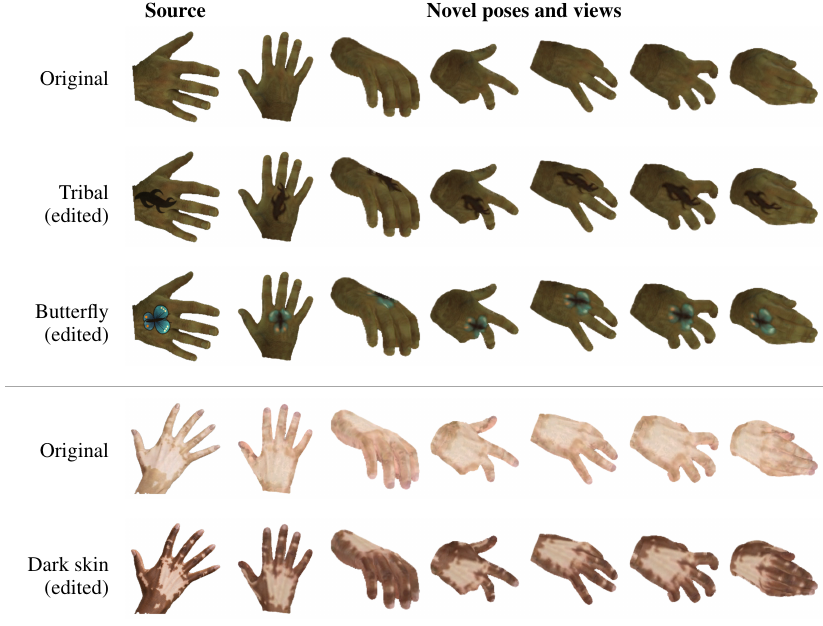}
  \caption{Avatars reconstructed from fully estimated inputs (SAM~3 mask, WiLoR MANO and camera) for original
  and edited source images. Top: an unseen InterHand2.6M test subject, with hand-drawn tattoos
  composited onto the source image. Bottom: an in-the-wild image released with OHTA~\citep{zheng2024ohta},
  with its skin tone darkened by an image generation model. Edited rows reuse the mask, MANO, and camera
  of the original row, and each column shows the same pose and view.}
  \label{fig:edit}
\end{figure}

\textbf{Qualitative Comparisons.}
Figure~\ref{fig:qual} presents qualitative comparisons on InterHand2.6M and HanCo. Under the single-image setting, the fitted baselines exhibit over-smoothed geometry, distorted fingers, or texture artifacts. InstantHand retains coherent hand geometry, well-defined finger boundaries, and fine details such as wrinkles and skin texture. On HanCo, these results are obtained zero-shot, whereas the baselines are fitted to each subject. Figure~\ref{fig:edit} applies InstantHand to fully estimated inputs and to edited versions of the same source images. Because each edited image shares the mask, MANO, and camera of its original, comparing their renderings shows how the input appearance is carried to the avatar. Both the local patterns (top) and the global skin tone (bottom) follow the hand through novel poses and views. Figure~\ref{fig:supp_animation} in the appendix shows novel-pose animations of eight unseen subjects, including the in-the-wild input of Figure~\ref{fig:edit}. The supplementary video (Appendix~\ref{sec:supp_videos}) further includes $360^\circ$ rotations and continuous novel-pose animations.

\subsection{Ablation Study}

\textbf{Component ablation.}
Table~\ref{tab:ablation} reports ablations on InterHand2.6M. Removing the Soft Alpha Band regularizer reduces reconstruction quality for thin structures. Reliability gating ($\gamma$) and the learned appearance gate ($g_i$) play complementary roles: $\gamma$ weights a Gaussian according to its source-view contribution and foreground support, while $g_i$ predicts how strongly to use the sampled appearance evidence. Removing either produces a substantial drop. Although they appear only as a product in Eq.~\ref{eq:compose}, retraining each variant does not allow the remaining gate to compensate for the removed one. Disabling appearance transfer entirely (the Stage-1 model) yields 26.01\,dB, outperforming both single-gate variants; thus, appearance transfer is beneficial only when both gates are present. Synthetic data provides a modest additional gain. 
\begin{table}[t]
\centering
\caption{Ablation and source-cue robustness on InterHand2.6M.
Ablation rows remove one component and retrain both stages. Estimated-cue rows keep the trained full model and replace its source inputs with estimates.}
\label{tab:ablation}
\small
\setlength{\tabcolsep}{13pt}
\begin{tabular}{@{}lccc@{}}
  \toprule
  \textbf{Setting} & PSNR$\uparrow$ & SSIM$\uparrow$ & LPIPS$\downarrow$ \\
  \midrule
  w/o SAB ($\mathcal{L}_{\mathrm{band}}$) & 23.89 & 0.825 & 0.166 \\
  w/o reliability gating ($\gamma$)       & 23.18 & 0.829 & 0.171 \\
  w/o appearance gate ($g_i$)             & 24.52 & 0.833 & 0.152 \\
  w/o appearance transfer                 & 26.01 & 0.865 & 0.134 \\
  w/o synthetic data                      & 26.13 & 0.864 & 0.134 \\
  \midrule
  \textbf{Ours} (GT cues)       & \textbf{26.21} & \textbf{0.866} & \textbf{0.133} \\
  \midrule
  \multicolumn{4}{@{}l}{\textit{Estimated source cues}} \\
  SAM 3 mask       & 26.12 & 0.865 & 0.133 \\
  WiLoR MANO      & 26.07 & 0.863 & 0.134 \\
  Fully estimated & 25.99 & 0.863 & 0.134 \\
  \bottomrule
\end{tabular}
\end{table}
Further analyses are provided in Appendices~\ref{sec:supp_ablation} and~\ref{sec:supp_qualitative}.

\textbf{Robustness to source-cue errors.}
Table~\ref{tab:ablation} (bottom) evaluates sensitivity to estimated source inputs. We replace the ground-truth source mask with a SAM 3 estimate~\citep{carion2026sam3}, and the source MANO parameters and camera with WiLoR estimates~\citep{potamias2025wilor}, keeping the trained model and evaluation protocol fixed. Target poses and cameras remain ground truth in every setting. When all source inputs are estimated from a single RGB image, PSNR decreases by only 0.22 dB, while SSIM and LPIPS remain essentially unchanged. The decrease arises from the mask and MANO estimates; replacing the source camera with its estimate causes no further PSNR drop. Controlled input perturbations reduce PSNR by at most 0.16 dB. Appendix~\ref{sec:supp_source_sensitivity} provides the per-cue breakdown (Table~\ref{tab:supp_sensitivity}) and perturbation protocol, while Appendix~\ref{sec:supp_freihand} reports results on unseen FreiHAND images with fully estimated inputs.

\textbf{Source-fidelity diagnostic.}
Aggregate novel-pose metrics combine source-observed and unobserved regions, obscuring the fidelity of source appearance reconstruction. To isolate the effect of the appearance module, we render each avatar in its source view while keeping all non-color Gaussian attributes fixed at their canonical values, so metric differences reflect only the color update. With each evaluation frame serving in turn as the source, the module improves source-view PSNR by 1.96 dB on InterHand2.6M and 3.41 dB on zero-shot HanCo (Table~\ref{tab:supp_selfrecon}; protocol in Appendix~\ref{sec:supp_selfrecon_protocol}).

\begin{table}[t]
\centering
\caption{
Controlled comparison with feed-forward avatar recipes on InterHand2.6M.
Each recipe swaps its template (FLAME, SMPL-X) for MANO and keeps its published parameterization under otherwise matched training (Appendix~\ref{sec:supp_recipe_ports}); boundary metrics, distinct from the training bands, follow Appendix~\ref{sec:supp_boundary_metrics}.
}
\label{tab:recipe_port}
\small
\setlength{\tabcolsep}{4.0pt}
\begin{tabular}{@{}lcccccc@{}}
\toprule
Configuration & PSNR$\uparrow$ & SSIM$\uparrow$ & LPIPS$\downarrow$ & Bound.~F$\uparrow$ & Leakage$\downarrow$ & Int.~cov.$\uparrow$ \\
\midrule
\rowcolor{gray!8}
\textbf{Ours (constrained + SAB)} & \textbf{26.21} & \textbf{0.866} & \textbf{0.133} & \textbf{0.698} & \textbf{0.234} & 0.999 \\
Constrained, w/o SAB & 23.89 & 0.825 & 0.166 & 0.366 & 0.555 & 1.000 \\
Head-recipe port~\citep{he2025lam} & 23.60 & 0.809 & 0.182 & 0.299 & 0.608 & 1.000 \\
Body-recipe port~\citep{qiu2025LHM} & 23.36 & 0.805 & 0.182 & 0.267 & 0.627 & 1.000 \\
\bottomrule
\end{tabular}
\end{table}

\textbf{Comparison with feed-forward avatar recipes.}
Table~\ref{tab:recipe_port} examines whether avatar designs developed for heads and bodies transfer to hands under matched conditions. Compared with InstantHand, the head-recipe port loses 2.61 dB, reduces the boundary F-score from 0.698 to 0.299, and increases alpha leakage from 0.234 to 0.608, while interior coverage remains at 1.000. This pattern localizes the failure to thin, articulated contours rather than the hand interior, consistent with Sec.~\ref{sec:intro}. The head-recipe port also has a lower boundary F-score and greater leakage than the constrained model without SAB, indicating that both the constrained parameterization and boundary-band supervision contribute to boundary quality. Even with boundary supervision retained, freeing per-axis scale, rotation and opacity jointly, or all three reduces PSNR by 0.47--0.71 dB (Appendix~\ref{sec:supp_attr_ablation}). The body-recipe port performs similarly to the head-recipe port, within 0.24 dB in PSNR and with a boundary F-score of 0.267 versus 0.299, despite its canonical-space regularizers.

\textbf{Pose and temporal diagnostics.} Grouping test frames by pose change relative to the source reveals that quality degrades primarily with finger articulation: PSNR drops by 2.69 dB on InterHand2.6M and 1.93 dB on HanCo between the mildest and strongest articulation thirds. Changes in viewing direction have a smaller effect, with drops of at most 0.97 dB. Under continuous pose driving on zero-shot HanCo, frame-to-frame changes in our renderings exceed those in the ground truth by 4.7\% on average and at most 8.0\% per camera. All views and frames are rendered from a single fixed set of canonical Gaussians, providing a shared representation across views; we assess cross-view and temporal consistency empirically through these diagnostics and the supplementary video. Appendix~\ref{sec:supp_analyses} provides the full analyses and comparisons with baselines evaluated on the same frames.

\section{Conclusion}

We presented \emph{InstantHand}, a feed-forward framework for reconstructing an animatable 3D Gaussian hand avatar from a single image without per-subject optimization. A MANO-anchored canonical representation, together with the Soft Alpha Band regularizer, preserves compact, thin articulated structures, while Reliability-gated Appearance Transfer integrates observed source appearance according to each Gaussian's rendering contribution and completes unobserved regions using the learned prior and a global source identity. Once constructed, the avatar renders at 161 FPS. Its Gaussian colors and scales remain fixed across driving poses, preserving appearance consistency. InstantHand constructs avatars over 400$\times$ faster than optimization-based baselines while maintaining competitive reconstruction quality in zero-shot evaluation on HanCo and under fully estimated inputs.

\begin{ack}
This work is supported by NRF grants (No. RS-2025-00521013 30\%, No. RS-2025-02216916 10\%) and IITP grants (No. RS-2020-II201336 Artificial intelligence graduate school program (UNIST) 10\%; No. RS-2025-25442824 AI Star Fellowship Program (UNIST) 10\%; No. RS-2022-II220264 Comprehensive video understanding and generation with knowledge-based deep logic neural network 10\%; No. RS-2025-25442149 LG AI STAR Talent Development Program for Leading Large-Scale Generative AI Models in the Physical AI Domain 10\%), funded by the Korean government (MSIT). This work is also supported by the InnoCORE program of the Ministry of Science and ICT (26-InnoCORE-01 10\%, UNIST 1.260006.01 10\%).
\end{ack}

\bibliographystyle{unsrtnat}
\bibliography{main}

\clearpage
\appendix

\setcounter{figure}{0}
\setcounter{table}{0}
\setcounter{equation}{0}
\renewcommand{\thefigure}{S\arabic{figure}}
\renewcommand{\thetable}{S\arabic{table}}
\renewcommand{\theequation}{S\arabic{equation}}

\raggedbottom
\setcounter{topnumber}{5}
\setcounter{bottomnumber}{5}
\setcounter{totalnumber}{10}
\renewcommand{\topfraction}{0.95}
\renewcommand{\bottomfraction}{0.90}
\renewcommand{\textfraction}{0.05}
\renewcommand{\floatpagefraction}{0.80}
\setlength{\textfloatsep}{6pt plus 2pt minus 2pt}
\setlength{\floatsep}{6pt plus 2pt minus 2pt}
\setlength{\intextsep}{6pt plus 2pt minus 2pt}

\section*{Appendix}

This appendix gives the full implementation, data, and evaluation details of InstantHand.
Secs.~\ref{sec:supp_impl} and~\ref{sec:supp_dataset} describe the model and the training data, and
Sec.~\ref{sec:supp_eval} specifies every evaluation protocol used in the main paper and in this appendix.
Sec.~\ref{sec:supp_related} extends the related work, Sec.~\ref{sec:supp_baseline} describes the baseline
adaptations, and Sec.~\ref{sec:supp_ablation} reports additional ablations.
Secs.~\ref{sec:supp_discussion}--\ref{sec:supp_limitations} contain the discussion (including the ethical
considerations in Sec.~\ref{sec:supp_ethics}), additional qualitative results, and the failure cases and
limitations.
Sec.~\ref{sec:supp_analyses} analyzes the results by pose difficulty and over time and examines the
reliability and opacity of the canonical avatar.
Sec.~\ref{sec:supp_videos} describes the supplementary video, and Sec.~\ref{sec:supp_reproducibility}
summarizes reproducibility.
For a quick overview, we recommend the supplementary video (Sec.~\ref{sec:supp_videos}), which shows
$360^\circ$ rotations and continuous novel-pose driving, together with the novel-pose animation of unseen
subjects from fully estimated inputs (Fig.~\ref{fig:supp_animation}) and the failure cases
(Fig.~\ref{fig:failure_cases}).

\section{Implementation Details}
\label{sec:supp_impl}

\subsection{Training Protocol}
\label{sec:supp_training}

We train InstantHand in two stages, canonical representation learning followed by source-conditioned
Gaussian appearance modeling.
Both stages use AdamW ($\beta_1=0.9$, $\beta_2=0.95$) with a learning-rate schedule of 2{,}000 warmup
steps, 10{,}000 hold steps, and 8{,}000 cosine-decay steps, gradient clipping at norm $1.0$, and bfloat16
mixed precision.
Training is distributed over four NVIDIA RTX A6000 GPUs.

\paragraph{Stage 1: canonical representation learning.}
We train the MANO-anchored canonical Gaussian representation for 10 epochs of 2{,}000 optimization steps
each (20{,}000 steps in total), with a per-GPU batch size of 6, gradient accumulation over 2 steps, a learning rate
of $3\times10^{-4}$ that decays to $0.1\times$ its peak, and a weight decay of $0.05$.
This stage takes about 25 hours.
The objective of Eq.~\ref{eq:can_loss} of the main paper combines mask-aware RGB reconstruction, LPIPS,
silhouette supervision, Laplacian, offset, and scale regularization, and the Soft Alpha Band (SAB)
regularizer of Eq.~\ref{eq:sab}, with weights $\lambda_{\mathrm{rgb}}=1.0$, $\lambda_{\mathrm{lpips}}=0.2$,
$\lambda_{\mathrm{sil}}=0.6$, $\lambda_{\mathrm{lap}}=10{,}000$, $\lambda_{\mathrm{off}}=0.1$, and
$\lambda_{\mathrm{scale}}=0.75$, and with $\lambda_{\mathrm{out}}=0.25$ and $\lambda_{\mathrm{in}}=0.05$ for
SAB.

\paragraph{Stage 2: source-conditioned appearance modeling.}
We initialize from the Stage-1 checkpoint and \emph{freeze the entire Stage-1 canonical predictor},
including the base-color head $f_{\mathrm{rgb}}$.
The canonical geometry and base colors are therefore fixed, and only the source-conditioned appearance
module of Sec.~\ref{sec:source_conditioned_gaussian_appearance} of the main paper is optimized.
We train for 10 epochs of 2{,}000 steps each with a per-GPU batch size of 4, a learning rate of $1\times10^{-4}$
that decays to $0.2\times$ its peak, and a weight decay of $0.01$.
This stage takes about 13 hours.
The objective of Eq.~\ref{eq:app_loss} of the main paper combines source-view reconstruction, novel-view
reconstruction, and feature-statistic consistency, with $\lambda_{s1}=1.0$, $\lambda_{sp}=0.5$,
$\lambda_{n1}=1.0$, $\lambda_{np}=0.25$, and $\lambda_{\mathrm{stat}}=0.05$.
All loss terms are defined in Sec.~\ref{sec:supp_losses}.

\paragraph{Data preparation.}
Each training sample consists of one source image and six target frames of the same subject, which
provides supervision under varying poses and viewpoints.
All images are cropped around the hand using the bounding box of the 2D keypoints, enlarged by a factor of
1.3, and resized to $256\times256$, and the camera intrinsics are adjusted accordingly.
When available, we use the provided masks, MANO parameters, and camera annotations.
Otherwise, we estimate hand masks with SAM 3~\citep{carion2026sam3} and MANO and camera parameters with
WiLoR~\citep{potamias2025wilor}.
Left-hand samples are flipped horizontally to keep a consistent canonical hand orientation.
The same flip is applied during training and inference, so a single network serves both hands.
Dataset construction and sampling are detailed in Sec.~\ref{sec:supp_dataset}.

\paragraph{Canonical Gaussian parameterization values.}
The fixed and bounded attributes of $\mathbf{G}^{can}$ in Eqs.~\ref{eq:gcan} and~\ref{eq:decoder} of the
main paper take the following values.
The opacity is fixed at $\alpha_0=1$, and the canonical rotation is fixed to the identity quaternion
$\mathbf{r}_0=(1,0,0,0)$.
The scale activation $\rho$ is a clamped exponential,
$s_i=\rho\bigl(f_s(\mathbf{z}^g_i)\bigr)=\mathrm{clamp}\!\left(\exp\!\left(f_s(\mathbf{z}^g_i)\right),\,5\times10^{-4},\,s_{\max}\right)$
in canonical MANO units, with $s_{\max}=0.008$.
The offset head $f_\Delta$ ends in the per-axis bounded activation $0.008\,(\mathrm{sigmoid}(\cdot)-0.5)$
on the output of its final linear layer.
This bound keeps each center within $\pm4$\,mm of its anchor, well below the finger width, so the offsets
can refine the local surface but cannot move a center across a finger boundary.
Both bounds lie below the finger scale, and the boundary leakage that remains even under these constraints
is what the Soft Alpha Band targets (Sec.~\ref{sec:supp_sab}).
Sec.~\ref{sec:supp_attr_ablation} quantifies the cost of relaxing each of these constraints.

\subsection{Loss Definitions}
\label{sec:supp_losses}

For completeness, we define every term of Eqs.~\ref{eq:can_loss} and~\ref{eq:app_loss} of the main paper.
Throughout, $\hat{I}$ denotes a render and $I$ the corresponding ground-truth image.
All losses are computed at $256\times256$ resolution, and the reconstruction terms are restricted to the
foreground region given by the provided mask.

\paragraph{Stage-1 terms (Eq.~\ref{eq:can_loss}).}
$\mathcal{L}_{1}$ is a mask-aware $L_1$ reconstruction term on the
target views, $\mathcal{L}_{\mathrm{lpips}}$ the perceptual loss on the same
renders, and $\mathcal{L}_{\mathrm{sil}}$ an $L_1$ silhouette term on the rendered
alpha against the ground-truth mask.
The three geometry regularizers of $\mathcal{L}_{\mathrm{geo}}$ are
\begin{equation}
\mathcal{L}_{\mathrm{lap}}
=\frac{1}{N}\sum_{i=1}^{N}
\Bigl\lVert \Delta\mathbf{x}_i
-\tfrac{1}{\deg(i)}\!\sum_{j\sim i}\!\Delta\mathbf{x}_j
\Bigr\rVert_2^{2},
\qquad
\mathcal{L}_{\mathrm{off}}
=\frac{1}{3N}\sum_{i=1}^{N}\lVert\Delta\mathbf{x}_i\rVert_2^{2},
\label{eq:supp_lap_off}
\end{equation}
\begin{equation}
\mathcal{L}_{\mathrm{scale}}
=\frac{1}{N}\sum_{i=1}^{N}\bigl[\,s_i-s_{\mathrm{th}}\,\bigr]_{+}^{2},
\label{eq:supp_scale}
\end{equation}
where $j\sim i$ ranges over the $\deg(i)$ one-ring neighbors of anchor $i$ on the subdivided canonical
mesh and $[\cdot]_{+}=\max(\cdot,0)$.
$\mathcal{L}_{\mathrm{lap}}$ penalizes the uniform graph Laplacian of the offset field,
$\mathcal{L}_{\mathrm{off}}$ keeps the offsets small, and $\mathcal{L}_{\mathrm{scale}}$ penalizes only
scales above the threshold $s_{\mathrm{th}}=0.004$ in canonical MANO units, which is half the hard ceiling
$s_{\max}=0.008$ of Sec.~\ref{sec:supp_training}.
$\mathcal{L}_{\mathrm{band}}$ is defined in Sec.~\ref{sec:supp_sab}.

\paragraph{Stage-2 reconstruction terms (Eq.~\ref{eq:app_loss}).}
The source-view terms
$\lambda_{s1}\mathcal{L}_{1}^{\mathrm{src}}+\lambda_{sp}\mathcal{L}_{\mathrm{lpips}}^{\mathrm{src}}$
are computed between the rendered source view and the source image, and the
novel-view terms
$\lambda_{n1}\mathcal{L}_{1}^{\mathrm{novel}}+\lambda_{np}\mathcal{L}_{\mathrm{lpips}}^{\mathrm{novel}}$
apply the same masked $L_1$ and LPIPS losses to novel-view renderings against
their ground-truth target frames.

\paragraph{Feature-statistic consistency $\mathcal{L}_{\mathrm{stat}}$.}
This term matches foreground-masked statistics of features from the frozen DINOv3 encoder, the same
encoder that conditions our model.
Let $F_\ell(\cdot)$ denote channel $\ell$ of the DINOv3 feature map of an image, $N_{\mathrm{ch}}$ the
number of feature channels, and $M$ the ground-truth hand mask downsampled to the feature resolution.
The per-channel masked mean and standard deviation are
\begin{equation}
\mu_\ell(X;M)\;=\;\frac{\sum_{\mathbf{p}} M(\mathbf{p})\,F_\ell(X)(\mathbf{p})}{\sum_{\mathbf{p}} M(\mathbf{p})},
\label{eq:supp_mu}
\end{equation}
\begin{equation}
\sigma_\ell(X;M)\;=\;
\sqrt{\frac{\sum_{\mathbf{p}} M(\mathbf{p})\,\bigl(F_\ell(X)(\mathbf{p})-\mu_\ell(X;M)\bigr)^{2}}
{\sum_{\mathbf{p}} M(\mathbf{p})}}.
\label{eq:supp_sigma}
\end{equation}
The loss is the mean absolute difference of the two statistics over channels,
\begin{equation}
\mathcal{L}_{\mathrm{stat}}(X,Y)=
\frac{1}{N_{\mathrm{ch}}}\sum_{\ell}\bigl|\mu_\ell(X;M_X)-\mu_\ell(Y;M_Y)\bigr|
+\frac{1}{N_{\mathrm{ch}}}\sum_{\ell}\bigl|\sigma_\ell(X;M_X)-\sigma_\ell(Y;M_Y)\bigr|.
\label{eq:supp_stat}
\end{equation}
In Stage 2, it is applied between the rendered source view and the source image with
$\lambda_{\mathrm{stat}}=0.05$.
The source-image features are detached, so gradients flow only through the render.
The term regularizes the global foreground appearance, such as tone and contrast, without requiring pixel
alignment.

\subsection{Soft Alpha Band Details}
\label{sec:supp_sab}

The bands $B_{\mathrm{out}}$ and $B_{\mathrm{in}}$ of Eq.~\ref{eq:sab} of the main paper are constructed
from the ground-truth silhouette $M$ by morphological dilation and erosion.
$B_{\mathrm{out}}$ is the 2-pixel dilation band outside $M$, and $B_{\mathrm{in}}$ is the interior eroded
by 2 pixels, with all widths defined at $256\times256$ resolution.
We use the tolerances $\tau_{\mathrm{out}}=0.03$ and $\tau_{\mathrm{in}}=0.60$ and smooth-$L_1$ (Huber)
penalties $\psi(\cdot)$ whose quadratic-to-linear transition lies at $0.03$ for the outer term and at
$0.05$ for the inner term.
The tolerances are deliberately asymmetric.
Alpha in the outer band is penalized above a near-zero level, whereas the interior is only required to
stay above $0.60$, because leakage is the failure mode being targeted while interior coverage only needs a
floor.

The schedule $\eta(n)$ over the optimization step $n$ is zero for the first 5{,}000 steps and then
increases linearly to $1$ between steps 5{,}000 and 12{,}000 (25\% and 60\% of the 20{,}000 Stage-1
steps).
The band penalty is thus activated only after a coarse geometry has formed.
Applying it from initialization destabilizes early optimization, when the rendered silhouette is still far
from the target.

Both bands are constructed per frame from the ground-truth silhouette.
Under crossed fingers or near-contact poses, every region inside $M$, including contact regions between
fingers, is therefore untouched by the outer penalty, which acts only on true background.
Since hands are opaque, no valid translucency is suppressed, and the interior floor $\tau_{\mathrm{in}}$
prevents erosion near contact boundaries.
Figure~\ref{fig:supp_sab_qual} (Sec.~\ref{sec:supp_sab_qual}) shows the qualitative effect of SAB.

\subsection{Observation Reliability Details}
\label{sec:supp_reliability}

The observation reliability of Eq.~\ref{eq:gamma} of the main paper fuses each Gaussian's
accumulated rendering contribution $C_i$ and foreground support $m_i$ into a
confidence $\gamma_i\in[0,1]$ through a square-root transform, median-relative
thresholding, and percentile normalization.

\paragraph{Design requirements.}
The reliability is designed to satisfy four requirements.
(i)~$\gamma_i$ should increase monotonically with the evidence a Gaussian actually contributes to source
pixels.
(ii)~It should be insensitive to changes in projected hand size, camera distance, and image scale.
(iii)~It should be robust to the heavy-tailed distribution of rendering contributions.
(iv)~It should be bounded in $[0,1]$, so that the reliability-weighted term in Eq.~\ref{eq:compose} of the
main paper interpolates convexly between observed and base colors.
Each component of Eq.~\ref{eq:gamma} addresses one of these requirements with a standard
robust-statistics tool.

\emph{Square-root transform.} The distribution of $\{C_i\}$ is heavy-tailed, as a small number of
Gaussians near the silhouette and on prominent surface regions accumulate disproportionately large
contributions.
Using $C_i$ directly would let these few Gaussians dominate the normalization and push the confidence of
moderately observed Gaussians toward zero.
The square root, the classical variance-stabilizing transform for nonnegative count-like mass, compresses
this tail (requirement iii).
As a result, $\gamma_i$ reflects \emph{whether} a Gaussian is reliably observed rather than \emph{how
strongly} it dominates rasterization.

\emph{Median-relative thresholding.} Thresholding contributions relative to the median of the nonzero
contributions, rather than with an absolute cutoff, adapts to each source image and is therefore
insensitive to the global scale of the rendering contributions (requirement ii).
Absolute thresholds are brittle because the total contribution mass varies with hand pose, projected hand
size, and camera distance.
A fixed sigmoid or softmax over $C_i$ would instead require a temperature, which is not scale-invariant
across images.

\emph{Percentile normalization.} Normalizing $\sqrt{C_i}\,m_i$ by its $\zeta$-th percentile over
$\mathcal{O}$ instead of its maximum amounts to quantile normalization with winsorization at the
$\zeta$-th percentile (requirements iii and iv).
It is robust to outliers that survive the variance-stabilizing transform, and clipping to $[0,1]$ gives a
bounded confidence that is comparable across poses and image scales.

\paragraph{Hyperparameters.}
With $(\kappa,\tau_m,\zeta)=(0.5,0.5,95)$, a Gaussian enters $\mathcal{O}$ if its rendering
contribution exceeds half the median nonzero contribution and more than half of its projected footprint
lies on the hand, and the scores are normalized by the 95th percentile.
Both thresholds are set to the midpoint $0.5$, and the 95th percentile is a conventional choice for robust
normalization.
Sec.~\ref{sec:supp_gamma_sensitivity} reports the sensitivity around these values.

\paragraph{Graded weights and the observed-set boundary.}
Unlike a binary visibility mask, $\gamma$ assigns \emph{graded} weights within
$\mathcal{O}$ and is zero outside it.
Gaussians at the edge of $\mathcal{O}$ contribute only weakly to the source image, so the transition from
source-driven to prior-driven color takes place where the source evidence is weakest.
Alpha blending across many overlapping Gaussians with different $\gamma$ values usually smooths this
transition in the rendered image.
When the observed tone differs from that of the prior, for example because shading is baked into the source
(Sec.~\ref{sec:supp_limitations}), the transition can remain visible as a seam, most noticeably where the
boundary crosses a thin structure.
Sec.~\ref{sec:supp_canonical_diag} quantifies the color step at this boundary.
The source-level shift $\boldsymbol{\Delta}^{aid}_{s}$ is a single per-channel shift shared by all
Gaussians (Sec.~\ref{sec:supp_architecture}) and therefore introduces no spatial variation of its own.
The $360^\circ$ rotations in the supplementary video (Sec.~\ref{sec:supp_videos}) show the transition
between the source-observed and prior-completed sides directly.

\paragraph{Role separation.}
As noted after Eq.~\ref{eq:compose} of the main paper, the reliability $\gamma_i$ and the appearance gate
$g_i$ act on the source evidence only through their product $g_i\gamma_i$, but the two factors differ in
kind.
$\gamma_i$ is a closed-form estimate of how reliably a Gaussian is observed.
It is computed from the rasterizer outputs and the source mask, receives no gradient, and does not predict
appearance.
Appearance itself is predicted by the learned modules, namely the local evidence tokens, the global source
identity token, and the source-conditioned prior bank $\mathcal{D}_s$, and the appearance gate $g_i$
decides how much of the observed color to adopt where evidence exists.
We do not learn $\gamma_i$, since a learned confidence could be satisfied from dataset priors rather than
from the observation process.
A single learned scalar in place of $g_i\gamma_i$ would lose the one factor that training cannot adjust.
The ablations confirm that neither factor absorbs the other.
With every variant retrained through the full two-stage pipeline, removing $\gamma$ costs $3.03$\,dB and
removing $g_i$ costs $1.69$\,dB (Table~\ref{tab:ablation} of the main paper).

\subsection{Architecture Details}
\label{sec:supp_architecture}

This section specifies every learned module in Eqs.~\ref{eq:img_tokens}--\ref{eq:compose} and
Figs.~\ref{fig:pipeline} and~\ref{fig:source_appearance} of the main paper.
In Fig.~\ref{fig:pipeline}, the conditional transformer consists of $\Phi_g$ and $\Phi_a$, the point
embedding is $\phi_q$, and the scale, offset, and color heads are $f_s$, $f_\Delta$, and
$f_{\mathrm{rgb}}$.
In Fig.~\ref{fig:source_appearance}, Prior Refinement is $\mathcal{U}$, Identity Conditioning comprises
$\mathcal{B}$ and $\mathcal{U}_c$, Appearance Prediction is $f_{out}$, and Appearance Composition is
Eq.~\ref{eq:compose}.

\paragraph{Image encoder and fusion head.}
$E_{\mathrm{img}}$ applies a frozen DINOv3 ViT-L/16 backbone (patch size~16) to the $256\times256$ crop,
giving a $16\times16$ grid of patch tokens.
The patch tokens of blocks 4, 11, 17, and 23 are each passed through a $1\times1$ convolution,
concatenated along channels, and fused by a second $1\times1$ convolution into a
$d_{\mathrm{img}}=1{,}024$-channel map at patch resolution.
This map is flattened into the $N_f=256$ image tokens $\mathbf{F}_s$ of Eq.~\ref{eq:img_tokens}.
Only this fusion head is trained (in Stage~1), and the backbone stays frozen throughout.

\paragraph{Query embedding and transformer streams (Eqs.~\ref{eq:anchors}--\ref{eq:decoder}).}
The query anchors $\mathbf{q}_i$ of Eq.~\ref{eq:anchors} are the $N=49{,}281$ vertices of the $k=3$
subdivided $\mathrm{MANO}(\boldsymbol{\beta},\boldsymbol{\theta}_{\mathrm{mean}})$ mesh.
The point-embedding layer $\phi_q$ encodes each anchor with sinusoidal features at eight octaves per axis,
concatenates the raw coordinates, and applies a linear layer with LayerNorm to obtain a $d=512$-dimensional
token.
$\Phi_g$ consists of two SD3-style joint-attention blocks (8 heads, feed-forward ratio~4) in which the $N$
query tokens and the $N_f$ image tokens, linearly projected from $d_{\mathrm{img}}$ to $d$, attend to one
another.
The image stream is updated in the first block and serves only as keys and values in the second, and a
final LayerNorm yields $\mathbf{Z}^g$.
$\Phi_a$ is a single cross-attention layer without a feed-forward block (queries $\mathbf{Z}^g$, keys and
values $\mathbf{F}_s$, 8 heads) that produces $\mathbf{Z}^a$.
The Gaussian decoder of Eq.~\ref{eq:decoder} applies one shared trunk MLP (two hidden layers of width 512,
SiLU) separately to $\mathbf{z}^{g}_i$ and $\mathbf{z}^{a}_i$, followed by one linear output layer per
attribute, namely $f_\Delta$ (3 values), $f_s$ (1 value, replicated to the isotropic scale), and
$f_{\mathrm{rgb}}$ (3 values, sigmoid), with the output bounds of Sec.~\ref{sec:supp_training}. 
\paragraph{Footprint aggregation and local evidence (Eqs.~\ref{eq:agg}, \ref{eq:footprint},
and~\ref{eq:local_evidence}).}
$\mathrm{Agg}_{\Omega_i}$ samples a $7\times7$-pixel window centered at the projected Gaussian center
$\mathbf{u}_i$ with Gaussian weights
$w_i(\mathbf{p})=\exp\bigl(-\lVert\mathbf{p}-\mathbf{u}_i\rVert^2/2\sigma_i^2\bigr)$, where $\sigma_i$ is
half the projected radius reported by the rasterizer, with a minimum of $0.5$ pixels.
The same window is applied to $I_s$, $\hat{I}_s$, $M_s$, and $\mathbf{F}^{x}_{s}$.
$\mathbf{F}^{x}_{s}$ is produced by a ResNet-18 U-Net (ImageNet-pretrained encoder) at $1/4$ of the input
resolution ($64\times64$) with 64 channels.
It is trained in Stage~2 and is separate from the DINOv3 encoder.
$f_{loc}$ is a two-layer MLP ($3+3+3+64+1\to d_a\to d_a$, GELU, LayerNorm) followed by a residual MLP block
($d_a\to4d_a\to d_a$, GELU), with $d_a=384$.

\paragraph{Global source appearance identity.}
The global source appearance identity token $\mathbf{t}^{aid}_{s}$ is computed once per source image from
the concatenation of
(i)~the $M_s$-masked spatial mean of $\mathbf{F}_s$ ($d_{\mathrm{img}}$ values),
(ii)~the $M_s$-masked spatial mean of $\mathbf{F}^{x}_{s}$ (64),
(iii)~the $M_s$-masked mean and standard deviation of $I_s$ (6),
(iv)~the mean and standard deviation of the base colors $\mathbf{c}^{0}_i$ over all $N$ Gaussians (6),
(v)~the $\gamma$-weighted mean and standard deviation over Gaussians of the residuals
$\bar{\mathbf{c}}^{src}_{i}-\bar{\mathbf{c}}^{ren}_{i}$ (6), and
(vi)~the $\gamma$-weighted mean of the local evidence tokens $\mathbf{z}^{loc}_i$ ($d_a$).
A two-layer MLP (GELU, LayerNorm) maps this vector to $\mathbf{t}^{aid}_{s}\in\mathbb{R}^{d_a}$.

\paragraph{Prior Refinement and Identity Conditioning (Eq.~\ref{eq:prior_refine}).}
Prior Refinement $\mathcal{U}$ projects $\mathbf{Z}^a$ to width $d_a$ with a two-layer MLP and applies two
cross-attention blocks (pre-LayerNorm on queries and memory, 8 heads, residual feed-forward of ratio~4)
whose memory is the $N+1$ tokens $\mathbf{Z}^{loc}\cup\{\mathbf{t}^{aid}_{s}\}$.
In Identity Conditioning, the bank $\mathcal{D}_s=\mathcal{B}(\mathbf{t}^{aid}_{s})$ consists of $64$
learned prototypes in $\mathbb{R}^{d_a}$, each shifted by the same two-layer MLP image of
$\mathbf{t}^{aid}_{s}$ (additive conditioning).
$\mathcal{U}_c$ then forms the query
$\mathrm{MLP}\bigl([\mathbf{z}^{loc}_i,\tilde{\mathbf{z}}^{a}_i,\mathbf{t}^{aid}_{s},
\mathrm{MLP}(\mathbf{c}^{0}_i)]\bigr)$ ($4d_a\to d_a\to d_a$) and applies four cross-attention blocks of
the same form whose memory is the $65$ tokens $\{\mathbf{t}^{aid}_{s}\}\cup\mathcal{D}_s$, giving
$\mathbf{h}_i$.

\paragraph{Appearance Prediction and Composition (Eqs.~\ref{eq:app_pred} and~\ref{eq:compose}).}
$f_{out}$ consists of three two-layer MLP heads ($d_a\to d_a\to$ output) that produce
$\mathbf{c}^{obs}_i=\mathrm{sigmoid}(\cdot)$, $g_i=\mathrm{sigmoid}(\cdot)$, and
$\boldsymbol{\delta}_i=0.1\tanh(\cdot)$.
The output bias of the $g_i$ head is initialized to $-1.5$ so that transfer starts close to the base
colors.
$\mathrm{Pool}$ in Eq.~\ref{eq:app_pred} is the unweighted mean over all $N$ Gaussians, and the composed
color of Eq.~\ref{eq:compose} is clipped to $[0,1]$.
The cross-attention blocks and the residual MLP of the appearance module use dropout $0.1$.
Stage~1 trains the fusion head, $\phi_q$, $\Phi_g$, $\Phi_a$, and the Gaussian decoder, and Stage~2 trains
the appearance module, including the ResNet-18 U-Net.

\section{Datasets and Sampling}
\label{sec:supp_dataset}

\subsection{Training Datasets}
\label{sec:supp_training_datasets}

\begin{figure}[!h]
    \centering
    \includegraphics[width=0.95\linewidth,height=0.42\textheight,keepaspectratio]{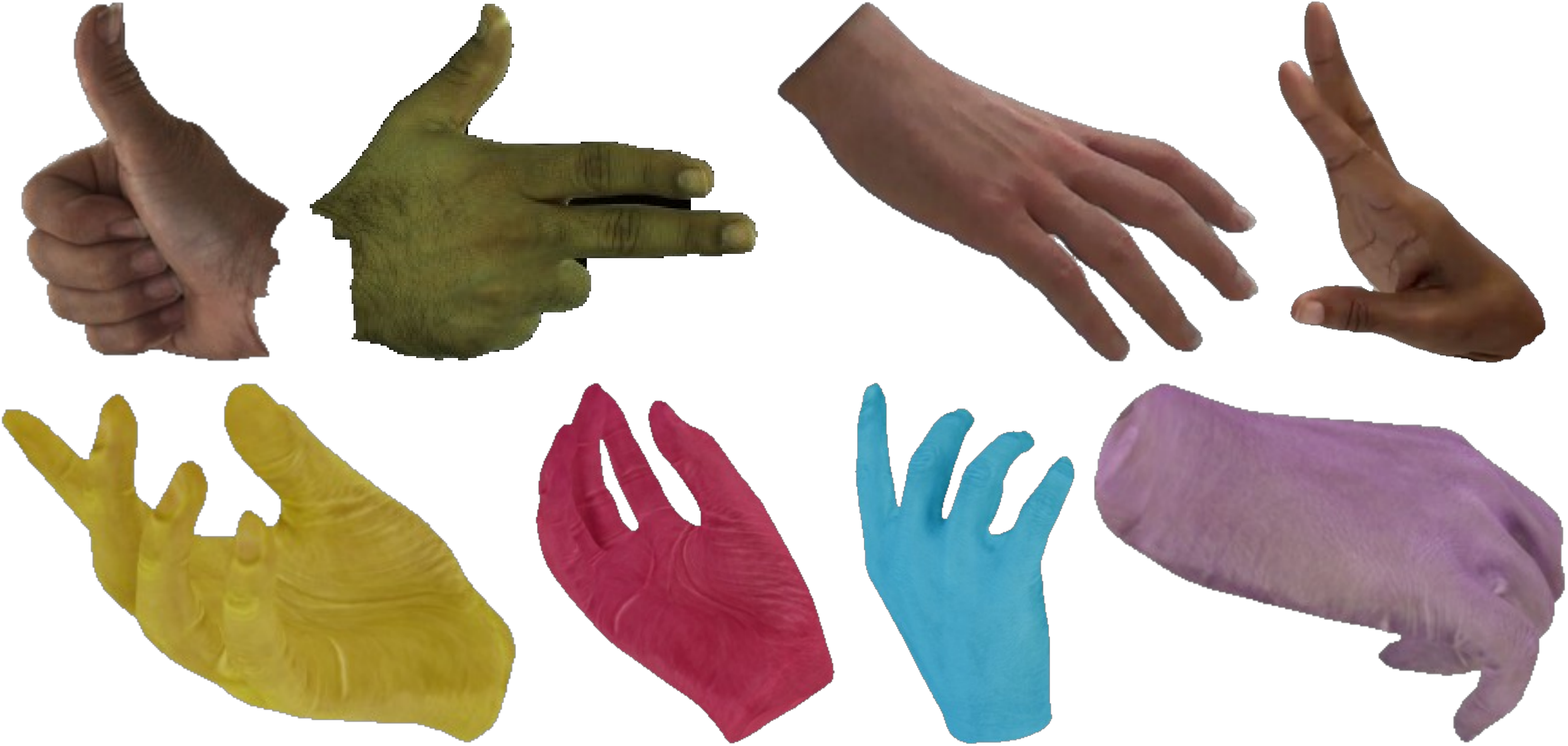}
    \caption{Examples of real (top) and synthetic (bottom) training data.
    Synthetic samples are used only in the source-conditioned appearance stage
    (Sec.~\ref{sec:supp_synthetic}).}
    \label{fig:supp_dataset_examples}
\end{figure}

We train InstantHand using a mixture of real and synthetic hand datasets.
The real datasets include PALM~\citep{fan2026palm},
InterHand2.6M~\citep{Moon_2020_ECCV_InterHand2.6M},
HO3Dv3~\citep{hampali2020ho3d,2021ho3dv3}, Goliath~\citep{martinez2024Goliath},
and Hand Appearance (HARP)~\citep{karunratanakul2023harp}.
For Goliath and InterHand2.6M, left and right hands are treated as separate
subjects after preprocessing.
In total, the real training datasets contain 304 subjects and 339{,}532 frames.
For InterHand2.6M, we train only on the official training split, so the evaluation subject
(\texttt{test/Capture0}) and all HanCo subjects are unseen during training.
Figure~\ref{fig:supp_dataset_examples} shows representative training examples
from the real datasets (top row) and from the synthetic appearance set of
Sec.~\ref{sec:supp_synthetic} (bottom row).
Hand regions are shown segmented on a white background for visualization.

PALM and InterHand2.6M provide large-scale multi-view observations with diverse
poses and identities, while HO3Dv3 introduces hand--object interactions and
partial occlusions.
Goliath and HARP provide additional appearance diversity.
Synthetic hand data is used only for training the source-conditioned Gaussian
appearance module, as described in Sec.~\ref{sec:supp_synthetic}.

\subsection{Synthetic Hand Data Generation}
\label{sec:supp_synthetic}

For the source-conditioned appearance stage, we augment training with synthetic
source--target pairs generated following \citet{Zhao_2025_CVPR}.
In addition to realistic hand appearances, we sample diverse hand colors to expose the appearance module
to a broader range of source appearances
(bottom row of Fig.~\ref{fig:supp_dataset_examples}).
This augmentation is used only for training the source-conditioned Gaussian appearance module, and the
canonical Gaussian representation stage is trained on real hand datasets only.

Each synthetic sample is converted into the same source--target format as real
training data. A single source image provides the appearance condition, while
target frames provide supervision under different poses and viewpoints.
All synthetic images are cropped, resized, and processed using the same
preprocessing pipeline as real images.

\subsection{Dataset Sampling Strategy}
\label{sec:supp_sampling}

We use a subject-centric sampling strategy to construct source--target training
pairs. Each dataset is organized by subject, and each subject contains one or more
groups. A group denotes a subject-internal observation bucket, such as a camera,
sequence, session, or multi-camera unit depending on the dataset.
For each training iteration, we first sample a dataset according to a
stage-specific dataset weight, then uniformly sample a subject from the selected
dataset. A source frame is sampled from the subject-level source pool, and six
target frames are sampled from the same subject according to dataset-specific
self, same-group, and cross-group target counts.

\begin{table}[!h]
    \centering
    \caption{Dataset sampling policies used for training.
    Targets per source image are reported as self / same-group / cross-group.}
    \label{tab:supp_sampling_policy}
    \small
    \setlength{\tabcolsep}{5pt}
    \begin{tabular}{@{}llccc@{}}
    \toprule
    Stage & Dataset & Weight & Source stride & Targets \\
    \midrule
    Canonical & PALM           & 0.45 & 1  & 1 / 1 / 4 \\
    Canonical & InterHand2.6M  & 0.30 & 1  & 1 / 2 / 3 \\
    Canonical & HO3Dv3         & 0.10 & 10 & 1 / 3 / 2 \\
    Canonical & Goliath        & 0.10 & 5  & 1 / 5 / 0 \\
    Canonical & HARP           & 0.05 & 5  & 1 / 5 / 0 \\
    \midrule
    Appearance & PALM          & 0.35 & 1  & 1 / 1 / 4 \\
    Appearance & InterHand2.6M & 0.30 & 1  & 1 / 2 / 3 \\
    Appearance & Synthetic     & 0.15 & 1  & 1 / 1 / 4 \\
    Appearance & HO3Dv3        & 0.08 & 10 & 1 / 3 / 2 \\
    Appearance & Goliath       & 0.08 & 5  & 1 / 5 / 0 \\
    Appearance & HARP          & 0.04 & 5  & 1 / 5 / 0 \\
    \bottomrule
    \end{tabular}
\end{table}

The self target corresponds to the source frame itself and provides source-view
supervision. Same-group targets are sampled from the same subject group as the
source, while cross-group targets are sampled from different groups of the same
subject. For multi-view datasets, cross-group targets usually correspond to different cameras or observation
groups.
For monocular datasets, they correspond to different subject-internal sequences when available.
This strategy keeps identity and hand shape consistent while exposing the model to
varying poses and viewpoints.
Table~\ref{tab:supp_sampling_policy} summarizes the sampling weights and
target-frame composition used in each stage.

\section{Evaluation Protocol}
\label{sec:supp_eval}

We follow the single-image animatable hand reconstruction protocol established by
prior hand avatar benchmarks~\citep{handavatar,zheng2024ohta}, including their
sequence, designated source-frame, cropping, and masking choices, to ensure direct
comparability with reported baselines.
The InterHand2.6M source frame is the one provided with the official OHTA evaluation configuration.
For HanCo, whose protocol leaves the source frame unspecified, we fix one common source frame for all
methods (Sec.~\ref{sec:supp_eval_hanco}).
Given a single source image, the model reconstructs a canonical Gaussian hand
representation, which is then rendered under novel target poses and viewpoints for
evaluation. All methods receive the identical source image, ground-truth
annotations, and evaluation targets.

\subsection{InterHand2.6M Evaluation}
\label{sec:supp_eval_interhand}

We follow the evaluation protocol used in HandAvatar~\citep{handavatar} and
OHTA~\citep{zheng2024ohta}.
We use the following sequence and, as the source image, the example provided in
the official OHTA repository:
\begin{center}
\small
\begin{tabular}{@{}ll@{}}
Sequence     & \texttt{test/Capture0/ROM03\_RT\_No\_Occlusion} \\
Source image & \texttt{test\_Capture0\_ROM03\_RT\_No\_Occlusion\_cam400272\_image15012.jpg}
\end{tabular}
\end{center}

Following HandAvatar, the hand region is cropped using the provided evaluation
bounding box. The bounding box is converted into a square region with a
$1.3\times$ enlargement factor, and the cropped image is resized to $256\times256$.
Unless otherwise specified, the same cropping protocol is used for all experiments.
We use the ground-truth camera parameters and MANO parameters provided by the dataset, and the foreground
evaluation masks are ray masks generated following the HandAvatar evaluation protocol.
The source image itself is excluded from quantitative evaluation.
Table~\ref{tab:main_quant} of the main paper reports the results under this protocol.
The corresponding evaluation with \emph{estimated} source-side cues is reported in Table~\ref{tab:ablation}
of the main paper and, with boundary-level metrics, in Sec.~\ref{sec:supp_source_sensitivity}.

\subsection{HanCo Evaluation}
\label{sec:supp_eval_hanco}

For HanCo we follow the zero-shot cross-dataset evaluation setting of
OHTA~\citep{zheng2024ohta}. We use sequence 0191 and evaluate frames from cam3,
cam5, cam6, and cam7, where the cameras are positioned above the hands.
Following OHTA, we do not use cameras below the hands due to inconsistent lighting
conditions and color calibration.
We use the 76th and last frame of cam6 (frame index 75) as the common source image for all methods,
together with the ground-truth MANO parameters and hand masks provided by HanCo.
Unlike the InterHand2.6M protocol, the HanCo evaluation set retains all 76 frames
of each of the four cameras (304 frames), so the source frame itself is one of the
evaluation frames.
It is retained for every compared method, so relative comparisons are unaffected,
and it contributes one of the 304 frames to the aggregate metrics.

Preprocessing and evaluation-mask generation are identical to the InterHand2.6M
protocol of Sec.~\ref{sec:supp_eval_interhand}: the hand region is cropped to a
square with a $1.3\times$ enlargement factor and resized to $256\times256$, and
metrics are computed over ray masks generated following the HandAvatar evaluation
protocol, applied consistently to all methods.
The only difference is the bounding-box source: since HanCo does not provide the
HandAvatar evaluation bounding boxes, the square crop is instead derived from the
minimum and maximum coordinates of the ground-truth foreground mask.
Because the OHTA HanCo protocol designates the held-out sequence (0191) but not the
source frame, absolute numbers are not directly comparable to previously reported HanCo values.
All methods are therefore re-evaluated by us from the identical source frame under identical inputs
(Sec.~\ref{sec:supp_baseline_inputs}).
Table~\ref{tab:hanco_results} of the main paper reports the results under this protocol.

\subsection{Metric Computation}
\label{sec:supp_metrics}

Following the HandAvatar protocol, pixels outside the ray mask are set to the background
color in both the rendered and the ground-truth image, and PSNR, SSIM, and LPIPS are
computed on the resulting images, which reduces background bias.
Because our rasterizer ignores the principal point, we render with a centered principal point and map the
render to the original crop with the exact homography, so all methods are scored on the same pixels.
All evaluations are performed at $256\times256$ resolution.

\subsection{Boundary-aware Silhouette Metric Definitions}
\label{sec:supp_boundary_metrics}

Boundary F-score and boundary IoU are computed between the binarized rendered alpha
($A>0.5$) and the ground-truth mask, with boundaries extracted as 1-pixel contours.
The boundary F-score is the standard contour-matching F-measure: predicted and
ground-truth boundary pixels are matched within a distance tolerance of 2 pixels via
distance transforms.
Boundary IoU is a variant of the boundary-restricted IoU of
\citet{cheng2021boundary}: each 1-pixel contour is dilated by 3 pixels and the IoU is
computed between the two resulting bands.
The band is thus symmetric about the silhouette rather than restricted to the mask
interior, so it also registers outward alpha leakage, which an interior-only band ignores.
Alpha leakage is the mean excess of the rendered alpha above a tolerance of $0.1$
over the outer band, constructed as a 5-pixel dilation of the ground-truth
silhouette excluding the silhouette itself, and interior coverage is the mean
rendered alpha over the interior eroded by 5 pixels.
The evaluation bands are deliberately wider than, and distinct from, the 2-pixel
training bands of Sec.~\ref{sec:supp_sab}, so the boundary evaluation does not
simply re-measure the region that the SAB loss optimizes.
All metrics are computed at $256\times256$ and, unless stated otherwise, averaged over the InterHand2.6M
novel-pose/novel-view evaluation set.
They are reported in Table~\ref{tab:recipe_port} of the main paper and in Tables~\ref{tab:supp_sensitivity}
and~\ref{tab:supp_densify}, the latter also on the HanCo evaluation set.

\subsection{Self-reconstruction Protocol}
\label{sec:supp_selfrecon_protocol}

Self-reconstruction is a controlled source-view appearance diagnostic: it measures
how faithfully the input image is reflected in the avatar when no pose or viewpoint
gap is involved.
Each frame of the evaluation data is used in turn as the input: the avatar is
reconstructed from that frame and rendered back into the same pose and camera, so
the target is the input itself.
All non-color attributes are fixed to the Stage-1 canonical values, so
any metric difference between the w/o transfer and Ours rows of
Table~\ref{tab:supp_selfrecon} is attributable \emph{solely} to the
source-conditioned color update.
The w/o transfer row renders the Stage-1 base colors $\mathbf{c}^0_i$ (the
w/o appearance transfer configuration of Table~\ref{tab:ablation} in the main paper),
and the Ours row renders the composed colors $\mathbf{c}^*_i$ (Eq.~\ref{eq:compose} of the main paper).
Table~\ref{tab:supp_selfrecon} reports the average over the evaluation frames of
each benchmark (the InterHand2.6M sequence of Sec.~\ref{sec:supp_eval_interhand}
and the HanCo sequence of Sec.~\ref{sec:supp_eval_hanco}), with the same cropping
protocol and ray masks as Sec.~\ref{sec:supp_metrics}.

This diagnostic complements the novel-pose/novel-view ablation in Table~\ref{tab:ablation} of the main
paper by isolating whether the color update follows the observed source appearance while geometry is
fixed.
Figure~\ref{fig:supp_selfrecon_qual} (Sec.~\ref{sec:supp_selfrecon_qual}) shows representative frames.

\begin{table}[!tb]
    \centering
    \caption{Self-reconstruction with geometry fixed
    (Sec.~\ref{sec:supp_selfrecon_protocol}). w/o transfer renders the Stage-1 base colors, and Ours renders
    the composed colors.}
    \label{tab:supp_selfrecon}
    \small
    \setlength{\tabcolsep}{6pt}
    \begin{tabular}{@{}llccc@{}}
    \toprule
    Dataset & Setting & PSNR$\uparrow$ & SSIM$\uparrow$ & LPIPS$\downarrow$ \\
    \midrule
    InterHand2.6M     & w/o transfer & 26.58 & 0.869 & 0.133 \\
                      & Ours         & \textbf{28.54} & \textbf{0.903} & \textbf{0.108} \\
    \midrule
    HanCo (zero-shot) & w/o transfer & 20.47 & 0.839 & 0.187 \\
                      & Ours         & \textbf{23.88} & \textbf{0.898} & \textbf{0.153} \\
    \bottomrule
    \end{tabular}
\end{table}

\subsection{Timing and Efficiency Measurement}
\label{sec:supp_timing}

The 2.84\,s reported in the main paper is the wall-clock \emph{avatar-network} inference time on a single
NVIDIA A100 (80\,GB), covering image encoding, MANO-anchored token computation, Gaussian decoding,
source-view rendering, and source-conditioned appearance composition.
It excludes disk I/O, data loading, input preprocessing, one-time model loading, and CUDA initialization,
and it is averaged over repeated runs after warm-up (standard deviation 0.007\,s).

\paragraph{Cost breakdown.}
Building the canonical Gaussians alone takes 1.38\,s under the same protocol.
Source-conditioned appearance composition, including the source-view rendering it requires, therefore
accounts for the remaining 1.46\,s, about half of the avatar-network time.

\paragraph{End-to-end raw image to avatar.}
Because the reported 2.84\,s assumes that MANO, camera, and mask inputs are already
available, we additionally report the complete raw-image-to-avatar cost on the same
GPU (Table~\ref{tab:supp_runtime}).
Upstream estimation adds approximately $0.46$\,s ($0.38$\,s for the SAM 3 foreground mask and $0.08$\,s
for WiLoR MANO and camera estimation), giving an end-to-end construction time of approximately $3.30$\,s.
Upstream estimation is thus about $14\%$ of the end-to-end cost and is not the bottleneck.
This entire cost is incurred \emph{once per avatar}, for example when a user's hand is enrolled at the
start of an AR/VR or telepresence session, and subsequent animation re-runs neither network and renders
at 161\,FPS.
The construction time is therefore not a per-frame latency, and we do not claim real-time acquisition.

\begin{table}[!tb]
    \centering
    \caption{Raw-image-to-avatar runtime on a single NVIDIA A100 (80\,GB)
    under the protocol of Sec.~\ref{sec:supp_timing}.}
    \label{tab:supp_runtime}
    \small
    \setlength{\tabcolsep}{6pt}
    \begin{tabular}{@{}llc@{}}
    \toprule
    Stage & Component & Time \\
    \midrule
    Upstream estimation
      & SAM 3 foreground mask~\citep{carion2026sam3}        & 0.38\,s \\
      & WiLoR MANO $+$ camera~\citep{potamias2025wilor}    & 0.08\,s \\
      & \emph{subtotal}                                   & \emph{0.46\,s} \\
    \midrule
    Avatar construction
      & Canonical Gaussian construction                   & 1.38\,s \\
      & Source-conditioned appearance composition         & 1.46\,s \\
      & \emph{subtotal} (Table~\ref{tab:hanco_results} of the main paper)       & \emph{2.84\,s} \\
    \midrule
    \textbf{End-to-end} & raw image $\rightarrow$ animatable avatar & \textbf{$\approx$3.30\,s} \\
    \midrule
    Animation & novel-pose rendering (per avatar, repeated) & 160.71\,FPS \\
    \bottomrule
    \end{tabular}
\end{table}

All baseline measurements in Table~\ref{tab:hanco_results} of the main paper
(per-subject fitting time, rendering speed, and peak memory) are taken on the same
NVIDIA A100 (80\,GB).
Relative to our 2.84\,s avatar-network inference, the fitting times of HandAvatar, OHTA, and BIGS$^{*}$
correspond to speedups of $1964\times$, $1183\times$, and $486\times$ ($5579/2.84$, $3360/2.84$, and
$1380/2.84$).
The ``over 400$\times$'' figure quoted in the main paper is the most conservative of these.

\paragraph{Rendering speed and memory.}
FPS measures novel-pose rendering (MANO-driven deformation and rasterization) at
$256\times256$ with batch size~1 over the HanCo evaluation frames, excluding
data loading: 161\,FPS ($\approx6.2$\,ms per frame).
Peak GPU memory, including the CUDA context, is 1.79\,GB during rendering and
4.78\,GB during avatar construction.
Since all compared methods receive the same ground-truth inputs
(Sec.~\ref{sec:supp_baseline_inputs}), upstream cost is identical across the
compared systems and orthogonal to the reported comparison.

\subsection{Baseline Input Protocols}
\label{sec:supp_baseline_inputs}

Table~\ref{tab:supp_input_protocol} states, for every compared method, its
originally intended input modality and the input actually provided in our
evaluation.
All methods receive the identical single source frame and identical ground-truth
MANO, camera, and mask annotations, following the single-image protocol
established by prior hand avatar benchmarks~\citep{handavatar,zheng2024ohta}, and no method receives
additional frames.
For Handy, a parametric shape and appearance model built from large-scale 3D hand
scans, the single-frame evaluation follows the setting established by
OHTA~\citep{zheng2024ohta}: the provided MANO annotations are converted into Handy's
format, and its texture latent code is fitted to the single source frame via
differentiable rendering.

\begin{table}[!tb]
    \centering
    \caption{Input protocol for every compared method.}
    \label{tab:supp_input_protocol}
    \small
    \setlength{\tabcolsep}{6pt}
    \begin{tabular}{@{}lll@{}}
    \toprule
    Method & Original input & Input here \\
    \midrule
    HARP        & Monocular video   & Single frame \\
    LiveHand    & Multi-view video  & Single frame \\
    HandAvatar  & Monocular video   & Single frame \\
    Handy       & Single image      & Single frame \\
    OHTA        & Single image      & Single frame \\
    BIGS$^{*}$  & Monocular video   & Single frame \\
    Ours        & Single image      & Single frame \\
    \bottomrule
    \end{tabular}
\end{table}

\paragraph{Provenance of baseline numbers.}
The provenance of the reported baseline numbers differs between the two benchmarks.
On InterHand2.6M, we adopt the single-frame results of HARP, LiveHand, HandAvatar,
Handy, and OHTA established under this protocol by \citet{zheng2024ohta},
and evaluate our method under the identical protocol.
BIGS$^{*}$ postdates that work, so its InterHand2.6M numbers are measured by us
under the same protocol using the single-hand adaptation of
Sec.~\ref{sec:supp_bigs}.
On HanCo, the exact evaluation configuration behind previously reported numbers
(evaluation frames, source frame, and mask generation) is not fully specified
publicly, so directly citing them would not yield a controlled comparison.
We therefore re-evaluate every compared method ourselves under the fully specified protocol of
Sec.~\ref{sec:supp_eval_hanco}, using the same source frame, ground-truth annotations, crops, and masks for
all methods.
All qualitative comparisons (Fig.~\ref{fig:qual} of the main paper) are rendered by us
from our own runs of each baseline under the protocols of
Secs.~\ref{sec:supp_eval_interhand} and~\ref{sec:supp_eval_hanco}, so the
qualitative and quantitative comparisons share the same source frames, crops,
and masks.

\subsection{Protocols of the Additional Analyses}
\label{sec:supp_diag_protocols}

The pose-stratified and temporal analyses of Sec.~\ref{sec:supp_analyses} reuse the
renders, ground truth, and ray masks of the main evaluation
(Secs.~\ref{sec:supp_eval_interhand} and~\ref{sec:supp_eval_hanco}), and only the reporting changes.
The canonical diagnostics use the avatars reconstructed from the two protocol source frames.

\paragraph{Pose strata.}
For every evaluation frame we measure two changes of the driving pose relative to the
source pose.
The \emph{palm-orientation change} is the geodesic distance between the source and
target global hand rotations, each expressed in the frame of its own camera, so it
captures both the rotation of the hand and the change of camera.
The \emph{finger-articulation change} is the mean geodesic distance over the 15 finger
joints of MANO, excluding the global rotation.
The frames are split into equal-count terciles (Easy, Medium, Extreme) of each
measure, and every method is scored on the same frames.
For the palm terciles, we also report the median change of the viewing direction in the hand frame, which
measures how far the camera has moved around the hand irrespective of in-plane roll.
Tables~\ref{tab:supp_pose_strata} and~\ref{tab:supp_articulation_methods} (Sec.~\ref{sec:supp_pose_strata})
report the results.

\paragraph{Temporal stability.}
On the continuous HanCo sequence we form, per camera, the 75 pairs of consecutive
evaluation frames (300 pairs in total).
For a pair $(t,t{+}1)$ with ray masks $M_t$ and $M_{t+1}$, the frame-to-frame change
of the prediction is
\begin{equation}
d_{\mathrm{pred}}
=\frac{1}{\lvert M_t\cup M_{t+1}\rvert}
\sum_{\mathbf{p}\in M_t\cup M_{t+1}}
\frac{1}{3}
\bigl\lVert \hat I_{t+1}(\mathbf{p})-\hat I_{t}(\mathbf{p})\bigr\rVert_1,
\label{eq:supp_temporal}
\end{equation}
where the $L_1$ norm runs over the three color channels of the same composited 8-bit images used for the
metrics.
We define $d_{\mathrm{gt}}$ identically on the ground truth, and the excess variation is
$E_t=d_{\mathrm{pred}}-d_{\mathrm{gt}}$.
For each camera, which records one continuous video, we average $E_t$ over its
consecutive pairs and report the result relative to the mean ground-truth change of that camera.
The summary over cameras averages the magnitudes of these per-camera values, so that
opposite biases of different cameras do not cancel.
As a calibration, we add to our own renders a per-frame brightness offset inside the ray mask, drawn
uniformly within $\pm1\%$, $\pm2\%$, or $\pm4\%$, and recompute $E_t$.
Table~\ref{tab:supp_temporal} (Sec.~\ref{sec:supp_temporal}) reports the results.

\paragraph{Canonical boundary diagnostic.}
The Gaussians are partitioned by their reliability into a high-reliability set ($\gamma_i\ge0.5$), a
transition set, and a low-reliability set ($\gamma_i<0.1$).
Neighbors are the six nearest Gaussians in canonical space, and color differences are CIE76 $\Delta E$
values in CIELAB between the final colors $\mathbf{c}^*_i$.
We compare three kinds of neighbor pairs.
Pairs within the high-reliability set measure the natural texture variation and serve as the reference,
pairs that straddle the threshold $\gamma_i=0.5$ measure the transition, and \emph{strict} pairs connect a
high-reliability and a low-reliability Gaussian directly.
Sec.~\ref{sec:supp_canonical_diag} reports the results.

\paragraph{Reliability and opacity renderings.}
The reliability maps (Fig.~\ref{fig:supp_color_reliability}) render $\gamma_i$ as a scalar field with the
same rasterizer, and the alpha-weighted mean of $\gamma$ along each ray is color-mapped after compositing,
so blending never produces colors outside the color map.
The opacity maps (Fig.~\ref{fig:supp_opacity}) show the accumulated alpha $A$ of the render, the quantity
behind the boundary metrics of Sec.~\ref{sec:supp_boundary_metrics}.
The avatar is posed with the source pose and rendered from the source view and from
an orbit of the source camera about its vertical image axis through the centroid of the
posed Gaussians.

\section{Extended Related Work}
\label{sec:supp_related}

\paragraph{Hand avatar reconstruction.}
A large body of work relies on multi-view inputs or monocular video sequences,
combined with neural implicit representations or explicit mesh-based
pipelines~\citep{corona2022lisa,handavatar,mundra2023livehand,karunratanakul2023harp,guo2023handnerf,kalshetti2024intrinsichand}.
HandAvatar~\citep{handavatar} models pose-dependent appearance using implicit
representations with self-occlusion-aware texture modeling.
HARP~\citep{karunratanakul2023harp} employs a mesh-based formulation augmented
with learned reflectance maps to capture high-frequency details, while
LiveHand~\citep{mundra2023livehand} leverages multi-view video and neural
rendering to synthesize photorealistic hand appearance.
Despite achieving high visual fidelity, these approaches depend on dense
observations and per-subject optimization, leading to substantial computational
overhead and limited scalability.
Single-image methods such as OHTA~\citep{zheng2024ohta} introduce data-driven
priors for one-shot hand avatar reconstruction, but still require test-time
fitting or inversion.
Gaussian-based approaches such as
GaussianHand~\citep{Zhao2025GaussianHand} adopt explicit 3D Gaussian primitives
for efficient rendering but continue to depend on multi-view supervision or
temporal sequences and often require subject-specific optimization.
As summarized in Table~\ref{tab:method_comparison}, existing methods remain constrained by dense inputs or
per-subject optimization, whether they use implicit fields, meshes, or Gaussian primitives.

\begin{table}[!t]
\centering
\caption{Comparison of representative 3D hand avatar reconstruction methods under their original input settings.
BIGS$^{*}$: our single-hand adaptation of BIGS (Sec.~\ref{sec:supp_bigs}).
$^{\S}$GaussianHand reports results from as few as 5--20 camera views.}
\label{tab:method_comparison}
\vspace{0.4em}
\footnotesize
\setlength{\tabcolsep}{10pt}
\begin{tabular}{@{}lccc@{}}
\toprule
\textbf{Method} & \textbf{Input} & \textbf{Feed-forward} & \textbf{Repr.} \\
\midrule
LiveHand~\citep{mundra2023livehand}         & Multi-view video & \xmark & NeRF-like \\
GaussianHand$^{\S}$~\citep{Zhao2025GaussianHand}   & Multi-view video & \xmark & 3DGS \\
HandAvatar~\citep{handavatar}               & Monocular video  & \xmark & Implicit \\
HARP~\citep{karunratanakul2023harp}         & Monocular video  & \xmark & Mesh \\
BIGS$^{*}$~\citep{on2025bigs}               & Monocular video  & \xmark & 3DGS \\
OHTA~\citep{zheng2024ohta}                  & Single image     & \xmark & Implicit \\
\midrule
\rowcolor{gray!8}
\textbf{Ours} & \textbf{Single image} & \cmark & \textbf{3DGS} \\
\bottomrule
\end{tabular}
\end{table}

\paragraph{Feed-forward 3D Gaussian avatar reconstruction.}
Feed-forward avatar methods regress the parameters of animatable 3D Gaussians
from one or a few images, replacing per-subject optimization with amortized inference.
Among them, LHM~\citep{qiu2025LHM} extends
Gaussian representations to animatable full-body avatars, while
GUAVA~\citep{GUAVA} focuses on upper-body reconstruction.
For head avatars, approaches such as GAGAvatar~\citep{chu2024gagavatar} and
LAM~\citep{he2025lam} demonstrate high-fidelity rendering via feed-forward
Gaussian prediction.
Extending these designs to hands exposes the two failure modes discussed in
Sec.~\ref{sec:intro} of the main paper: thin, articulated structures are sensitive
to small geometric errors, and appearance regressed from image features over-relies
on learned priors under single-view ambiguity.
The controlled recipe ports of Sec.~\ref{sec:supp_recipe_ports} quantify the former.

\section{Baseline Adaptations}
\label{sec:supp_baseline}

\subsection{BIGS-based Hand Reconstruction Baseline}
\label{sec:supp_bigs}

BIGS~\citep{on2025bigs} originally targets bimanual hand--object interaction
reconstruction from monocular videos, and therefore does not directly match our
single-image, single-hand evaluation setting.
To construct a reference optimization-based 3DGS baseline, we adapt BIGS to a
single-hand setting by retaining only the Gaussian hand avatar component and
removing the object and second-hand branches.

The adapted baseline represents the hand in a canonical Gaussian space.
Following the Gaussian hand avatar modeling design, Gaussian attributes are decoded
from a triplane feature representation using MLP heads: the geometry head predicts
Gaussian offsets, rotations, and scales, the appearance head predicts colors and
opacities, and the deformation head predicts MANO-related skinning weights.
The canonical Gaussians are then transformed to posed space using MANO-based linear
blend skinning and rendered with differentiable Gaussian rasterization.

The original formulation is designed for monocular videos without known camera
poses, where the camera coordinate system and hand trajectory must be estimated or
aligned. In our benchmark setting, ground-truth annotations are available, so for controlled fitting and
evaluation we use the ground-truth MANO parameters, camera parameters, and hand masks whenever available.
This removes errors from external hand-pose, camera, and mask estimation, and
focuses the comparison on the reconstruction and rendering formulation itself.
Accordingly, the fixed virtual-camera construction and the 2D-keypoint-based global
trajectory alignment of the original pipeline are bypassed, and the temporal
smoothness regularization defined across consecutive video frames is inactive under
single-frame fitting.

The adapted baseline is fitted to the single source frame by minimizing a masked
reconstruction objective restricted to the foreground hand region, combining an
$L_1$ term and an SSIM term, together with a silhouette loss on the rendered
opacity and a regularizer that encourages the predicted skinning weights to remain
close to the corresponding MANO skinning priors.
Fitting uses Adam for 15{,}000 iterations, with learning rates of
$1.6\times10^{-4}$ for the Gaussian position parameters, $1.0\times10^{-4}$ for the
remaining Gaussian attribute parameters, and $1.0\times10^{-3}$ for the triplane and
MLP decoder weights.

During evaluation, the fitted BIGS-based baseline is rendered using the same target
MANO poses, target cameras, hand-centered crops, and foreground masks as
InstantHand.
Since this protocol adapts BIGS to a single-hand benchmark and differs from the
original bimanual hand--object interaction setting, we report it as a reference
optimization-based 3DGS baseline (BIGS$^{*}$) rather than as BIGS in its original setting.
Its results appear in Tables~\ref{tab:main_quant} and~\ref{tab:hanco_results} of the main paper, and its
input protocol in Table~\ref{tab:supp_input_protocol}.

\subsection{Controlled Recipe Ports of One-shot Head and Body Avatar Methods}
\label{sec:supp_recipe_ports}

A natural question is whether existing feed-forward one-shot \emph{head} and
\emph{full-body} avatar frameworks can simply be retargeted to hands.
To answer this empirically rather than by argument, we construct controlled
hand adaptations of both families and evaluate them under matched conditions.
All quantitative results are reported in Table~\ref{tab:recipe_port} of the main paper.
This section specifies the two ports and states what the resulting gap does and does not show.

\paragraph{Matched-condition protocol.}
In both ports, the training data, MANO anchors, image encoder, optimization
schedule, and compute budget are identical to ours, and only the representation and supervision recipe
differ.
Free per-Gaussian parameters are initialized from our constrained solution so that
the comparison does not confound optimization difficulty with initialization.
Both ports are evaluated under the InterHand2.6M protocol of
Sec.~\ref{sec:supp_eval_interhand} with the boundary metrics of
Sec.~\ref{sec:supp_boundary_metrics}.

\paragraph{Head recipe~\citep{he2025lam}.}
We replace FLAME with MANO while retaining the method's free per-Gaussian opacity,
rotation, and anisotropic scale, together with its direct image-feature-based RGB
prediction. Our boundary-aware supervision and our reliability-gated color pathway
are both removed, so this port evaluates the published parameterization rather than
a hybrid.

\paragraph{Body recipe~\citep{qiu2025LHM}.}
Directly reproducing LHM~\citep{qiu2025LHM} is infeasible at our scale.
It uses a 0.5--1B-parameter model trained on approximately 300K videos with 32--64 A100
GPUs, and its released model requires full-body images and SMPL-X parameters rather than
cropped hand images and MANO parameters.
We therefore construct a controlled recipe-level port.
SMPL-X anchors are replaced by MANO anchors, and the freely regressed Gaussian
attributes and canonical-space regularization of~\citet{qiu2025LHM} are retained as
released.
Writing $s^{(1)}_i\ge s^{(2)}_i\ge s^{(3)}_i$ for the sorted per-axis scales of a freed
Gaussian, the as-spherical-as-possible term takes its released ratio-hinge form,
$\max\bigl(s^{(1)}_i/s^{(3)}_i-\tau_{\mathrm{ASAP}},\,0\bigr)$ with
$\tau_{\mathrm{ASAP}}=5$, at the released hand weighting $\lambda_{\mathrm{ASAP}}=10^{5}$.
Because this term is a ratio of scales, it transfers to MANO unchanged.
The position-anchoring term
$\frac{1}{N}\sum_i\max(\lVert\Delta\mathbf{x}_i\rVert_2-\tau_{\mathrm{ACAP}},\,0)$ and the
offset range are stated in~\citet{qiu2025LHM} relative to a $1.8$\,m body
($\tau_{\mathrm{ACAP}}=1.8/32$\,m, weight $10^{3}$, range $1.0$\,m).
We apply the same numbers as fractions of the template extent ($\approx0.19$\,m for
MANO): $\tau_{\mathrm{ACAP}}=\mathrm{extent}/32\approx5.9$\,mm,
$\lambda_{\mathrm{ACAP}}=10^{3}\cdot1.8/\mathrm{extent}\approx9.5\times10^{3}$, and an
offset range of $\mathrm{extent}\cdot(1.0/1.8)\approx0.106$\,m.
This choice preserves the range-to-threshold ratio of the original exactly.
This soft anchoring replaces the hard offset bound and L2 offset prior of the head
recipe, since the two are one design choice.
The mask weight follows Eq.~7 of~\citet{qiu2025LHM} (0.5), and the backbone, data,
anchors, schedule, and compute are matched to ours.
At the weights of Eq.~10 of~\citet{qiu2025LHM} (50/10), the anisotropy hinge never
activates (mean scale ratio $\approx3<\tau_{\mathrm{ASAP}}$), and the anchoring term
contributes $\approx0.25\%$ of the loss, so the free attributes are effectively unregularized.
We therefore report the hand weighting released in the official implementation
of~\citet{qiu2025LHM}, under which the anchoring term is active.
We report this as a recipe port, not as a reproduction of the original system at its
original scale.
Matched renderings of both ports are shown in Fig.~\ref{fig:supp_recipe_ports_qual}.

\paragraph{Reading the reported gap.}
Both ports are worse than our constrained model without SAB in boundary F-score and leakage
(boundary F $0.267$--$0.299$ vs.\ $0.366$, leakage $0.608$--$0.627$ vs.\ $0.555$).
The two ports are independent configurations, each retaining its own published parameterization, yet they
score close to each other (PSNR within $0.24$\,dB, boundary F $0.267$ vs.\ $0.299$).
Their anchoring terms differ in kind.
The head recipe uses hard anchoring, an L2 prior within a $\pm4$\,mm bound, and the body recipe uses soft
anchoring, a hinge that is inactive within $\tau_{\mathrm{ACAP}}\approx5.9$\,mm.
At our framing, this leaves a free shell of $\approx6$\,px, three times the 2-px training band.
Neither term observes the rendered silhouette, however, and neither recovers boundary precision.
Supervision on the silhouette itself does.
Under the constrained parameterization, SAB raises the boundary F-score from $0.366$ to $0.698$
(Table~\ref{tab:recipe_port} of the main paper and Fig.~\ref{fig:supp_sab_qual}).
The contribution of the attribute constraints with boundary supervision held fixed
is isolated in Sec.~\ref{sec:supp_attr_ablation}.

\section{Additional Ablation Studies}
\label{sec:supp_ablation}

Unless otherwise specified, all variants are evaluated on InterHand2.6M using the
same protocol as the main paper.

\subsection{Effect of the Number of Gaussians}
\label{sec:supp_mano_subdivision}

\begin{figure}[!h]
    \centering
    \includegraphics[width=\linewidth,height=0.36\textheight,keepaspectratio]{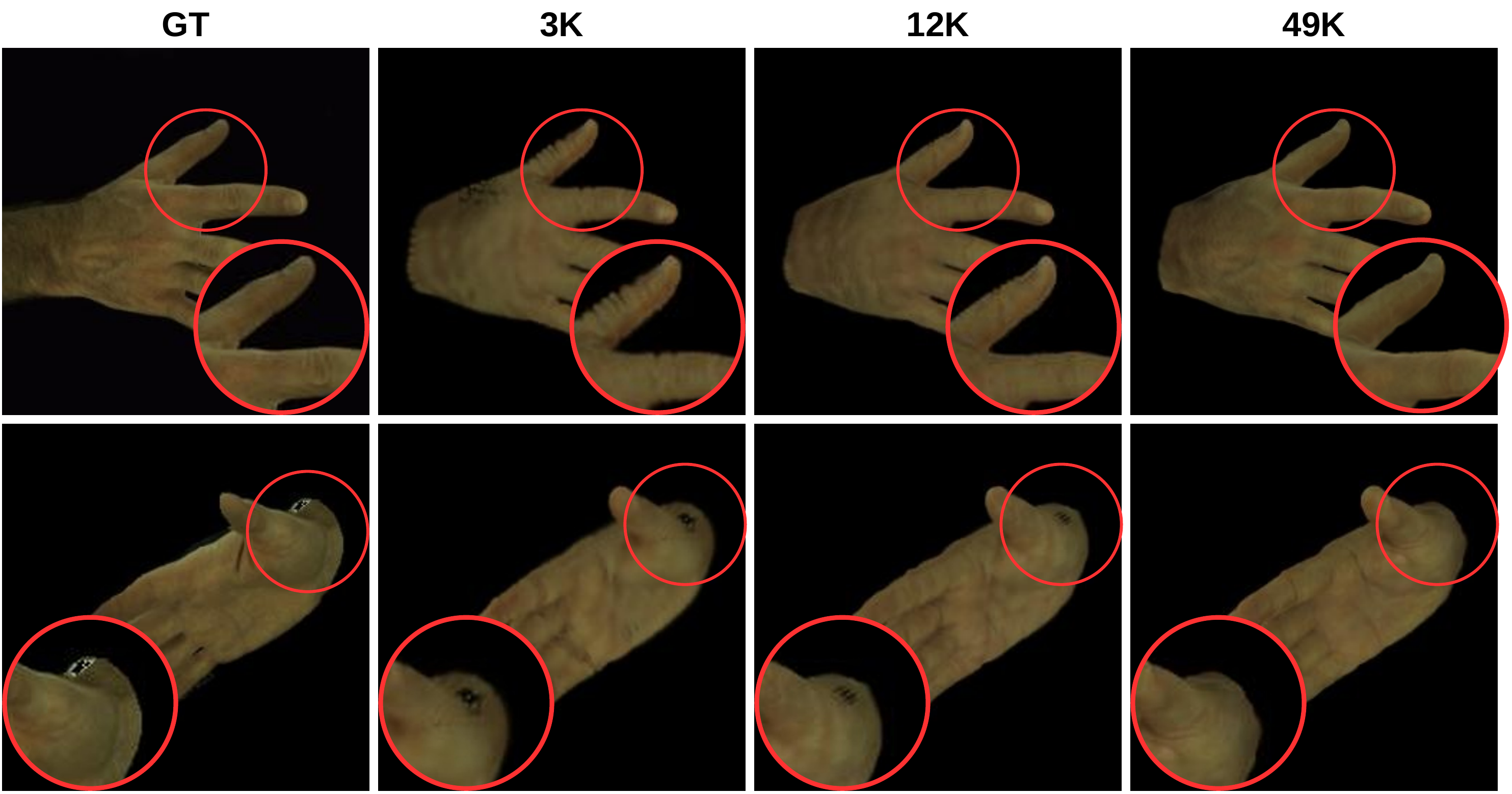}
    \caption{Qualitative effect of the number of Gaussians.
    Columns: ground truth (GT) and the 3K, 12K, and 49K settings of
    Table~\ref{tab:supp_mano_subdivision}. Red circles mark regions magnified in the corner insets.}
    \label{fig:supp_mano_subdivision_qual}
\end{figure}

\begin{table}[!h]
    \centering
    \caption{Ablation on the number of Gaussians (MANO subdivision level).
    All variants are full models differing only in the subdivision level.}
    \label{tab:supp_mano_subdivision}
    \small
    \setlength{\tabcolsep}{6pt}
    \begin{tabular}{@{}ccccc@{}}
        \toprule
        $k$ & \#Gaussians & PSNR$\uparrow$ & SSIM$\uparrow$ & LPIPS$\downarrow$ \\
        \midrule
        1 & 3{,}093  & 26.11 & 0.859 & 0.168 \\
        2 & 12{,}337 & 25.92 & 0.860 & 0.147 \\
        3 & 49{,}281 & \textbf{26.21} & \textbf{0.866} & \textbf{0.133} \\
        \bottomrule
    \end{tabular}
\end{table}

InstantHand anchors Gaussian primitives on a subdivided MANO surface ($\mathcal{S}_k$ in
Eq.~\ref{eq:anchors} of the main paper), so the subdivision level directly controls the density of query anchors and therefore the
number of canonical Gaussians. Subdivision levels $k=1$, $k=2$, and $k=3$ produce
3{,}093, 12{,}337, and 49{,}281 Gaussian anchors, respectively, and the main model uses $k=3$.
A lower subdivision level provides a compact anchor set but may underrepresent thin
fingers, hand boundaries, and local surface details, while a higher level provides
denser surface coverage.
Each variant is trained with the full two-stage pipeline and evaluated as a
complete model under the novel-pose/novel-view protocol, and only the subdivision level differs.

Table~\ref{tab:supp_mano_subdivision} and Fig.~\ref{fig:supp_mano_subdivision_qual}
show the effect.
The figure compares full-model reconstructions, rendered on the same black background, with ground
truth.
SSIM and LPIPS improve consistently with denser anchors (LPIPS
$0.168\to0.147\to0.133$), while PSNR stays within $0.29$\,dB.
In the sparse 3K setting, the gaps between neighboring splats appear as serrated
finger contours (red circles), and surface detail is coarser.
Such local, high-frequency defects barely move pixel-averaged PSNR but are captured by LPIPS.
With denser anchors the contours become smooth and continuous, and at 49K the
finger silhouettes are sharp.
Training at the next level, $k=4$ (196{,}993 anchors), runs out of memory on our
training GPUs (RTX A6000, 48\,GB) and was therefore not feasible with our resources.
Rendering the trained $k=3$ prediction at $k=4$ density without retraining is examined in
Sec.~\ref{sec:supp_limitations} (Table~\ref{tab:supp_densify}).

\subsection{Gaussian Attribute Constraints}
\label{sec:supp_attr_ablation}

Our canonical parameterization deliberately fixes opacity and canonical rotation
and constrains scales to be isotropic (Sec.~\ref{sec:supp_training}), as in prior
monocular Gaussian-avatar designs~\citep{hu2024gaussianavatar,moon2024exavatar}.
This subsection isolates the cost of relaxing each constraint in the hand setting.

\begin{figure}[!h]
    \centering
    \includegraphics[width=0.85\linewidth,height=0.40\textheight,keepaspectratio]{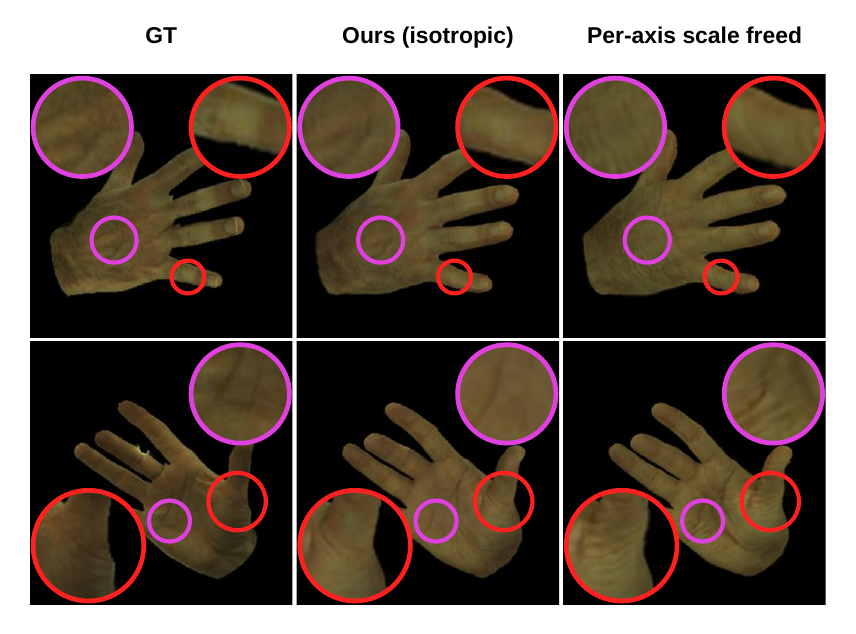}
    \caption{Artifacts under per-axis Gaussian scales.
    Ground truth, the isotropic model (ours), and the per-axis-scale-freed variant
    of Table~\ref{tab:supp_attr_ablation} on identical frames. Magenta circles mark interior regions and red
    circles silhouette regions, each magnified in the matching corner inset.}
    \label{fig:supp_aniso_artifacts}
\end{figure}

\begin{table}[!h]
    \centering
    \caption{Effect of relaxing the Gaussian attribute constraints
    (InterHand2.6M). Boundary supervision is retained in every row.}
    \label{tab:supp_attr_ablation}
    \small
    \setlength{\tabcolsep}{6pt}
    \begin{tabular}{@{}lccc@{}}
        \toprule
        Configuration & PSNR$\uparrow$ & SSIM$\uparrow$ & LPIPS$\downarrow$ \\
        \midrule
        Ours (isotropic; rotation, opacity fixed) & \textbf{26.21} & \textbf{0.866} & \textbf{0.133} \\
        Per-axis scale freed                      & 25.58 & 0.856 & 0.142 \\
        Rotation and opacity freed                & 25.74 & 0.862 & \textbf{0.133} \\
        All three freed                           & 25.50 & 0.856 & 0.148 \\
        \bottomrule
    \end{tabular}
\end{table}

\paragraph{Matched protocol.}
Boundary supervision is retained in \emph{every} configuration of
Table~\ref{tab:supp_attr_ablation}, with all other settings unchanged and
per-axis scales initialized isotropically, so the differences isolate the
attribute parameterization.
The effect of boundary supervision under the constrained parameterization is
reported separately in Table~\ref{tab:recipe_port} of the main paper, where removing SAB lowers PSNR by
$2.32$\,dB, to $23.89$\,dB.

Under matched supervision, freeing per-axis scale alone reduces PSNR by
$0.63$\,dB.
Freeing rotation and opacity reduces PSNR by $0.47$\,dB.
Because rotation does not alter the covariance of an isotropic Gaussian, this row primarily measures the
effect of learning opacity.
Freeing all three produces the largest degradation, $0.71$\,dB, and worsens SSIM
and LPIPS as well.
Curved finger geometry is instead captured by increasing the density of small
isotropic Gaussians (LPIPS $0.168\to0.133$ from 3K to 49K, Table~\ref{tab:supp_mano_subdivision}).
Rendering the trained 49K prediction at 197K density with scales $0.6\,s_i$, without retraining, lowers it
further to $0.129$ (Table~\ref{tab:supp_densify}).

\paragraph{Artifacts under per-axis scales.}
Because the metric gap alone does not identify \emph{how} the anisotropic variant
fails, we compare the variant with only per-axis scales freed and the isotropic model on identical
frames, so that observed differences are attributable specifically to the scale parameterization.
Figure~\ref{fig:supp_aniso_artifacts} shows two artifact types, with boundary
supervision retained in both models.
The magenta circles fall on the dorsum (top) and the palm (bottom), and the red circles on a fingertip and
the thumb.
The isotropic model preserves the vein and crease structure of the ground truth,
whereas the per-axis variant replaces it with parallel bands and widens the
silhouette fringe.
\emph{(i) Striated interiors.} Broad, smooth regions such as the back of the hand
exhibit quasi-parallel, regularly spaced bands, producing a corduroy-like ripple
consistent with elongated splats becoming similarly aligned.
These patterns are distinguishable from anatomical texture, since veins branch and
follow the hand geometry whereas the artifacts remain approximately parallel and
uniformly spaced.
The distortions occur \emph{within} the rendered surface rather than as holes or
transparent regions, which is why opacity-based measures such as interior coverage
barely register them, while LPIPS degrades from $0.133$ to $0.142$.
\emph{(ii) Feathered silhouettes.} The per-axis model produces a wider
semi-transparent fringe around the hand boundary, consistent with elongated
Gaussians extending across the true contour, so finger boundaries appear smeared
while the matched isotropic renders remain substantially sharper.

\subsection{Sensitivity to Source-side MANO and Mask Errors}
\label{sec:supp_source_sensitivity}

We evaluate how errors in the source-side MANO parameters and hand mask affect
novel-pose/novel-view rendering on InterHand2.6M.
All experiments use the protocol-designated source frame of
Sec.~\ref{sec:supp_eval_interhand}, fixed network weights, and the same evaluation
protocol and metrics as the main paper.
Only the specified source cue is changed: the source RGB image and crop, target
MANO parameters and cameras, and ground-truth target evaluation masks remain fixed.
Results are averaged over the same target frames.
The estimated-cue substitution rows appear in Table~\ref{tab:ablation} (bottom) of the main paper, and
Table~\ref{tab:supp_sensitivity} reports the complete study with boundary-level
metrics.

\paragraph{Controlled perturbations.}
For pose perturbations we add independent zero-mean Gaussian noise directly to the
45 articulated MANO axis--angle components of the source frame,
\begin{equation}
\boldsymbol{\theta}'=\boldsymbol{\theta}+\boldsymbol{\xi},
\qquad \boldsymbol{\xi}\sim\mathcal{N}(0,\sigma_{\mathrm{pose}}^2),
\label{eq:supp_posenoise}
\end{equation}
where $\sigma_{\mathrm{pose}}\in\{5^\circ,10^\circ\}$ is the component-wise standard deviation.
The global wrist orientation, translation, shape, and camera remain unchanged, and
each pose-noise condition is averaged over three random draws.
For mask perturbations we apply two iterations of binary erosion or dilation to the
source mask before the otherwise identical source preprocessing, and the source image and crop remain
unchanged.

\paragraph{Estimated source-cue substitution.}
We separately replace the ground-truth source cues with outputs from practical
upstream estimators.
SAM 3~\citep{carion2026sam3} replaces the source hand mask, and
WiLoR~\citep{potamias2025wilor} replaces the source MANO pose and shape fields.
The mask--MANO combinations form a $2\times2$ substitution study that separates the individual and
combined effects of the estimated cues.
We additionally report the \emph{fully estimated} configuration, in which the jointly regressed WiLoR
camera also replaces the ground-truth source camera, so that every source-side input originates from a
single RGB image.
This camera differs from the calibrated ground truth in both intrinsics and extrinsics.
Unlike the controlled perturbations, these estimator outputs contain structured and
spatially correlated errors.

\begin{table}[!tb]
    \centering
    \caption{Sensitivity to source-side MANO and mask errors on
    InterHand2.6M.
    (a) Controlled perturbations around the ground-truth source cues.
    (b) Ground-truth source cues replaced by SAM 3 and WiLoR outputs. The last row also replaces the source
    camera.}
    \label{tab:supp_sensitivity}
    \small
    \setlength{\tabcolsep}{4.5pt}
    \begin{tabular}{@{}lcccccc@{}}
    \multicolumn{7}{@{}l}{(a) Controlled perturbations around ground truth}\\
    \toprule
    Source perturbation & PSNR$\uparrow$ & SSIM$\uparrow$ & LPIPS$\downarrow$ & Bound.~F$\uparrow$ & B-IoU$\uparrow$ & Leakage$\downarrow$ \\
    \midrule
    Clean GT inputs                                     & 26.21 & 0.866 & 0.133 & 0.698 & 0.562 & 0.234 \\
    Pose noise ($\sigma_{\mathrm{pose}}=5^\circ$)       & 26.12 & 0.864 & 0.134 & 0.698 & 0.562 & 0.234 \\
    Pose noise ($\sigma_{\mathrm{pose}}=10^\circ$)      & 26.05 & 0.863 & 0.134 & 0.698 & 0.562 & 0.234 \\
    Mask erosion (2\,px)                                & 26.28 & 0.867 & 0.132 & 0.704 & 0.565 & 0.229 \\
    Mask dilation (2\,px)                               & 26.24 & 0.864 & 0.134 & 0.697 & 0.561 & 0.237 \\
    \bottomrule
    \end{tabular}

    \vspace{0.6em}

    \setlength{\tabcolsep}{4.5pt}
    \begin{tabular}{@{}llccccc@{}}
    \multicolumn{7}{@{}l}{(b) Estimated source-cue substitution}\\
    \toprule
    Source mask & Source MANO & PSNR$\uparrow$ & SSIM$\uparrow$ & LPIPS$\downarrow$ & Bound.~F$\uparrow$ & Leakage$\downarrow$ \\
    \midrule
    GT   & GT    & 26.21 & 0.866 & 0.133 & 0.698 & 0.234 \\
    SAM 3 & GT    & 26.12 & 0.865 & 0.133 & 0.695 & 0.239 \\
    GT   & WiLoR & 26.07 & 0.863 & 0.134 & 0.689 & 0.235 \\
    SAM 3 & WiLoR & 25.99 & 0.862 & 0.135 & 0.686 & 0.239 \\
    \multicolumn{2}{@{}l}{SAM 3 $+$ WiLoR $+$ WiLoR camera} & 25.99 & 0.863 & 0.134 & 0.685 & 0.240 \\
    \bottomrule
    \end{tabular}
\end{table}

Table~\ref{tab:supp_sensitivity}(a) shows that controlled pose perturbations produce
gradual but small appearance changes: relative to clean ground-truth inputs,
$\sigma_{\mathrm{pose}}=5^\circ$ and $10^\circ$ reduce PSNR by $0.09$ and $0.16$\,dB
and SSIM by $0.002$ and $0.003$, while boundary F-score, boundary IoU, and leakage
are unchanged at the reported precision, as expected since the source pose enters
only the appearance pathway.
Two-pixel mask erosion and dilation change PSNR by at most $0.07$\,dB and the
boundary F-score by at most $0.006$, so the model is insensitive to uniform mask
morphology at this scale.

Structured estimator errors produce a somewhat larger but still limited gap.
With ground-truth MANO fixed, replacing the source mask with SAM 3 decreases PSNR by
$0.09$\,dB and boundary F by $0.003$.
With the ground-truth mask fixed, replacing the source MANO with WiLoR decreases PSNR by $0.14$\,dB and
boundary F by $0.009$.
Unlike pose noise, which leaves the boundary metrics unchanged, WiLoR also replaces the shape
$\boldsymbol{\beta}$ that places the canonical anchors (Eq.~\ref{eq:anchors} of the main paper), which is
consistent with its change in boundary F.
Using both estimated cues decreases PSNR by $0.22$\,dB and boundary F by $0.012$.
At the reported precision, the combined degradation is close to the sum of the individual changes, which
suggests limited interaction between the two source-cue errors.
Replacing the source camera as well, which changes the source intrinsics and
extrinsics, leaves PSNR unchanged at the reported precision and moves each boundary
metric by only $0.001$.
The fully estimated configuration thus sits $0.22$\,dB below clean ground-truth inputs.

Qualitative results under fully estimated inputs on unseen images are shown in
Figs.~\ref{fig:freihand_qualitative} and~\ref{fig:supp_animation}.

\subsection{Inference-time Sensitivity of Observation Reliability}
\label{sec:supp_gamma_sensitivity}

We keep the trained checkpoint fixed and vary the reliability hyperparameters of Eq.~\ref{eq:gamma} of the
main paper one at a time around the default $(\kappa,\tau_m,\zeta)=(0.5,0.5,95)$.
For each setting, $\gamma$ is recomputed once from the Stage-1 source-view render
and the complete appearance-composition forward pass is rerun, while no network weights or canonical
Gaussian attributes are updated.
This is therefore a local inference-time sensitivity analysis rather than a
comparison of separately retrained models.
We run the sweep under two protocols, the InterHand2.6M protocol used for the other ablations and the
held-out HanCo protocol of Sec.~\ref{sec:supp_eval_hanco}, so that the analysis covers both subject-level
and dataset-level generalization.

\begin{table}[!tb]
    \centering
    \caption{Inference-time sensitivity to observation-reliability
    hyperparameters. One parameter is varied at a time with all weights fixed, and bold marks the default.}
    \label{tab:supp_gamma_sensitivity}
    \small
    \setlength{\tabcolsep}{6pt}
    \begin{tabular}{@{}ccccc@{}}
    \toprule
    $\kappa$ & $\tau_m$ & $\zeta$ & PSNR$\uparrow$ & LPIPS$\downarrow$ \\
    \midrule
    \multicolumn{5}{@{}l}{(a) InterHand2.6M (novel-pose/novel-view)}\\
    0.25 & 0.5 & 95 & 26.21 & 0.133 \\
    \textbf{0.5} & \textbf{0.5} & \textbf{95} & \textbf{26.21} & \textbf{0.133} \\
    1.0  & 0.5 & 95 & 26.20 & 0.133 \\
    0.5  & 0.3 & 95 & 26.22 & 0.133 \\
    0.5  & 0.7 & 95 & 26.19 & 0.133 \\
    0.5  & 0.5 & 90 & 26.22 & 0.134 \\
    0.5  & 0.5 & 99 & 26.19 & 0.132 \\
    \midrule
    \multicolumn{5}{@{}l}{(b) HanCo (zero-shot, sequence 0191)}\\
    0.25 & 0.5 & 95 & 20.16 & 0.187 \\
    \textbf{0.5} & \textbf{0.5} & \textbf{95} & \textbf{20.18} & \textbf{0.188} \\
    1.0  & 0.5 & 95 & 20.21 & 0.188 \\
    0.5  & 0.3 & 95 & 20.18 & 0.188 \\
    0.5  & 0.7 & 95 & 20.18 & 0.188 \\
    0.5  & 0.5 & 90 & 20.17 & 0.188 \\
    0.5  & 0.5 & 99 & 20.20 & 0.187 \\
    \bottomrule
    \end{tabular}
\end{table}

Table~\ref{tab:supp_gamma_sensitivity} shows a plateau rather than a peak.
Across the seven settings, PSNR spans $0.03$\,dB ($26.19$--$26.22$) and LPIPS
$0.002$, even though $\kappa$ varies by a factor of four and $\tau_m$ by
more than a factor of two.
For scale, removing reliability gating entirely costs $3.03$\,dB
(Table~\ref{tab:ablation} of the main paper), roughly two orders of magnitude more than any
hyperparameter change here.
The behavior of reliability gating is therefore governed by its structure, namely square-root
compression, median-relative thresholding, and percentile normalization (Sec.~\ref{sec:supp_reliability}),
rather than by the exact constants.
The same insensitivity holds under dataset-level transfer: on the held-out HanCo
sequence the same seven settings span $0.05$\,dB PSNR ($20.16$--$20.21$) and
$0.001$ LPIPS.
In practice, reliability gating requires no per-dataset tuning.

\subsection{Rationale for Two-stage Training}
\label{sec:supp_twostage}

The two-stage schedule enforces a structural separation between canonical
representation learning and source-conditioned appearance adaptation.
Stage 1 learns the canonical predictor, and Stage 2 freezes it in full, including the base-color head
$f_{\mathrm{rgb}}$, while optimizing only the source-conditioned appearance module of
Sec.~\ref{sec:source_conditioned_gaussian_appearance} of the main paper (training details in
Sec.~\ref{sec:supp_training}).
By construction, source-view appearance supervision in Stage 2 cannot alter the
canonical geometry or base colors, so novel-pose rendering starts from exactly the
Stage-1 representation.

The frozen canonical branch also gives the appearance module a well-defined prior
to fall back on: wherever source evidence is missing or unreliable, the composed
color reverts to the Stage-1 base color plus the source-level shift $\boldsymbol{\Delta}^{aid}_{s}$ shared
by all Gaussians.
The full model improves source-view fidelity (Table~\ref{tab:supp_selfrecon}), and
observation reliability controls what reaches novel views.
A color taken from the source is only as trustworthy as the evidence behind it.
When transfer is not weighted by $\gamma$, novel-pose PSNR falls from $26.21$ to $23.18$\,dB
(Table~\ref{tab:ablation} of the main paper), which is consistent with colors of weakly observed Gaussians
being carried to novel views.
The two stages thus keep the two roles separate: Stage 1 provides a stable prior
for every Gaussian, and Stage 2 learns only where, and how strongly, source
evidence should replace it, together with a source-level shift that propagates
global appearance cues to every Gaussian.

\section{Discussion}
\label{sec:supp_discussion}

\subsection{Relation to Boundary Losses}
\label{sec:supp_boundary_relation}

Boundary-focused objectives and metrics are established in segmentation, where
regional losses are complemented by contour-based losses~\citep{kervadec2019boundary}
and by boundary-aware evaluation~\citep{cheng2021boundary}.
Rendered alpha, however, differs from the pixel-labeling setting those losses assume,
and each element of SAB (Eq.~\ref{eq:sab} of the main paper and Sec.~\ref{sec:supp_sab}) addresses one of
the resulting differences.
Per-pixel alpha is not an independent prediction but a composite of many projected
3D primitives, so band supervision acts through the rasterizer on primitive centers
and scales rather than relabeling pixels.
A one-sided outer penalty admits a degenerate optimum with no segmentation analogue, namely shrinking the
hand until nothing crosses the contour.
The paired interior coverage term closes this loophole, so leakage cannot be ``fixed'' by shrinking the
hand.
Soft alpha at boundaries is correct anti-aliased rendering rather than an error, and the tolerances and the
robust penalty leave it unpenalized.
Finally, the schedule $\eta(n)$ reflects the 3D setting by activating the penalty only once the bands are
geometrically meaningful for the evolving primitives.
All SAB ablations keep the constrained parameterization of Sec.~\ref{sec:supp_training} fixed, so the
boundary-aware silhouette evaluation (Sec.~\ref{sec:supp_boundary_metrics} and Table~\ref{tab:recipe_port}
of the main paper) measures the effect of SAB under identical constraints.

\subsection{View-dependent Appearance, Illumination, and Multi-frame Extension}
\label{sec:supp_viewdep}

Our appearance model predicts view-independent Gaussian colors and does not employ
view-dependent representations such as spherical harmonics (SH).
The two address different problems.
The ambiguity we target is observational.
A single source image observes essentially one side of the hand, so the core difficulty is deciding where
source evidence is reliable and how to complete unobserved regions, which is what reliability-gated
transfer and the global source identity address.
A similar observational problem arises for human avatars built from monocular video, where
FlexiAvatar~\citep{tiruneh2026flexiavatar} optimizes only the visible body regions and
completes entirely unseen regions with diffusion-generated texture.
View-dependent appearance modeling instead captures viewpoint- and
illumination-dependent reflectance after the underlying appearance has been recovered, and it typically
requires multi-view observations to be constrained.
Under single-image supervision, per-Gaussian SH coefficients would be severely underdetermined.
The two mechanisms are therefore complementary rather than alternatives:
view-dependent terms could be attached to our composed colors without
changing the geometry stabilization or the reliability-gated transfer.

A related limitation is that shading baked into the source image is transferred as
albedo and thus carried to novel poses.
This is as much a property of the setting as of the model: a single image does not
determine illumination and albedo separately.
Explicit albedo--shading decompositions such as those of
HandAvatar~\citep{handavatar} and OHTA~\citep{zheng2024ohta} model self-occlusion
shading learned from video or multi-identity data, but the scene illumination of a
single source image, such as its exposure or color cast, is not identifiable from
that image alone.
The per-subject baselines, which are fitted to the same source frame, are not exempt.
In Part~4 of the supplementary video (Sec.~\ref{sec:supp_videos}), they also depart from the ground truth in
tone and shading, with dark discoloration for OHTA and a uniform tone without shading for HandAvatar.
Disentangling illumination from intrinsic appearance is therefore likewise future
work, and the corresponding failures are documented in
Sec.~\ref{sec:supp_limitations}.
Finally, the formulation already extends to multiple source frames at inference time, without
retraining or additional parameters.
Since $\gamma$ is computed per source frame from the same rasterizer by-products, each additional frame is
mapped onto the shared canonical Gaussians, the most reliable observation is selected for each Gaussian,
and the colors are composed with Eq.~\ref{eq:compose} of the main paper.
Related work similarly registers observations from disjoint viewpoints into a shared 3D space for
multi-person mesh tracking across video shots~\citep{on2026multi}.
Frames beyond the first are selected to maximize the union coverage, the fraction of canonical Gaussians
with nonzero reliability ($\gamma_i>0$) in at least one source frame, and they span cameras $12^\circ$--$178^\circ$ apart
as well as different poses and time steps.
Table~\ref{tab:supp_multiview} reports the results under the InterHand2.6M protocol of
Sec.~\ref{sec:supp_eval_interhand}.

\begin{table}[!tb]
    \centering
    \caption{Multi-frame evidence integration at inference time (InterHand2.6M).
    The same trained model is used without retraining or additional parameters, and frames beyond the
    first are selected to maximize the union coverage of observed Gaussians ($\gamma_i>0$).}
    \label{tab:supp_multiview}
    \small
    \setlength{\tabcolsep}{6pt}
    \begin{tabular}{@{}ccccc@{}}
    \toprule
    Source frames & Coverage & PSNR$\uparrow$ & SSIM$\uparrow$ & LPIPS$\downarrow$ \\
    \midrule
    1 & 0.45 & 26.21 & 0.866 & 0.133 \\
    2 & 0.68 & 26.53 & 0.867 & 0.131 \\
    4 & 0.71 & 26.52 & 0.867 & 0.132 \\
    8 & 0.96 & \textbf{26.57} & \textbf{0.868} & \textbf{0.130} \\
    \bottomrule
    \end{tabular}
\end{table}

A single complementary view that covers the unseen palm or dorsum improves all three metrics
($+0.32$\,dB PSNR).
The gain follows newly observed surface rather than the number of frames.
From two to four frames, coverage grows by only $0.03$ and PSNR is effectively unchanged ($-0.01$\,dB),
whereas from four to eight frames coverage grows by $0.25$ and all three metrics improve again.
The gain per unit of coverage nevertheless diminishes near full coverage, from $+0.32$\,dB for the first
$0.23$ of added coverage to $+0.05$\,dB for the last $0.25$, for two reasons.
First, overlapping observations of an already covered region are handled by reliability-gated selection
rather than learned fusion.
The reliability separates observed from unobserved Gaussians within a single view but is not designed to
rank several valid observations of the same region, so near-ties are resolved in favor of the base view.
Second, the last uncovered regions are visible only at grazing angles, where they receive low reliability
and occupy little area in the evaluation views.
Learned multi-view fusion, trained with explicit multi-view or temporal conditioning, could combine
overlapping observations instead of selecting among them and is a natural direction for future work.

\subsection{Ethical Considerations}
\label{sec:supp_ethics}

Hand imagery can act as a soft biometric: skin texture, vein-like patterns, and
nail features may be identifying, and a reconstructed avatar could in principle
support impersonation in hand-centric interfaces.
All training and evaluation data come from published research datasets used
under their licenses, and our model neither identifies individuals nor is
trained for recognition.
Feed-forward hand avatar reconstruction may benefit virtual reality,
telepresence, digital human animation, and human--computer interaction by
enabling efficient avatar creation from minimal input.
At the same time, as with other avatar and generative modeling technologies, reconstructed avatars could be
misused for identity impersonation or deceptive synthetic media.
Deployments that capture user hands should obtain informed consent and safeguard
reconstructed avatars as biometric-adjacent data.

\clearpage
\section{Additional Qualitative Results}
\label{sec:supp_qualitative}

\subsection{Qualitative Effect of the Soft Alpha Band}
\label{sec:supp_sab_qual}

\begin{figure}[!h]
    \centering
    \includegraphics[width=0.74\linewidth,height=0.45\textheight,keepaspectratio]{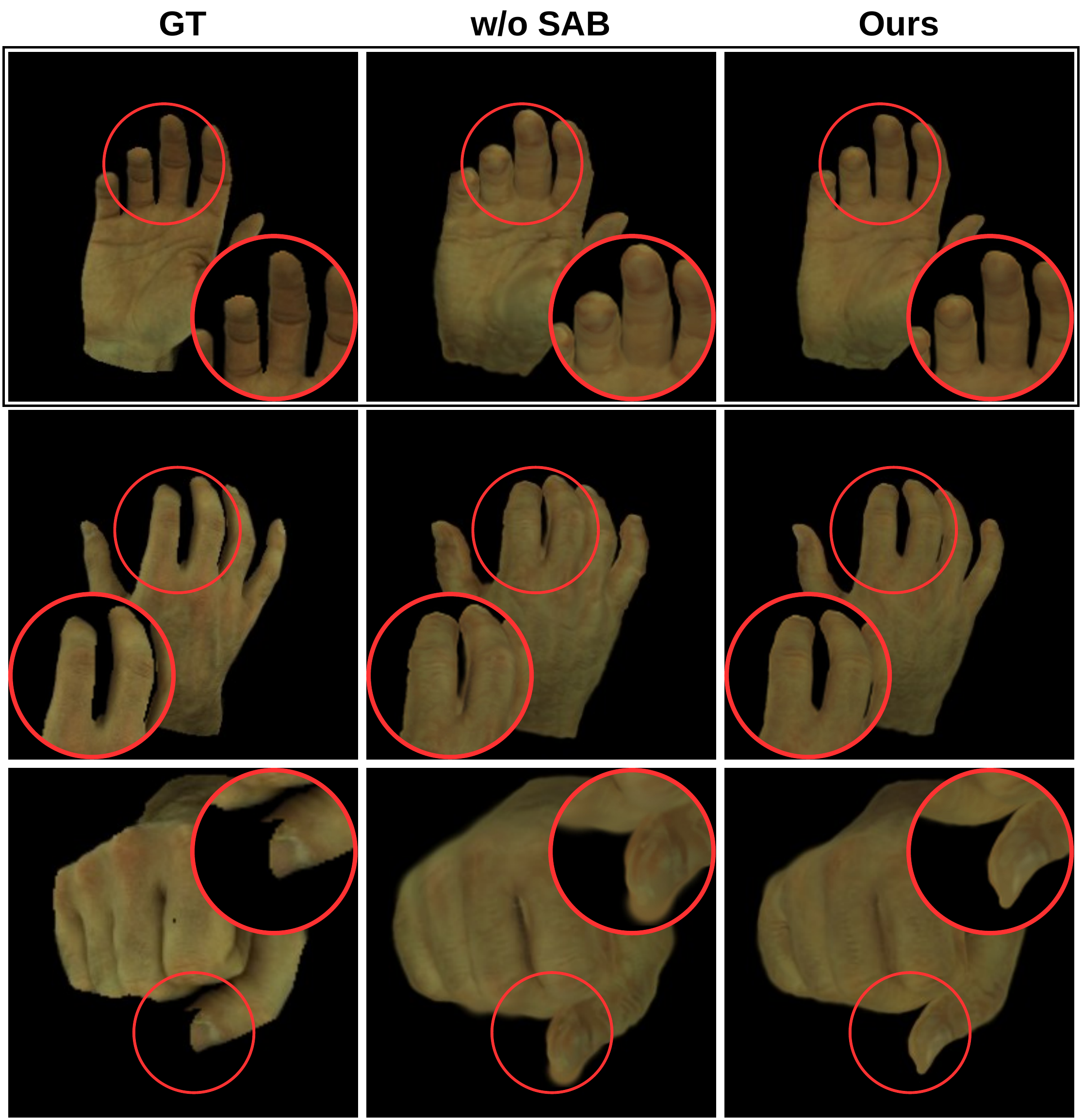}
    \caption{Qualitative effect of the Soft Alpha Band (SAB).
    Columns: ground truth (GT), without SAB (w/o SAB), and with SAB (Ours). The black box marks the input
    frame, and the remaining rows are novel poses. Red circles mark
    boundary regions magnified in the corner insets.}
    \label{fig:supp_sab_qual}
\end{figure}

Figure~\ref{fig:supp_sab_qual} visualizes the boundary-level failure mode diagnosed
in the main paper.
Since the ``w/o SAB'' model retains the same constrained parameterization, the
comparison isolates the effect of the regularizer.
Without SAB, residual leakage concentrates in the thin band around finger
boundaries and appears as blurred, apparently thickened contours, most visibly
where adjacent fingers meet (red circles), whereas the full model keeps these silhouettes sharp.
This qualitative gap is consistent with Table~\ref{tab:recipe_port} of the main
paper, where SAB raises the boundary F-score from $0.366$ to $0.698$ and more than
halves alpha leakage ($0.555\to0.234$), while interior coverage remains essentially
unchanged ($1.000$ vs.\ $0.999$).
SAB therefore suppresses leakage without eroding valid hand regions, rather than
improving the metrics by shrinking the silhouette.

\clearpage
\subsection{Qualitative Comparison with the Recipe Ports}
\label{sec:supp_recipe_ports_qual}

\begin{figure}[!h]
    \centering
    \includegraphics[width=0.95\linewidth,keepaspectratio]{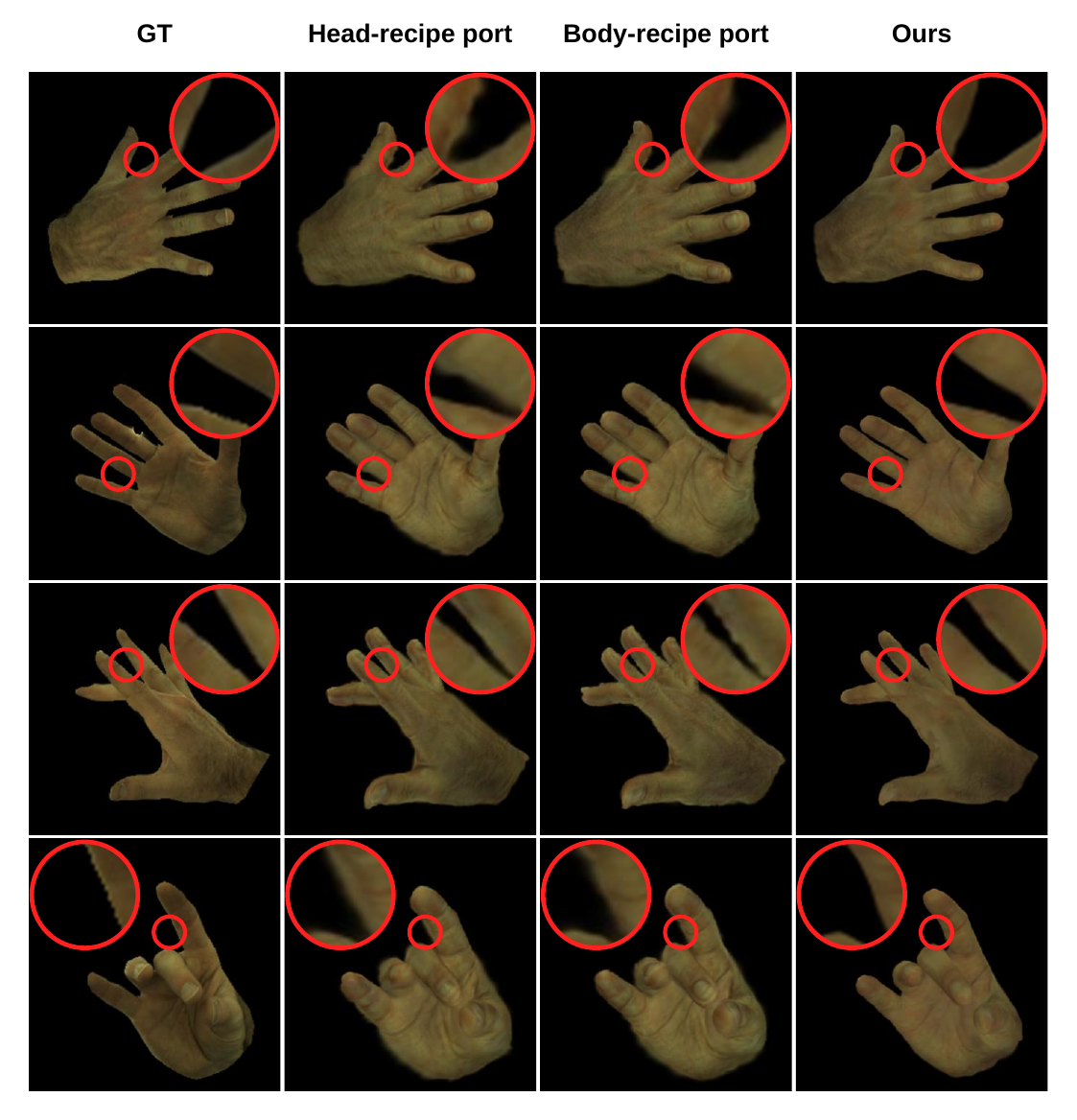}
    \caption{Qualitative comparison with the recipe ports of
    Table~\ref{tab:recipe_port} of the main paper.
    InterHand2.6M novel-pose/novel-view frames, with the source camera in the top row. Red circles mark
    inter-finger gaps and contours magnified in the corner insets.}
    \label{fig:supp_recipe_ports_qual}
\end{figure}

Figure~\ref{fig:supp_recipe_ports_qual} shows matched renderings of the two recipe
ports (Sec.~\ref{sec:supp_recipe_ports}) alongside ours, on four frames chosen to
expose inter-finger gaps and finger contours from the dorsal, palmar, and lateral sides.
Each red circle marks the same location in every column.
Both ports render the hand interior well, consistent with their interior coverage of
$1.000$, but blur the silhouette into a translucent fringe: the gaps between adjacent
fingers partially fill with alpha, and finger and thumb contours widen, most visibly
between the ring and little fingers (second row) and at the thumb--index junction
(top row).
The two ports look alike, in line with their close boundary F-scores of $0.299$ and
$0.267$.
Ours keeps the gaps open and the contours sharp on the same frames.
The constrained model without SAB, which lies between the ports and ours in
boundary F-score and leakage, is compared separately in Fig.~\ref{fig:supp_sab_qual}.

\clearpage
\subsection{Qualitative Self-reconstruction Diagnostics}
\label{sec:supp_selfrecon_qual}

\begin{figure}[!h]
    \centering
    \includegraphics[width=0.74\linewidth,height=0.50\textheight,keepaspectratio]{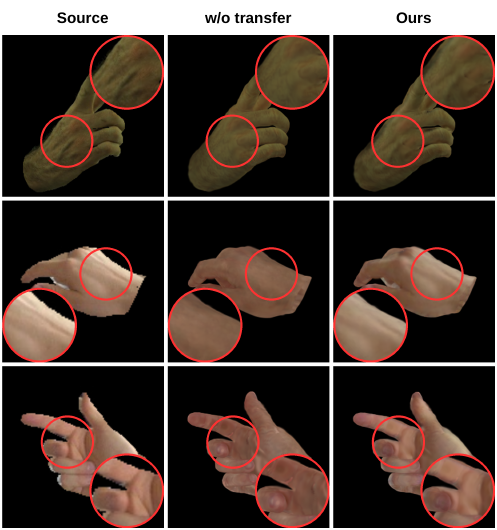}
    \caption{Qualitative source-view appearance diagnostic with geometry
    fixed (protocol in Sec.~\ref{sec:supp_selfrecon_protocol}).
    Columns: source observation (Source), Stage-1 base colors (w/o transfer), and
    composed colors (Ours). Red circles mark regions magnified in the insets.}
    \label{fig:supp_selfrecon_qual}
\end{figure}

Figure~\ref{fig:supp_selfrecon_qual} shows the diagnostic of
Table~\ref{tab:supp_selfrecon} on representative frames.
Because all non-color attributes are identical between w/o transfer and Ours, every
visible difference is attributable to the source-conditioned appearance module:
relative to the Stage-1 base colors, the full model recovers the observed skin tone and
source-observed local texture (circled regions), in line with the source-view gains
of Table~\ref{tab:supp_selfrecon}.

\clearpage
\subsection{Branch-level Diagnostics of the Appearance Module}
\label{sec:supp_branch_qual}

\begin{figure}[!h]
    \centering
    \includegraphics[width=\linewidth,keepaspectratio]{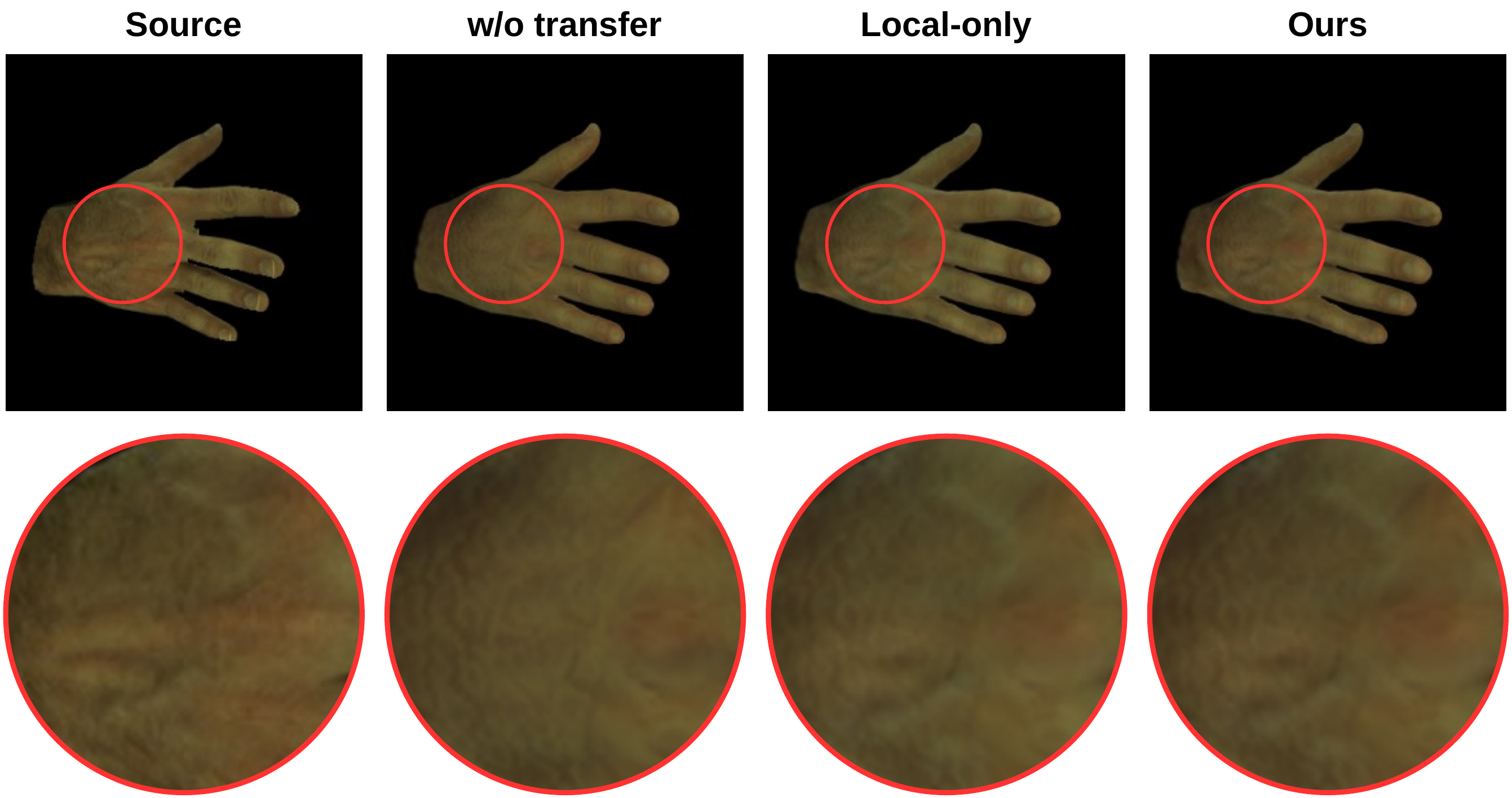}
    \caption{Branch-level analysis of source-conditioned local detail
    transfer.
    Columns: source observation (Source), Stage-1 base colors (w/o transfer), the
    local-only variant (Local-only), and the full model (Ours) under the same source pose and view. The red
    circle marks the region magnified below.}
    \label{fig:supp_branch_qual}
\end{figure}

Complementing the module-level evaluation above, we dissect the source-conditioned
appearance module into branch-isolated variants.
These are diagnostic probes of the appearance composition, not separate model configurations for the main
comparison.
The w/o transfer variant uses the Stage-1 base colors without
source-conditioned refinement.
The local-only variant uses Gaussian footprint-aligned local evidence and is designed
to transfer source-observed local details such as skin wrinkles and vein-like texture.
The full model combines footprint-aligned local observations,
rendering-contribution-based reliability, the global source identity, and the Stage-1 base colors.

Figure~\ref{fig:supp_branch_qual} compares the variants under the same source pose
and view, so that source-observed details can be traced from the input to each
branch.
The circled region on the dorsum contains subject-specific texture, such as
vein-like patterns and skin creases, that the w/o transfer variant attenuates.
The local-only branch and the full model both recover it from the source
observation and are nearly indistinguishable in this view, consistent with the
local-evidence pathway carrying source-observed detail in the full model.
The contribution of the global source identity cannot be seen from the source view, since it matters most for Gaussians that the source does not observe.
It appears instead on the prior-completed sides in the palm-facing renderings of
Fig.~\ref{fig:supp_animation} and the $360^\circ$ rotations of the supplementary video
(Sec.~\ref{sec:supp_videos}).
This branch analysis is qualitative, and its quantitative counterparts are Table~\ref{tab:ablation} of the
main paper and Table~\ref{tab:supp_selfrecon}.

\clearpage
\subsection{Generalization to Unseen Images under Fully Estimated Inputs}
\label{sec:supp_freihand}

\begin{figure}[!h]
    \centering
    \includegraphics[width=\linewidth,keepaspectratio]{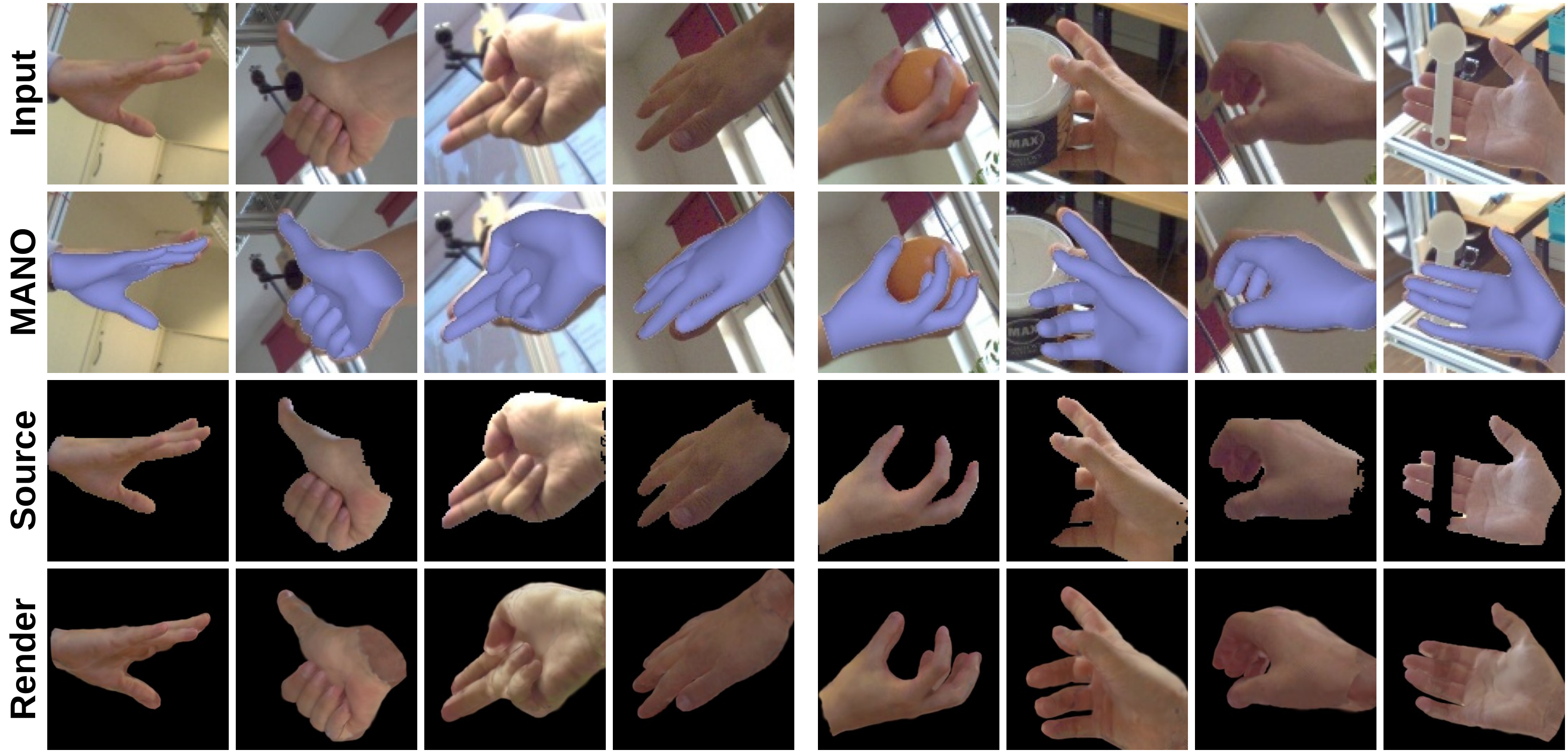}
    \caption{Qualitative generalization to unseen
    FreiHAND~\citep{Freihand2019} test images under fully estimated inputs.
    Rows: input frame (Input), WiLoR~\citep{potamias2025wilor} pose overlay (MANO),
    SAM 3~\citep{carion2026sam3} source crop (Source), and our render (Render). No ground-truth annotations
    or test-time optimization are used.}
    \label{fig:freihand_qualitative}
\end{figure}

We show source-view reconstructions of unseen FreiHAND~\citep{Freihand2019} test
images to evaluate generalization beyond the benchmark datasets of the main paper.
Unlike in the diagnostic of Sec.~\ref{sec:supp_selfrecon_protocol}, every input here is estimated.
Given only a single input image, our method reconstructs plausible canonical Gaussian
hand avatars in a fully feed-forward manner without subject-specific optimization or
test-time fitting.

For these results, MANO parameters and camera estimates are obtained using
WiLoR~\citep{potamias2025wilor}, and hand masks using SAM 3~\citep{carion2026sam3}.
The reconstructed avatars are rendered with the estimated camera parameters.
Figure~\ref{fig:freihand_qualitative} shows eight representative examples.
Each column documents every estimated input alongside the output, and FreiHAND's ground-truth MANO
annotations are not used.
Failure cases are shown in the same format in Fig.~\ref{fig:failure_cases} (Sec.~\ref{sec:supp_limitations}).
The left four columns show unoccluded hands, and the right four involve hand--object
interaction.
The examples cover side views, thin finger structures, articulated and near-closed
poses, and diverse skin tones.
The overlays also reveal that the estimated poses are imperfect, as fingertips visibly
deviate from the mesh in several columns.
Nevertheless, the reconstructed avatars remain faithful to the observed appearance.
In the rightmost column, the held object crosses the fingers near their bases and
splits the estimated mask into disjoint pieces.
The avatar still completes the masked-out finger segments with coherent appearance,
because appearance is anchored on the canonical MANO surface and reliability gating
falls back to the source-conditioned appearance prior where source evidence is missing.
Unlike the benchmark evaluations, these results use estimated rather than ground-truth pose, camera, and
mask inputs, so they illustrate the behavior of the full pipeline under real, uncontrolled upstream
estimation errors.
The corresponding quantitative comparison is Table~\ref{tab:ablation} of the main paper
and Table~\ref{tab:supp_sensitivity}(b).
Novel-pose animation of avatars reconstructed under the same fully estimated inputs is
shown in Fig.~\ref{fig:supp_animation} (Sec.~\ref{sec:supp_animation}).

\clearpage
\subsection{Novel-pose Animation from a Single Image}
\label{sec:supp_animation}

\begin{figure}[!h]
    \centering
    \includegraphics[width=\linewidth,height=0.78\textheight,keepaspectratio]{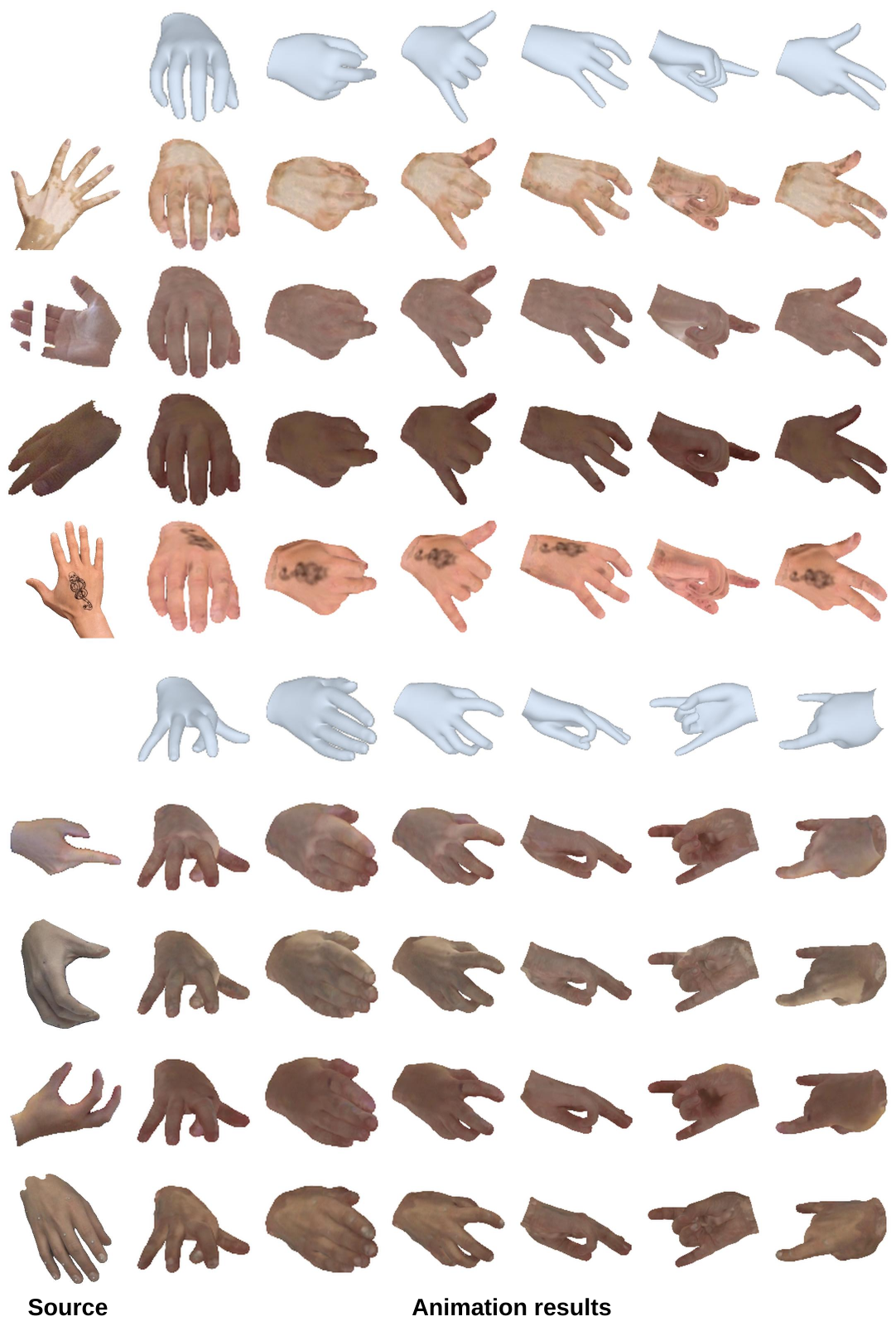}
    \caption{Novel-pose animation from a single image.
    In each block, the top row shows the driving MANO poses and each subject row the
    source image followed by our renderings in those poses. All inputs are fully estimated. Best viewed zoomed in.}
    \label{fig:supp_animation}
\end{figure}

Figure~\ref{fig:freihand_qualitative} shows how faithfully an avatar reproduces its own
source view, whereas Fig.~\ref{fig:supp_animation} shows whether avatars reconstructed
in the same way remain consistent once they are driven away from it.
Eight subjects are reconstructed with the fully estimated pipeline of
Sec.~\ref{sec:supp_freihand}, using WiLoR~\citep{potamias2025wilor} pose and camera and a
SAM 3~\citep{carion2026sam3} mask, with no ground truth and no test-time optimization.
From top to bottom, their sources are an in-the-wild image released with
OHTA~\citep{zheng2024ohta}, FreiHAND~\citep{Freihand2019}, FreiHAND, a
HARP~\citep{karunratanakul2023harp} subject excluded from the training split, FreiHAND,
ARCTIC~\citep{fan2023arctic} (not part of the training data), FreiHAND, and ARCTIC.
The subjects are split into two blocks of four, and each block is driven by its own six poses, rendered
from a camera fixed in the MANO frame.
The two blocks share no poses.
Within a block, every column therefore shows all four subjects in the same pose and from the same
viewpoint.
The top row of each block renders the driving MANO mesh from the same camera, and the
second block includes palm-facing views.

Subject-specific texture is carried across poses without drifting toward a generic
hand.
The vitiligo pattern (row~1), the tattoo (row~4), and, more faintly, the motion-capture
markers (row~8) stay attached to their anatomical locations under finger articulation and global rotation, including on
the wrist and on fingers that are foreshortened or partially hidden in the source view.
Fine detail is preserved only coarsely, as the line work of the tattoo is blurred and the markers appear as
small light dots that are best seen in the supplementary video (Sec.~\ref{sec:supp_videos}).
Surfaces that are unobserved in the source are completed by the source-conditioned prior
and remain coherent across views, as the palm-facing columns of the second block show
for dorsal-side inputs.
The second row deliberately uses the palm-facing input of the rightmost column of
Fig.~\ref{fig:freihand_qualitative}, whose SAM 3 mask is corrupted with missing finger
segments.
We use the mask as is, so the dorsal appearance of this avatar is entirely completed from the prior.
The avatar nevertheless remains animatable and coherent, which complements the mask-failure analysis of
Sec.~\ref{sec:supp_limitations}.
No per-frame refinement is involved, since all Gaussian attributes are fixed after the single feed-forward
pass.
The identity consistency across columns is therefore a property of the canonical representation rather
than of per-pose fitting.
The same eight avatars are shown under continuous driving in the supplementary video, and
four of them are also shown in full $360^\circ$ rotation (Sec.~\ref{sec:supp_videos}).

\section{Failure Cases and Limitations}
\label{sec:supp_limitations}

\begin{figure}[!h]
    \centering
    \includegraphics[width=0.9\linewidth,height=0.5\textheight,keepaspectratio]{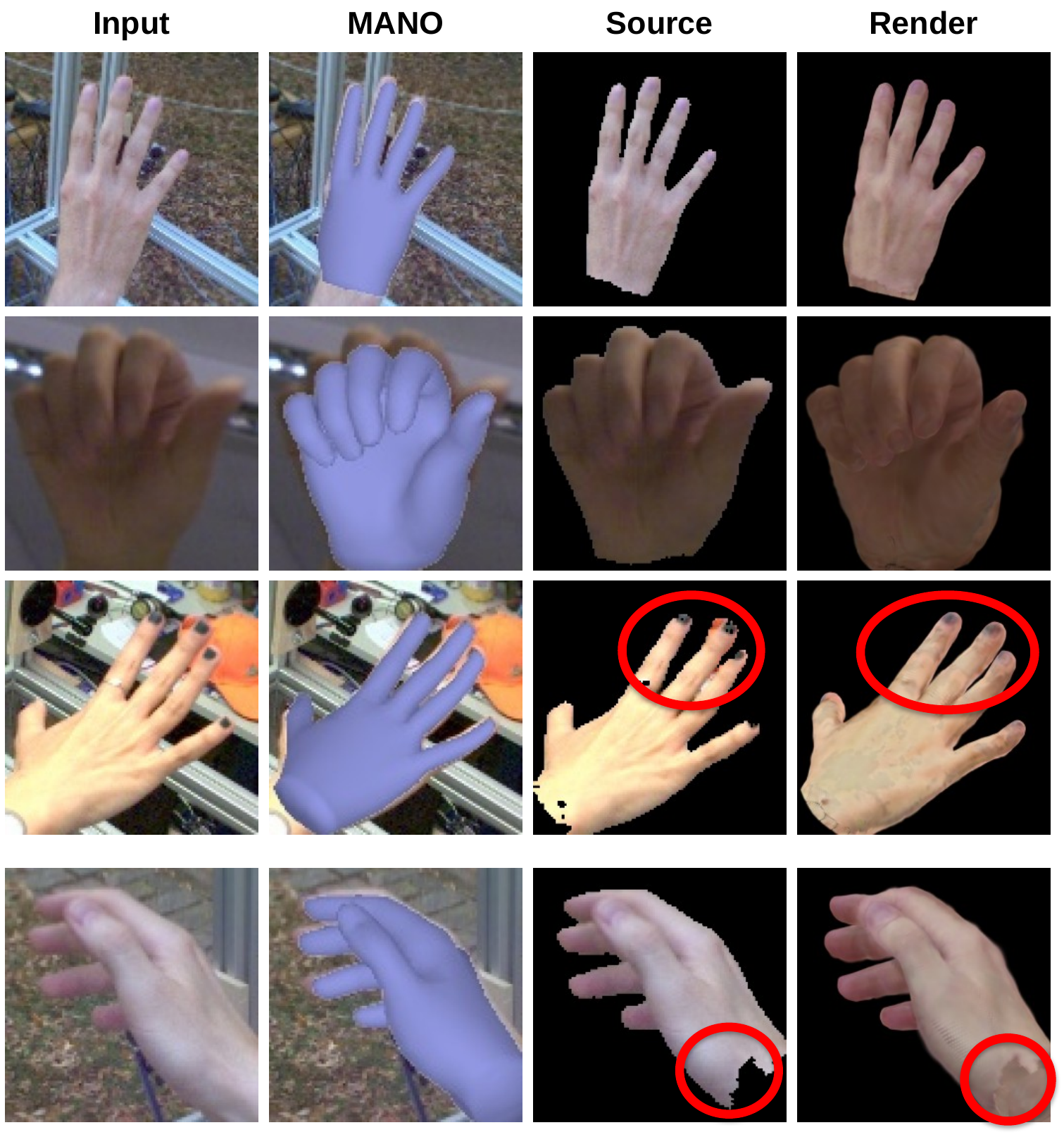}
    \caption{Representative appearance failure cases.
    Columns: Input, MANO, Source, and Render (the row labels of
    Fig.~\ref{fig:freihand_qualitative}). Red circles mark the affected regions in the third and bottom
    rows.}
    \label{fig:failure_cases}
\end{figure}

Representative failure cases are shown in Fig.~\ref{fig:failure_cases}, in the
input-documentation format of Fig.~\ref{fig:freihand_qualitative}: the input
frame (Input), the WiLoR~\citep{potamias2025wilor} pose overlay (MANO), the
SAM 3~\citep{carion2026sam3} source crop (Source), and our reconstruction (Render).

First, our method inherits the inherent ambiguity of single-image hand
reconstruction. Under severe self-occlusion the occluded surface is never observed,
so its appearance is completed from the source-conditioned prior rather than
transferred from evidence, and thin local structures that are only weakly observed
are reconstructed with reduced fidelity.
The upstream pose estimate on which the avatar is anchored also becomes less reliable in such views.
Where the boundary of the observed set crosses a thin or strongly foreshortened part,
such as the thumb in a dorsal view, source-driven and prior-completed colors meet on
the same structure.
If the observed tone differs from the prior, as when shading is baked into the source (the third point
below), the transition appears as a visible seam rather than a gradual blend.
Sec.~\ref{sec:supp_reliability} discusses this boundary, and Sec.~\ref{sec:supp_canonical_diag} measures
where such seams can occur (Fig.~\ref{fig:supp_color_reliability}).

Second, all Gaussian attributes are fixed after the single forward pass.
This keeps the identity consistent across poses, but the avatar does not model pose-dependent appearance.
Wrinkles that form and shading that changes under strong articulation are not synthesized, whereas
wrinkles visible in the source are transferred as texture.
The representation also ends where the MANO template ends, and the avatar has no forearm.
When the hand is viewed along the forearm, as in the $360^\circ$ rotations of the supplementary video, the
open wrist boundary is visible, including the inner side of the template surface on which the Gaussians
lie.
The MANO-based baselines likewise end at the wrist.
The anchor set is fixed as well, and the isotropic scales are predicted once in the
canonical pose.
Under strong flexion, skinning stretches the surface on the convex side of a joint and
spreads neighboring anchors apart, but the splats are not enlarged to follow this
stretch.
Where the spacing exceeds the splat footprint, neighboring rows of splats separate, and
the lattice becomes visible as fine striations at the knuckles.
Enlarging the splats relative to the anchor spacing would close these gaps, but the
same splats also form the silhouette.
Rendering the prediction on the next subdivision level, which halves the anchor spacing, without shrinking
the scales (row ``$k=4$, $s_i$'' of Table~\ref{tab:supp_densify}) doubles the footprint-to-spacing ratio.
This raises alpha leakage from $0.234$ to $0.293$ and lowers the boundary F-score from $0.698$ to $0.661$,
which is exactly the leakage that the scale ceiling and SAB are designed to suppress.
The striations are a local effect, and since curvature is handled by density rather than by anisotropy
(Sec.~\ref{sec:supp_attr_ablation}), a finer subdivision level reduces them.
Training at $k=4$, however, runs out of memory on our training GPUs (RTX A6000, 48\,GB) and was not
feasible with our resources.

\begin{figure}[!tb]
    \centering
    \includegraphics[width=0.7\linewidth,height=0.40\textheight,keepaspectratio]{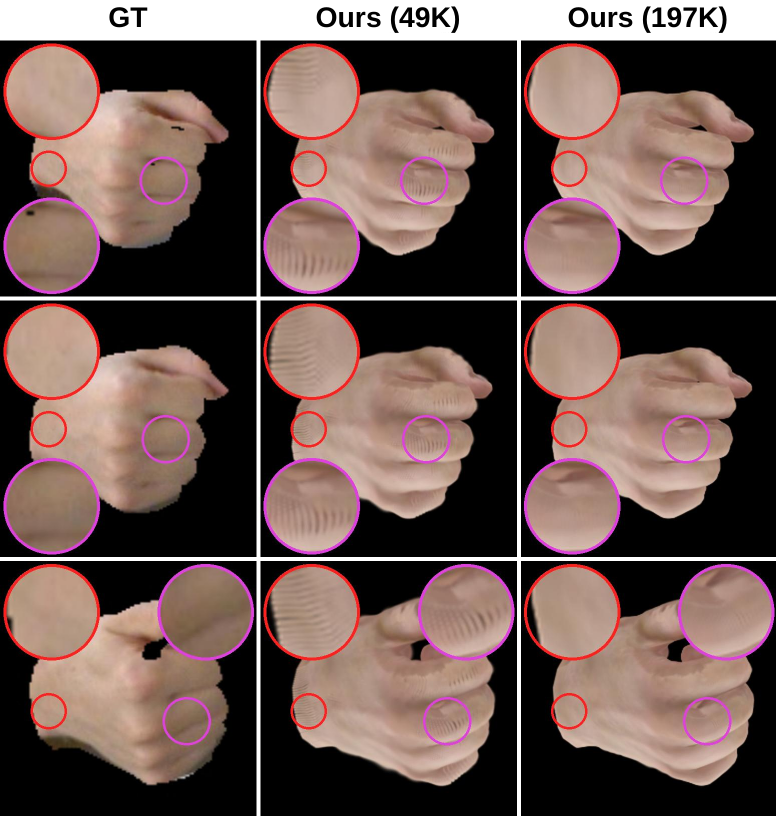}
    \caption{Inference-time densification.
    Columns: ground truth, the $k=3$ avatar used throughout the paper (49K
    Gaussians), and the same prediction densified onto the $k=4$ anchors (197K, scales $0.6\,s_i$,
    Table~\ref{tab:supp_densify}). Rows: three HanCo evaluation frames.
    Magenta circles mark knuckles and red circles the wrist, each magnified in the matching corner inset.}
    \label{fig:supp_densify}
\end{figure}

These striations can nevertheless be largely removed at inference without retraining.
We keep the trained $k=3$ network and its 49{,}281 predicted Gaussians and place Gaussians on all
196{,}993 anchors of the next subdivision level ($k=4$, continuing Table~\ref{tab:supp_mano_subdivision}).
The 49{,}281 anchors shared with $k=3$ keep their predicted attributes, and each of the 147{,}712 new
anchors lies at the midpoint of a $k=3$ edge and takes the average of all predicted attributes of the two
endpoints, including the composed colors $\mathbf{c}^{*}_{i}$.
Since the anchor spacing halves, every scale $s_i$ is then multiplied by a single fixed factor, and the
$k=4$ Gaussians are skinned with the $k=4$ weights and rendered as usual.
This densification is a fixed post-processing step on the predicted Gaussians.
It involves no per-subject or per-image optimization and uses neither the source image nor ground truth.
Table~\ref{tab:supp_densify} reports the sweep over the factor.
Each of $0.5\,s_i$, $0.6\,s_i$, and $0.7\,s_i$ improves LPIPS, boundary F, and alpha leakage on both
benchmarks, while PSNR changes by $-0.04$ to $-0.06$\,dB on InterHand2.6M and by $+0.12$ to $+0.30$\,dB on
HanCo, so the conclusion does not depend on the exact factor within this range.
We use $0.6\,s_i$, chosen from the InterHand2.6M results without using HanCo, for Fig.~\ref{fig:supp_densify}
and the supplementary video (Part~4, Sec.~\ref{sec:supp_videos}).
Figure~\ref{fig:supp_densify} shows three HanCo evaluation frames spanning the fist at the start of the
sequence, selected by pose rather than by any metric.
The jagged ground-truth boundary comes from the HanCo mask.
In the $k=3$ rendering, the anchor lattice appears as striations at the knuckles and near the wrist, where
the avatar ends at the MANO template boundary, and the densified rendering removes most of them.
Feeding the $k=4$ anchors to the network directly instead worsens all five metrics of
Table~\ref{tab:supp_densify} on both benchmarks, so the gain comes from the rendering density rather than
from the network.
The overhead is small.
Measured in the same run as the $k=3$ avatar, the rendering speed stays within 1\%, the Gaussian parameters
grow from 3.4\,MB to 13.4\,MB, and the GPU memory added by rendering a frame grows from 13\,MB to 38\,MB.
For comparison, the rendering peak of Sec.~\ref{sec:supp_timing}, which also includes the CUDA context and
the network, is 1.79\,GB.
Faint striations can remain at the most strongly flexed knuckles (bottom row of Fig.~\ref{fig:supp_densify}),
and the factor is a single global constant rather than adapted to the local anchor spacing.
All other results in this paper use the $k=3$ avatar.

\begin{table}[!tb]
    \centering
    \caption{Inference-time densification (Sec.~\ref{sec:supp_limitations}).
    The same trained $k=3$ network is used throughout.
    ``$k=4$ input'' feeds the $k=4$ anchors to the network.
    The other $k=4$ rows interpolate the $k=3$ prediction onto the $k=4$ anchors and multiply every scale
    $s_i$ by the stated factor, and ``$k=4$, $s_i$'' keeps the scales unchanged.
    Boundary metrics follow Sec.~\ref{sec:supp_boundary_metrics}.
    Bold marks the setting used for Fig.~\ref{fig:supp_densify} and the supplementary video.}
    \label{tab:supp_densify}
    \small
    \setlength{\tabcolsep}{3.5pt}
    \resizebox{\linewidth}{!}{%
    \begin{tabular}{@{}lccccc@{\hspace{10pt}}ccccc@{}}
    \toprule
    & \multicolumn{5}{c}{InterHand2.6M} & \multicolumn{5}{c}{HanCo (zero-shot)} \\
    \cmidrule(lr){2-6}\cmidrule(l){7-11}
    Setting & PSNR$\uparrow$ & SSIM$\uparrow$ & LPIPS$\downarrow$ & Bnd.~F$\uparrow$ & Leak.$\downarrow$
            & PSNR$\uparrow$ & SSIM$\uparrow$ & LPIPS$\downarrow$ & Bnd.~F$\uparrow$ & Leak.$\downarrow$ \\
    \midrule
    $k=3$ (Tables~\ref{tab:main_quant},~\ref{tab:hanco_results}) & 26.21 & 0.866 & 0.133 & 0.698 & 0.234 & 20.18 & 0.836 & 0.188 & 0.617 & 0.314 \\
    $k=4$ input      & 25.98 & 0.862 & 0.137 & 0.671 & 0.247 & 19.75 & 0.829 & 0.190 & 0.559 & 0.373 \\
    \midrule
    $k=4$, $s_i$         & 25.86 & 0.862 & 0.136 & 0.661 & 0.293 & 19.68 & 0.831 & 0.189 & 0.542 & 0.381 \\
    $k=4$, $0.85\,s_i$   & 26.06 & 0.864 & 0.132 & 0.697 & 0.244 & 20.04 & 0.837 & 0.184 & 0.613 & 0.321 \\
    $k=4$, $0.7\,s_i$    & 26.15 & 0.865 & 0.130 & 0.711 & 0.205 & 20.30 & 0.840 & 0.181 & 0.668 & 0.269 \\
    \textbf{$\boldsymbol{k=4}$, $\boldsymbol{0.6\,s_i}$} & \textbf{26.17} & \textbf{0.865} & \textbf{0.129} & \textbf{0.708} & \textbf{0.185} & \textbf{20.41} & \textbf{0.842} & \textbf{0.179} & \textbf{0.693} & \textbf{0.239} \\
    $k=4$, $0.5\,s_i$    & 26.15 & 0.863 & 0.129 & 0.701 & 0.168 & 20.48 & 0.842 & 0.178 & 0.708 & 0.215 \\
    \bottomrule
    \end{tabular}}
\end{table}

Third, Fig.~\ref{fig:failure_cases} collects appearance failures that arise while the
hand itself is fully observed.
These failures fall into two regimes, which the input columns separate explicitly.
In the first regime (top three rows), the Source crop retains the evidence that the
failure concerns, so the failure lies in the appearance composition itself.
The top two rows share one cause.
Illumination and intrinsic appearance are entangled in the composed colors, because shading baked into the
source is carried as albedo rather than separated from it (Sec.~\ref{sec:supp_viewdep}).
This entanglement has two signatures.
A strongly overexposed, washed-out source is rendered with a more saturated skin tone than observed, and an
underexposed tone is absorbed into the albedo and, since all attributes are fixed after the forward pass,
carried to every novel pose.
These two rows are left unmarked because the illumination shift affects the whole hand.
The third row concerns a rare cosmetic attribute that the transfer does not carry.
Dark nail pigment inside the source crop is composed as diffuse skin discoloration around the fingertips
and nails rather than reproduced as painted nails.
The nails in the source crop and the fingertip blotches they become are circled at the same image location
in the Source and Render columns.

In the second regime (bottom row), the failure is inherited from a corrupted mask input, consistent with
our assumption of externally provided segmentation.
The source crop is torn at the wrist, and evidence along the tear boundary leaks into the composed
appearance (the circles mark the wrist tear and the resulting brown discoloration).
The contrast with the rightmost column of Fig.~\ref{fig:freihand_qualitative} delimits where reliability
gating operates.
When fragmentation removes a region outright, reliability gating routes it to the source-conditioned
appearance prior and the region is completed coherently, as the second row of Fig.~\ref{fig:supp_animation}
shows for one such fragmented input.
Evidence that survives at a defect boundary, however, is locally indistinguishable from valid observation
and is accepted as such.
In all cases shown, the MANO-anchored geometry remains plausible, the torn wrist silhouette of the bottom
row is itself completed by the anchors, and the failures are confined to the appearance branch.

Separating illumination from intrinsic appearance is left for future work
(Sec.~\ref{sec:supp_viewdep}).
The cosmetic-attribute failure reflects the limited coverage of such attributes in
existing hand avatar datasets, and the mask-inherited failure diminishes as upstream segmentation improves.
Extending feed-forward Gaussian hand avatars to more diverse real-world conditions
and hand--object interaction scenarios remains an important direction for future work.

\section{Additional Analyses}
\label{sec:supp_analyses}

This section analyzes the evaluation results along three axes that aggregate metrics do not separate,
namely pose difficulty, temporal stability under continuous driving, and the reliability, opacity, and
boundary continuity of the canonical avatar.
The protocols are given in Sec.~\ref{sec:supp_diag_protocols}, and the pose-stratified and temporal
numbers use the same renders and ground truth as Tables~\ref{tab:main_quant} and~\ref{tab:hanco_results}
of the main paper.

\subsection{Pose-stratified Evaluation}
\label{sec:supp_pose_strata}

\begin{table}[!tb]
    \centering
    \caption{Pose-stratified evaluation of our method. Terciles follow Sec.~\ref{sec:supp_diag_protocols}, and ranges are in degrees.}
    \label{tab:supp_pose_strata}
    \small
    \setlength{\tabcolsep}{3.5pt}
    \begin{tabular}{@{}lcccccccc@{}}
    \toprule
    & \multicolumn{4}{c}{InterHand2.6M} & \multicolumn{4}{c}{HanCo (zero-shot)} \\
    \cmidrule(lr){2-5}\cmidrule(l){6-9}
    Tercile & Range & PSNR$\uparrow$ & SSIM$\uparrow$ & LPIPS$\downarrow$
            & Range & PSNR$\uparrow$ & SSIM$\uparrow$ & LPIPS$\downarrow$ \\
    \midrule
    \multicolumn{9}{@{}l}{\emph{Palm-orientation change}} \\
    Easy    & 0--79    & 26.37 & 0.863 & 0.133 & 0--82    & 20.67 & 0.842 & 0.169 \\
    Medium  & 80--115  & 25.67 & 0.858 & 0.140 & 82--128  & 19.70 & 0.826 & 0.212 \\
    Extreme & 116--179 & 26.61 & 0.876 & 0.126 & 129--180 & 20.15 & 0.840 & 0.183 \\
    \midrule
    \multicolumn{9}{@{}l}{\emph{Finger-articulation change}} \\
    Easy    & 0--11  & 27.37 & 0.888 & 0.106 & 0--22  & 20.89 & 0.851 & 0.177 \\
    Medium  & 11--17 & 26.56 & 0.875 & 0.119 & 22--43 & 20.67 & 0.845 & 0.174 \\
    Extreme & 17--56 & 24.68 & 0.834 & 0.175 & 43--56 & 18.96 & 0.812 & 0.213 \\
    \midrule
    All     &        & 26.21 & 0.866 & 0.133 &        & 20.18 & 0.836 & 0.188 \\
    \bottomrule
    \end{tabular}
\end{table}

\begin{table}[!tb]
    \centering
    \caption{Articulation-stratified comparison on the finger-articulation
    terciles of Table~\ref{tab:supp_pose_strata}.}
    \label{tab:supp_articulation_methods}
    \small
    \setlength{\tabcolsep}{3.2pt}
    \begin{tabular}{@{}lcccccccc@{}}
    \toprule
    & \multicolumn{4}{c}{PSNR$\uparrow$} & \multicolumn{4}{c}{LPIPS$\downarrow$} \\
    \cmidrule(lr){2-5}\cmidrule(l){6-9}
    Tercile & Ours & OHTA & HandAvatar & BIGS$^{*}$ & Ours & OHTA & HandAvatar & BIGS$^{*}$ \\
    \midrule
    \multicolumn{9}{@{}l}{\emph{InterHand2.6M}} \\
    Easy    & 27.37 & 27.23 & 24.90 & 21.30 & 0.106 & 0.103 & 0.150 & 0.168 \\
    Medium  & 26.56 & 26.43 & 24.21 & 20.38 & 0.119 & 0.115 & 0.163 & 0.191 \\
    Extreme & 24.68 & 24.66 & 22.24 & 19.67 & 0.175 & 0.169 & 0.223 & 0.277 \\
    \midrule
    \multicolumn{9}{@{}l}{\emph{HanCo (ours zero-shot, baselines fitted to the source frame)}} \\
    Easy    & 20.89 & 20.12 & 19.88 & 18.60 & 0.177 & 0.177 & 0.194 & 0.224 \\
    Medium  & 20.67 & 20.53 & 20.77 & 17.99 & 0.174 & 0.172 & 0.185 & 0.228 \\
    Extreme & 18.96 & 19.53 & 20.32 & 16.51 & 0.213 & 0.209 & 0.216 & 0.287 \\
    \bottomrule
    \end{tabular}
\end{table}

Table~\ref{tab:supp_pose_strata} splits both benchmarks into equal-count terciles of
the palm-orientation and finger-articulation change from the source pose.
On both benchmarks, the evaluation subject is unseen during training, and HanCo is also
unseen as a dataset (Sec.~\ref{sec:supp_training_datasets}).
The terciles therefore show how the quality of an avatar reconstructed from a single
image of an unseen hand, without test-time optimization, changes as the pose and view
move away from that image.
The effect of palm orientation is non-monotonic.
On both datasets, the middle tercile is the lowest, $0.70$ and $0.97$\,dB below the easiest on
InterHand2.6M and HanCo, respectively.
In this tercile, the median viewing direction has turned by $84^\circ$ relative to the hand on both
datasets, so the hand is seen nearly edge-on.
The extreme tercile is viewed mostly from the side that the dorsal source images do not observe, with a
median viewing change of $106^\circ$ on InterHand2.6M and $121^\circ$ on HanCo.
This tercile is $0.24$\,dB above the easiest on InterHand2.6M and $0.52$\,dB below it on HanCo.
Finger articulation has a larger effect, concentrated in the extreme tercile.
Relative to the easiest tercile, quality decreases by $0.81$ and $0.22$\,dB in the middle tercile on
InterHand2.6M and HanCo, respectively, and by $2.69$ and $1.93$\,dB in the extreme tercile.

Table~\ref{tab:supp_articulation_methods} scores all methods on the same articulation terciles of both
datasets.
On InterHand2.6M, the per-frame renders of the baselines average to their numbers in
Table~\ref{tab:main_quant} of the main paper within $0.01$\,dB.
On HanCo, our avatar is reconstructed zero-shot, whereas each baseline is optimized on the source frame of
the subject (Table~\ref{tab:hanco_results} of the main paper).
In LPIPS, the extreme articulation tercile is the hardest for every method on both datasets, not only for
ours.
On InterHand2.6M, all four methods also lose PSNR from the easiest to the extreme tercile, by
$1.63$--$2.69$\,dB, and ours remains the highest in PSNR in every tercile.
On HanCo, BIGS$^{*}$ and ours lose $2.09$ and $1.93$\,dB, whereas OHTA and HandAvatar change by $-0.59$ and
$+0.44$\,dB.
The PSNR gain of HandAvatar is not matched by LPIPS, which worsens from $0.194$ to $0.216$ over the same
terciles.
Its renderings have a uniform tone without shading (Sec.~\ref{sec:supp_viewdep}) and change less than the
footage over time (Sec.~\ref{sec:supp_temporal}).
Such smooth renderings keep the squared pixel error low even when the appearance of the hand changes, so
PSNR can rise while LPIPS worsens.
In the extreme tercile, ours is below OHTA and HandAvatar in PSNR, at $18.96$\,dB against $19.53$ and
$20.32$\,dB.
Our larger drop on HanCo is consistent with the limitation that our colors are fixed after the forward pass,
so wrinkles and shading that appear under strong articulation are not synthesized
(Sec.~\ref{sec:supp_limitations}).
These data, however, do not isolate the cause, since on InterHand2.6M OHTA and HandAvatar, which model
self-occlusion shading (Sec.~\ref{sec:supp_viewdep}), lose about as much PSNR as ours ($2.57$ and
$2.66$\,dB).

\subsection{Temporal Stability}
\label{sec:supp_temporal}

\begin{table}[!tb]
    \centering
    \caption{Excess temporal variation on HanCo
    (Eq.~\ref{eq:supp_temporal}, 75 consecutive-frame pairs per camera). Ours is zero-shot,
    and the baselines are fitted to the source frame. GT change:
    mean change between consecutive ground-truth frames. Method columns give $E_t$ relative to it, where 0
    means as much change as the footage and positive values mean additional variation.
    cam6 is the source view, and the last row is the mean of the absolute per-camera values.}
    \label{tab:supp_temporal}
    \small
    \setlength{\tabcolsep}{5pt}
    \begin{tabular}{@{}lccccc@{}}
    \toprule
    Camera & GT change & Ours & OHTA & HandAvatar & BIGS$^{*}$ \\
    \midrule
    cam3               & 0.0362 & $+1.1\%$ & $-5.1\%$ & $-17.6\%$ & $+37.8\%$ \\
    cam5               & 0.0319 & $+7.3\%$ & $+5.0\%$ & $-8.7\%$  & $+50.5\%$ \\
    cam6 (source view) & 0.0360 & $+2.6\%$ & $-2.2\%$ & $-23.3\%$ & $+30.7\%$ \\
    cam7               & 0.0412 & $+8.0\%$ & $+2.2\%$ & $-15.6\%$ & $+34.9\%$ \\
    \midrule
    Mean of $\lvert\cdot\rvert$ & 0.0363 & $4.7\%$ & $3.6\%$ & $16.3\%$ & $38.5\%$ \\
    \bottomrule
    \end{tabular}
\end{table}

Table~\ref{tab:supp_temporal} reports temporal stability on the continuous HanCo sequence.
Our avatar is evaluated zero-shot, as neither the subject nor the dataset is seen during training and no
test-time optimization is performed.
Every time step except that of the source frame shows a novel pose, and three of the four cameras show a
novel view.
Under these conditions, our renderings remain within $8.0\%$ of the ground-truth change on every camera,
with a mean absolute excess of $4.7\%$.
To calibrate this value, we add a per-frame brightness jitter of $\pm1\%$ to our own renderings.
The excess pooled over all pairs then increases from $+4.8\%$ to $+12.4\%$, and further to $+22.0\%$ and
$+43.1\%$ for $\pm2\%$ and $\pm4\%$.
A $1\%$ flicker thus adds $7.6$ points, which exceeds our mean absolute excess of $4.7\%$.
By contrast, the per-subject 3DGS fit BIGS$^{*}$ exceeds the ground-truth change by $30.7$--$50.5\%$ across
the four cameras, which is consistent with view-specific appearance noise from single-view optimization.
OHTA, which optimizes its avatar on this subject for 3360\,s (Table~\ref{tab:hanco_results} of the main
paper), is closest to zero, with a mean absolute excess of $3.6\%$, although its sign differs across cameras.
Its stability is therefore comparable to that of our feed-forward avatar.
HandAvatar changes less than the footage on every camera, by $8.7$--$23.3\%$, consistent with the uniform
tone without shading noted in Sec.~\ref{sec:supp_viewdep}.
Because the canonical Gaussians are fixed after the single forward pass, the representation itself cannot
drift, and the remaining variation arises from the pose-driven deformation and the rendering.

\subsection{Reliability, Opacity, and Boundary Continuity on the Canonical Avatar}
\label{sec:supp_canonical_diag}

\begin{figure}[!tb]
    \centering
    \includegraphics[width=\linewidth,height=0.50\textheight,keepaspectratio]{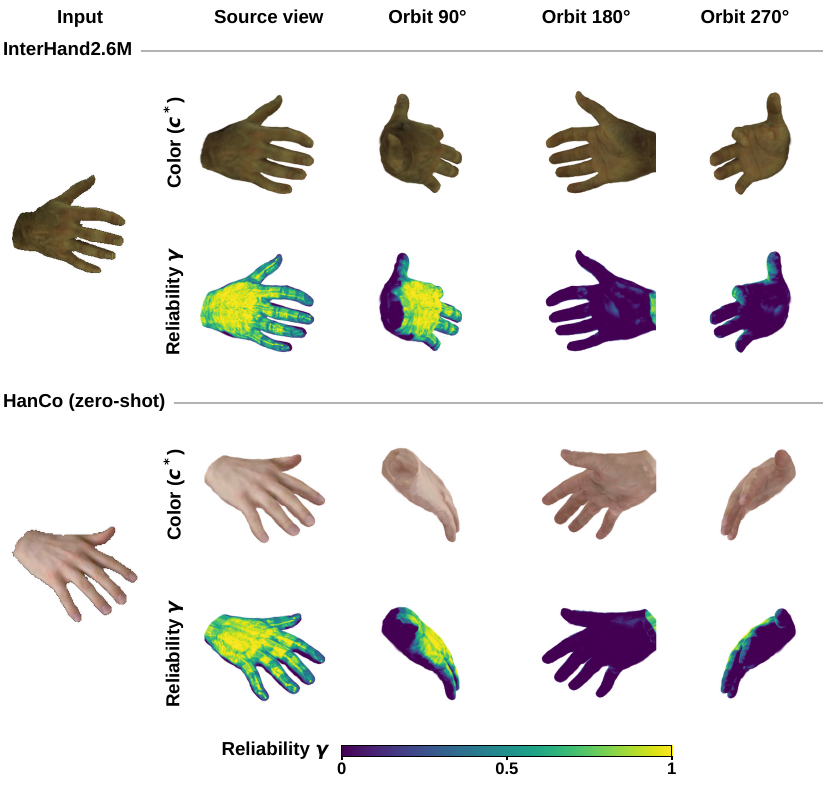}
    \caption{Canonical color and observation reliability. Final colors
    $\mathbf{c}^*$ (top) and reliability $\gamma$ (bottom) of each canonical avatar,
    rendered from the source view and from an orbit of the source camera about the
    hand, with the same viewpoints and crops in both rows.}
    \label{fig:supp_color_reliability}
\end{figure}

\begin{figure}[!tb]
    \centering
    \includegraphics[width=\linewidth,height=0.40\textheight,keepaspectratio]{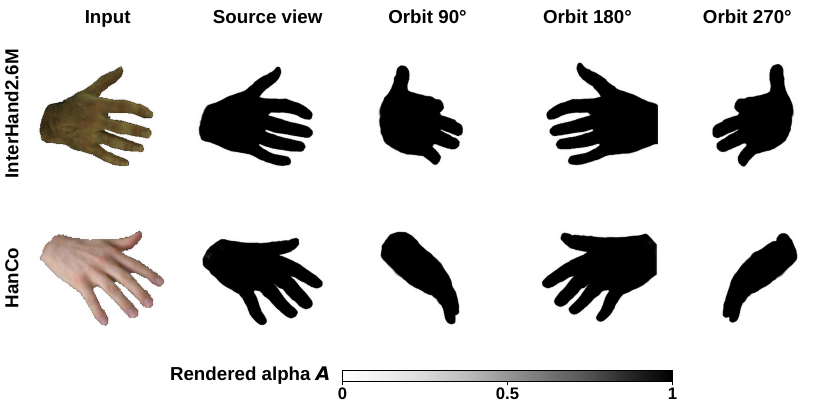}
    \caption{Rendered opacity (accumulated alpha $A$) of the canonical avatars
    from the views of Fig.~\ref{fig:supp_color_reliability}. The straight cut in the
    $180^\circ$ view is the wrist opening of the MANO template.}
    \label{fig:supp_opacity}
\end{figure}

Figure~\ref{fig:supp_color_reliability} shows that $\gamma$ is high on the dorsal surface visible in the
input, decreases at grazing angles, and is exactly zero on the side that the input does not observe.
Specifically, $55.1\%$ of the Gaussians on InterHand2.6M and $61.6\%$ on HanCo have $\gamma_i=0$, and these
Gaussians are colored by the Stage-1 prior and the source-level shift alone.
Because $\boldsymbol{\Delta}^{aid}_{s}$ is shared by all Gaussians, the color update of these Gaussians is
spatially uniform, with a standard deviation below $10^{-7}$ across Gaussians.
The module therefore cannot create a color boundary inside the unobserved region, and a seam between
transferred and prior-completed colors can arise only where observed and unobserved Gaussians meet.

At the reliability threshold $\gamma_i=0.5$, the color step is small.
The median $\Delta E$ between neighbors on either side of the threshold is $1.75$ on InterHand2.6M and
$1.91$ on HanCo, which is $1.05\times$ and $1.32\times$ the variation between neighbors inside the
high-reliability set ($1.67$ and $1.45$) and below the commonly cited just-noticeable difference of about
$2.3$.
The step is larger where a high-reliability Gaussian directly neighbors a low-reliability one without a
transition band.
This occurs for 79 pairs on InterHand2.6M and 209 pairs on HanCo, or $0.9\%$ and $3.1\%$ of the straddling
pairs, with a median $\Delta E$ of $8.8$ and $10.0$.
These pairs mark where the seams discussed in Sec.~\ref{sec:supp_limitations} can appear.
Both steps are larger on HanCo, at $1.32\times$ against $1.05\times$ the reference and $10.0$ against $8.8$
for the strict pairs.
This is consistent with a larger local correction on HanCo, where the mean
$\lVert\mathbf{c}^*_i-\mathbf{c}^0_i-\boldsymbol{\Delta}^{aid}_{s}\rVert$ over the Gaussians with
$g_i\gamma_i>0$ is $0.163$, against $0.056$ on InterHand2.6M.
The unobserved side, in contrast, receives only $\boldsymbol{\Delta}^{aid}_{s}$, which on HanCo reaches the
bound of $0.1$ per channel set by $\boldsymbol{\delta}_i=0.1\tanh(\cdot)$ (Sec.~\ref{sec:supp_architecture})
and amounts to a uniform brightening.

Figure~\ref{fig:supp_opacity} shows that the rendered opacity is equally high in every
view.
The mean of $A$ over the opaque region ($A\ge0.5$) is $0.99$ in every view of both datasets.
Semi-transparent pixels ($0.05\le A<0.5$) amount to $3.5$--$5.2\%$ of the opaque area, and this fraction is
similar on the observed and unobserved sides, for example $4.5\%$ in the source view and $3.9\%$ at
$180^\circ$ on InterHand2.6M.
The silhouettes show neither the translucent fringe of the recipe ports
(Fig.~\ref{fig:supp_recipe_ports_qual}) nor holes on the unobserved side.

\section{Supplementary Video}
\label{sec:supp_videos}
We provide a single supplementary video
(\texttt{InstantHand\_Supplementary\_Video.mp4}, 3\,min 9\,s) in four parts.
In every part, all of our renderings are projections of one fixed set of
canonical Gaussians per subject, with no view-dependent parameters, no
image-space refinement, and no per-frame fitting or update.
The avatar representation therefore cannot drift over time: frame-to-frame changes in
the renderings come only from the driving pose and the rendering itself, and
Sec.~\ref{sec:supp_temporal} quantifies them on HanCo.
Temporal stability and multi-view consistency are difficult to convey with
still frames, so we encourage readers to view the video directly.

\paragraph{Parts 1--2: benchmark subjects.}
Part~1 follows the InterHand2.6M protocol of
Sec.~\ref{sec:supp_eval_interhand} and Part~2 the zero-shot HanCo protocol of
Sec.~\ref{sec:supp_eval_hanco}, where the source subject is unseen and no
test-time optimization is performed.
In both parts the avatar is reconstructed from the protocol-designated source
frame with its ground-truth MANO, camera, and mask.
Each part shows (i) the single source input image together with the
reconstructed canonical avatar, (ii) full $360^\circ$ yaw and pitch rotations
of the canonical 3D avatar, and (iii) continuous novel-pose driving.
In the rotations, the prior-completed side comes into view as the hand turns
away from the source view, so the transition between the source-observed and
prior-completed sides can be inspected directly.
Each driving pose is rendered synchronously from six fixed cameras: the source
camera, the reference camera in which the driving sequence is defined, and
four fixed novel cameras at azimuth offsets of $+60^\circ$, $+120^\circ$,
$+150^\circ$, and $-60^\circ$ from that reference.
Both parts are driven by the same 35.2\,s MANO pose sequence, assembled from a diverse set of hand gestures
with short holds and interpolation between them.
The sequence is disjoint from the evaluation frames of either dataset, covers a wider range of poses than
either evaluation sequence, and shows the InterHand2.6M and HanCo avatars in identical poses.
No ground-truth images exist for these poses, so the driving segments are qualitative.
All quantitative novel-pose results (Tables~\ref{tab:main_quant}, \ref{tab:hanco_results},
and~\ref{tab:ablation} of the main paper) are computed on the evaluation sequences.

\paragraph{Part 3: unseen subjects under fully estimated inputs.}
Part~3 shows the eight unseen subjects of Fig.~\ref{fig:supp_animation}
(Sec.~\ref{sec:supp_animation}), reconstructed from fully estimated inputs
(SAM 3 mask and WiLoR MANO and camera, with no ground-truth annotations).
It first shows $360^\circ$ yaw rotations, together with a small pitch sweep, of
the canonical avatar for four subjects: a dorsal ARCTIC input, the palm-facing
FreiHAND input whose fragmented SAM 3 mask is used as is (row~2 of
Fig.~\ref{fig:supp_animation}), the dorsal in-the-wild OHTA input with a vitiligo
pattern, and a FreiHAND input with a darker skin tone.
In each, the side not visible in the source is completed from the
source-conditioned prior, so the rotations expose the transition between the
observed and completed sides.
Part~3 then drives all eight subjects, two at a time, with a shared MANO pose
sequence of diverse hand gestures, rendered from three cameras fixed in the MANO
frame at azimuths of $0^\circ$, $+45^\circ$, and $+90^\circ$.

\paragraph{Part 4: comparison with ground truth and per-subject baselines.}
Part~4 replays the HanCo evaluation sequence of
Table~\ref{tab:hanco_results} of the main paper from the source camera, placing the
ground-truth frames beside our zero-shot renderings and those of
HandAvatar~\citep{handavatar}, OHTA~\citep{zheng2024ohta}, and BIGS$^{*}$, which are
fitted to the same source frame, in the column order of Fig.~\ref{fig:qual} of the main paper.
Each column header lists the avatar-acquisition time of the method from Table~\ref{tab:hanco_results} of
the main paper.
Every frame of the sequence is shown in order, and all methods are driven by the
same frames, so the four renderings differ only in the method that produced them.
All panels are shown at the native $256\times256$ evaluation resolution without
resampling.
HanCo is recorded at 5\,Hz~\citep{zimmermann2021HanCo}, so consecutive evaluation frames
are 0.2\,s apart and the replay, which shows each frame for 0.1\,s, advances in visible
steps in every column, including the ground truth.
Part~4 closes with the same frames rendered by the same trained avatar with its
Gaussians densified at inference (Sec.~\ref{sec:supp_limitations}, 197K instead of 49K Gaussians, scales
$0.6\,s_i$), beside the ground truth and the $k=3$ rendering used
throughout the paper.
Part~4 is the only part that can be checked against real footage, because the
$360^\circ$ rotations and gesture sequences of Parts~1--3 were never recorded for
these subjects.

\section{Reproducibility Statement}
\label{sec:supp_reproducibility}

The full model is specified in the main paper (Eqs.~\ref{eq:img_tokens}--\ref{eq:app_loss}).
All loss terms are defined in Sec.~\ref{sec:supp_losses}, the Soft Alpha Band construction and schedule in
Sec.~\ref{sec:supp_sab}, the design rationale and thresholds of reliability gating (Eq.~\ref{eq:gamma}) in
Sec.~\ref{sec:supp_reliability}, and the network architecture in Sec.~\ref{sec:supp_architecture}.
Notation is consistent between the main paper and the appendix.
All real training datasets and both evaluation datasets are publicly available
(Sec.~\ref{sec:supp_dataset}), with preprocessing, annotation sources, subject and frame counts, and the
complete sampling policy documented in Secs.~\ref{sec:supp_training}--\ref{sec:supp_sampling}.
The synthetic training data are generated following the published protocol of \citet{Zhao_2025_CVPR}.
All loss weights, learning rates, schedules, and batch sizes are reported with their final values in
Sec.~\ref{sec:supp_training}.
Evaluation sequences, designated source frames, crops, masks, metric computation, boundary-aware silhouette
definitions, the scope of the self-reconstruction diagnostic, timing exclusions and the end-to-end runtime
breakdown, per-baseline input protocols, and the provenance of all baseline numbers are specified in
Secs.~\ref{sec:supp_eval}--\ref{sec:supp_baseline}.
The protocols of the pose-stratified, temporal, and canonical analyses are given in
Sec.~\ref{sec:supp_diag_protocols}.
The source-side MANO and mask sensitivity protocol and the inference-time reliability sensitivity settings
are specified in Secs.~\ref{sec:supp_source_sensitivity} and~\ref{sec:supp_gamma_sensitivity}, and the
inference-time densification in Sec.~\ref{sec:supp_limitations}.
All such variants keep the trained checkpoint fixed.
Training uses four NVIDIA RTX A6000 GPUs (approximately 25 and 13 hours for the two stages), and all timing
and efficiency measurements use a single NVIDIA A100 (80\,GB) under the protocol of
Sec.~\ref{sec:supp_timing}.
The reported $2.84$\,s is the avatar-network inference time once common MANO, camera, and mask inputs are
available, and the corresponding end-to-end raw-image-to-avatar time, including SAM 3 and WiLoR estimation,
is approximately $3.30$\,s (Table~\ref{tab:supp_runtime}).
Code, pretrained models, and the evaluation scripts are available at
\url{https://github.com/coolho1129/InstantHand}, and videos and further results are on the
project page, \url{https://coolho1129.github.io/InstantHand-page/}.

\end{document}